\documentclass{article} 
\usepackage{iclr2027_conference,times}

\usepackage{amsmath,amsfonts,bm}

\def\eqref#1{equation~\ref{#1}}

\def\1{\bm{1}}

\DeclareMathAlphabet{\mathsfit}{\encodingdefault}{\sfdefault}{m}{sl}
\SetMathAlphabet{\mathsfit}{bold}{\encodingdefault}{\sfdefault}{bx}{n}

\usepackage{hyperref}
\usepackage{url}
\usepackage{subcaption}
\usepackage{graphicx}
\usepackage{amsmath}
\usepackage{booktabs}
\usepackage{xurl}

\DeclareMathOperator{\relu}{ReLU}

\title{On Privacy in Data-Space Tabular Diffusion Models: Influential Factors, Attacker Knowledge, and Metrics}

\author{%
\textbf{Behnoosh Zamanlooy}$^{2,}$\footnote[2]{} \quad \textbf{Elaheh Bassak}$^{3,}$\footnote[2]{} \quad \textbf{Fatemeh Tavakoli}$^1$ \quad \textbf{Sara Kodeiri}$^1$ \\ \textbf{Marcelo Lotif}$^1$ \quad \textbf{Xi He}$^{1,4}$ \quad \textbf{Masoumeh Shafieinejad}$^{1,}\thanks{Corresponding Authors}$ \quad \textbf{D. B. Emerson}$^{1,*}$\\
$^1$Vector Institute \quad $^{2}$McMaster University \quad $^3$University of Toronto \quad $^4$University of Waterloo\\
\texttt{\{sara.kodeiri,marcelo.lotif,fatemeh.tavakoli\}@vectorinstitute.ai}\\
\texttt{\{masoumeh,david.emerson\}@vectorinstitute.ai} \\
\texttt{zamanlob@mcmaster.ca} \quad \texttt{e.bassak@mail.utoronto.ca} \quad \texttt{xi.he@uwaterloo.ca}
}

\iclrfinalcopy 
\begin{document}

\maketitle

\begin{abstract}

Tabular data plays an important role in many fields and industries, including those with elevated privacy considerations and risks. As such, there is a rising interest in generating high-quality synthetic proxies for real tabular data as a means of reducing privacy risk and proprietary data exposure. With data-space tabular diffusion models (TDMs) demonstrating leading performance in synthesizing such data, understanding and measuring the privacy risks associated with these models is imperative. Leveraging state-of-the-art membership inference attacks for such TDMs in both black- and white-box settings, this work quantifies the impact of training setup, synthesis choices, and attacker knowledge on privacy leakage. Moreover, the results demonstrate that adversaries need not have perfect knowledge of the training setup, identical data distributions, or massive compute resources to construct successful attacks. Finally, the substantial pitfalls associated with heuristic privacy metrics, such as distance-to-closest record, are identified.

\end{abstract}

\section{Introduction}

With the accelerating importance of machine learning (ML) models in all aspects of society, there has been a corresponding rise in interest in using advanced models to generate structured synthetic data \citep{Hernandez1, stoian2026a, fang2024large, shi2025comprehensivesurveysynthetictabular}. One of the most prominent modalities is tabular data, given its ubiquity across industries. High-fidelity data synthesis has many applications, including training data augmentation \citep{FAJARDO2021114463, DU2025101039, YEBIN2025115, IsmailFawaz2018, app11052158}, improving model robustness \citep{bartolo-etal-2021-improving, yang-etal-2020-generative, Ruiz1, moroianu2025improvingperformancerobustnessfairness}, and exploring hypothetical or novel circumstances. However, a primary driver of interest in synthetic tabular data is its potential as a proxy for real data with reduced privacy risks. In many settings, existing and emerging privacy legislation, such as GDPR, HIPAA, and PIPEDA \citep{GDPR2016a, pipeda2000, HIPAA1996}, regulate the collection, use, and release of sensitive information and datasets. Moreover, many institutions strictly manage internal risk associated with privacy breach or data misuse. As such, synthetic tabular data is regarded as a prospective solution to privacy risks. While synthetic data has many advantages and can increase barriers to privacy breach, it is not intrinsically privacy preserving \citep{stadler2022synthetic}, motivating significant research in privacy-preserving synthetic generation and privacy auditing techniques \citep{Shokri2017MIA, kazmi2024panoramia, Wen1, SANCHEZSERRANO2025110468, Ghatak1}.

Recent work has shown that data-space tabular diffusion models (TDMs) achieve leading performance in tabular synthesis along a number of metrics \citep{Pang2025ClavaDDPM, kotelnikov2023tabddpm,  ICLR2025_5c882988}. While some research suggests that diffusion models are robust to overfitting \citep{Bonnaire2025WhyDiffusionDontMemorize}, other work has shown that TDMs still leak information \citep{Ward1}, sometimes at higher rates than GAN-based counterparts \citep{zhu2025dptldmdifferentiallyprivatetabular}. Differentially private (DP) training can reduce privacy risks \citep{Jayaraman1, Annamalai1} but often comes with steep utility and efficiency tradeoffs, especially without large-scale public pretraining \citep{dockhorn2022differentially, Liu1, zhu2025dptldmdifferentiallyprivatetabular}. As such, it is important to identify and quantify the mechanisms that impact privacy leakage in TDMs to guide practical and safe use of such models.

In this work, state-of-the-art membership inference attacks (MIAs) for data-space TDMs are used as a lens to quantify privacy leakage. Such attacks are a fundamental privacy auditing technique for ML models \citep{Carlini2022MIA, Hayes2019Logan, Salem2019ML-Leaks, Shokri2017MIA}. Moreover, successful MIAs can be used to build deeper privacy attacks \citep{Salem1, Annamalai2}. Using leading white-box (WB) and black-box (BB) MIAs, we investigate the impact of training choices, architecture configurations, and generation decisions on privacy leakage. Further, the influence of attacker knowledge, compute power, and data access on MIA success is examined. Finally, the efficacy of distance-to-closest-record (DCR) \citep{SynthEval2024}, and other widely used pseudo-privacy metrics, in identifying rising or falling privacy risk is compared to MIAs. 

The results show that some traditional levers, such as training steps or dataset size, remain important for privacy in data-space TDMs. Others depend on the MIA technique or model, reinforcing the need to model a diversity of approaches, and some are surprisingly unimportant. Moreover, TDMs remain vulnerable to attacks even when adversaries have limited compute, imperfect knowledge, or corrupted data. Lastly, a nuanced and largely unreliable relationship between pseudo-privacy metrics, such as DCR, and MIA success is demonstrated. In some settings, DCR is well positioned to estimate rising MIA risk, while in others it fails entirely. These findings have broad implications for the practical use of TDMs and reinforce the importance of MIA auditing in synthetic data generation.

\section{Related Work}

MIAs have been established as a core technique for quantifying privacy risks in ML model training data \citep{Shokri2017MIA, Hayes2019Logan, Salem2019ML-Leaks, Chen2020GAN-Leaks, Carlini2022MIA}. The success of diffusion models has driven advances in MIAs tailored to these architectures, including TDMs. Model-agnostic, BB methods include DOMIAS \citep{Breugel2023MIAagainst}, RMIA \citep{pmlr-v235-zarifzadeh24a}, EPT \citep{german2025miaeptmembershipinferenceattack}, and others \citep{Ward1, Platzer2021Holdout}. For WB settings, the SecMI approach \citep{Duan2023SecMI} has been adapted for tabular models \citep{cheng2025membershipinferencediffusionmodelsbasedsynthetic, wu2025winning, yao2025dcrdelusionmeasuringprivacy, mahyar2026fermiexploitingrelationsmembership}. In this work, we focus on state-of-the-art approaches from the MIDST Challenge \citep{MIDST2025}, ensuring the experimental framework leverages strong MIAs in both WB and BB settings. 

Some existing work has considered factors that impact MIA success rates and utility in generative models for tabular data in controlled settings and for ML models more broadly. \citet{Salem2019ML-Leaks} propose dropout and model stacking as defenses against MIAs for classification models. For the simple set of models studied, it is shown that shadow models need not have identical architectures or data distributions for successful MIAs. In \citep{DelGobbo2025AComparative}, the effect of input-output ratio on the statistical similarity and predictive utility of synthetic data is studied. Overfitting is examined in \citep{Hu2022MIASurvey} as a key driver of MIA success across different models, including GANs and VAEs. A taxonomy of existing defenses is provided, with the majority of generative model defenses falling under DP techniques. They also exhibit that traditional methods that combat overfitting are not necessarily enough. Finally, for classification models, studies have shown that training data diversity plays an important role, with MIA success decreasing with growing training data \citep{Shokri2017MIA}. The present work focuses on a wide range of mechanisms and attack regimes that influence privacy leakage in data-space TDMs, going well beyond previous work.

Several studies investigate the interplay of DP and MIA success in the context of general ML models. Theoretical results demonstrate that small-$\epsilon$ DP provides an upper bound on MIA success \citep{erlingsson2020private, Yeom2017PrivacyRI}, while empirical studies suggest that larger $\epsilon$ values offer practical MIA protection \citep{Lowy2024WhyDD, Jayaraman2020RevisitingMI}. \citet{Humphries2020InvestigatingMI} investigate the impact of data dependencies on DP protections against MIAs for classifiers. The results indicate that DP does not fully protect against MIAs due to the correlative nature of the data distribution. While DP can provide privacy leakage protection, it can also significantly degrade synthetic data quality in diffusion models \citep{zhu2025dptldmdifferentiallyprivatetabular}. Moreover, experiments probing the mechanisms that reduce leakage in TDMs, as undertaken herein, provide a path to combining the benefits of high-$\epsilon$ DP with such levers to minimize privacy risk while maintaining maximum utility.

The impact of training decisions and attacker configurations on proxy metrics for privacy risks in synthetic data, especially distance-based measures, have been considered for TDMs. Experiments in 
\citep{fang2025understanding} investigate the influence of factors such as training epochs, dataset size, and model type on a thresholded form of NNDR \citep{steier2025syntheticdataprivacymetrics}. \citet{yao2025dcrdelusionmeasuringprivacy} examine DCR as a statistical indicator of privacy risk for such models. Therein, for a limited setup, DCR-based hypothesis tests fail to indicate leakage uncovered by MIAs. The experiments in this work significantly expand these studies in several directions. The relationship of multiple heuristic privacy metrics to MIA success across a variety of setups is investigated, revealing an unreliable relationship where, in some settings, these metrics reflect rising privacy risk and dramatically fail in others. Further, the impact of model training, attack configuration, and adversary knowledge are explored relative to MIA success. Finally, other BB MIAs that do not require internal parameters are considered.

\section{Methodology} \label{methodology}

This paper focuses on single-table synthesis using two frontier, data-space TDMs, ClavaDDPM \citep{Pang2025ClavaDDPM}, an improved form of TabDDPM \citep{kotelnikov2023tabddpm}, and TabDiff \citep{ICLR2025_5c882988}. The main lens through which privacy is assessed is that of MIAs. Provided a set of data points and a trained target model, the goal of an MIA is to distinguish between points that were members of the model's training dataset and those that were not. In the domain of synthetic data, such methods are commonly broken into two categories, WB and BB attacks. BB attacks strictly operate on the synthetic outputs of the target model, potentially with access to related public data. Alternatively, WB attacks are afforded full-model access, including weights. Experiments are conducted with the Tartan Federer (TF) \citep{wu2025winning} and Ensemble \citep{EnsembleMIA} approaches from the MIDST Challenge \citep{MIDST2025}. The TF method won the competition in both WB and BB settings, while the Ensemble technique placed second for BB attacks.

The Ensemble and TF attacks are fundamentally different in kind. The Ensemble method combines several distance-based approaches, probability mass estimations in DOMIAS, and pairwise likelihood ratio tests in RMIA. The approach only requires access to target-model generated synthetic data and real, similarly distributed background data as a reference. As such, it is not directly tied to diffusion architectures. However, its success can rely more heavily on auxiliary data availability. This is explored in Appendix \ref{ensemble_ablation_results}. Alternatively, whether applied as a WB or BB attack, the TF approach is tailored to TDMs, adapting the SecMI algorithm. Briefly, given a data point $x_0$, a forward process producing noisy samples at time $t$ such that $x_t = \sqrt{\alpha_t} x_0 + \sqrt{1-\alpha_t} \epsilon$ for some sampled Gaussian noise $\epsilon \sim \mathcal{N}(0, I)$, and a backward process model, $m_{\theta}(x_t, t)$ that predicts the noise in $x_t$, the TF attack extracts loss values for a candidate $x_0$ as $\Vert m_{\theta}(x_t, t) - \epsilon \Vert_2^2$ across a wide range of $\epsilon$ and $t$ values. Employing shadow models, these losses are used to train a dense neural network (DNN) for membership classification. In WB settings, target models are directly available to produce such losses at inference time. For the BB attack, proxies are constructed by training additional, target shadow models on the target's synthetic data from which the loss values are then extracted. For ClavaDDPM, the TF attack from \citep{MIDST2025} is applied, while the modified form in \citep{mahyar2026fermiexploitingrelationsmembership} is used for TabDiff. Additional details are provided in Appendix \ref{tf_appendix}.

\subsection{Attack Setup, Metrics, and Datasets} \label{datasets_and_defaults}

For a dataset $\mathcal{D}$, points are split into two subsets. The first, denoted $\mathcal{D}_t = \bigcup_{i=1}^{n_t} \{\mathcal{D}^i_t\}$, is used to train a set of target models, $\{m_{\theta}^i\}_{i=1}^{n_t}$. This dataset is strictly unavailable to an attacker. The second, $\mathcal{D}_h$, is a set of reference data available, in some form, to the attacker for the development of shadow models or calculating statistics between synthetic and real data. In both WB and BB setups, attackers are provided with synthetic datasets, $\mathcal{D}_s^i$, generated by $m_{\theta}^i$. Adversaries are also assumed to have knowledge of target-model architectures and training procedures, including hyperparameters. Traditionally, oracle knowledge of training and architecture elements is assumed. Below, however, settings with imperfect knowledge are also extensively explored. In the WB framework, all internal parameters for a target model $i$ are also revealed to attackers.

For each target model, membership challenge points, $\mathcal{C}_i$, are drawn from $\mathcal{D}$. Half are from of $\mathcal{D}_t^i$ and half are not. An MIA aims to provide an accurate distinction between such points across all target models. Following \citet{Carlini2022MIA}, MIA success is measured as the true-positive rate (TPR) at a false-positive rate (FPR) of $0.1$. Throughout, this is also simply referred to as attack success rate and privacy risk. Expected random classifier success in this metric is $0.1$. Several pseudo-privacy metrics are also measured, including DCR, NNDR, Hitting Rate, and EIR. DCR is reported in the main results as a representative metric, while definitions and results for others appear in Appendix \ref{other_privacy_heuristics}. Define $\mathcal{D}^i_h \subset \mathcal{D}_h$, a holdout datasets for model $i$. Let $x_s \in \mathcal{D}_s^i$. Denote by $x_t \in \mathcal{D}^i_t$ and $x_h \in \mathcal{D}^i_h$ points in the respective subsets closest, in some measure $d(\cdot, \cdot)$, to $x_s$. DCR is computed as the proportion of $x_s$ where $d(x_s, x_t) < d(x_s, x_h)$. An ideal DCR is given as $\vert \mathcal{D}^i_t \vert / \vert \mathcal{D}^i_t \cup \mathcal{D}^i_h \vert$. 

Several synthetic tabular-data quality metrics are also computed, including $\alpha$-precision and $\beta$-recall \citep{alaa2022faithful}, KS statistics and TVD \citep{Dankar1}, correlation and MI differences \citep{Zhu1, Ping1}, and machine-learning efficacy \citep{maheshwari2024efficacysyntheticdatabenchmark}. Definitions and values appear in Appendix \ref{quality_metrics}. The principle function of these metrics is to demonstrate that synthetic data quality does not collapse, decoupling MIA success from significant utility degradation.

Two large-scale tabular datasets are used in the experiments. The first is the Transactions table from the Berka dataset \citep{BerkaDataset} consisting of financial records from a Czech Bank, collected in 1999. The table has four numerical and four categorical columns with over $10^6$ rows of account transactions. The second table is the Diabetes dataset \citep{diabetes_dataset} with records derived from 130 US hospitals from 1999-2008. Each row represents a patient with diabetes and includes both numerical and categorical columns associated with demographics, clinical tests, and other health data. The dataset is smaller than Berka, containing just over $10^5$ records, but incorporates $47$ total features. For each dataset, row-level MIAs are conducted. Meaningfully, these datasets are larger than many in previous work. Preprocessing and MIA construction details appear in Appendix \ref{dataset_prepocessing}.

\subsection{Experiment Configurations: Training, Synthesis, and Attacker Knowledge} \label{experiment_configurations}

The influence of training setup, generation choices, and attacker knowledge on privacy risk is quantified through an extensive set of experiments. To start, a set of default configurations for ClavaDDPM, adapted from \citep{Pang2025ClavaDDPM}, and TabDiff, based on \citep{mahyar2026fermiexploitingrelationsmembership}, are fixed for each dataset and MIA. These defaults are outlined in Appendices \ref{experiment_details}--\ref{ensemble_appendix}. Sections \ref{steps_size_model} and \ref{diffusion_batch_ratio} consider the impact of varying configurations on the success of WB and BB TF attacks across models. In Section \ref{steps_size_model}, target-model training steps, training dataset size, and model composition are independently perturbed from the defaults. Details on model architecture variations appear in Appendix \ref{experiment_details}. Experiments in Section  \ref{diffusion_batch_ratio} vary training batch size and diffusion timesteps. The influence of adjusting the ratio of synthetic data generated to training set size with respect to BB attack success is also evaluated. Target models generate synthetic data at various multiples of their training data, providing growing pools of synthetic data for attackers relative to training set size. Section \ref{black_box_generalization} extends these results to the Ensemble attack setting, demonstrating generalization of the findings to other attack regimes. Several aspects specific to the Ensemble attack are also analyzed. Attack components are ablated to understand which contribute most to success. Different RMIA shadow model configurations are considered to understand the sensitivity of the algorithm to such choices. Finally, robustness to reference data volume is quantified. Results are reported in Appendix \ref{ensemble_ablation_results}.

Each setup above considers a standard MIA framework where attackers are assumed to be able to train many shadow models, perfect knowledge of the target-model training setup, and access to independently and identically distributed (IID) reference data. In Section \ref{resources_knowledge_data}, these assumptions are weakened in separate ways. First, MIA success as a function of the number of shadow models is considered, simulating resource constraints. Next, attackers train shadow models with configurations that diverge from the target model. Shadow models are trained for fewer or more iterations, use fewer diffusion timesteps, differ in size, or have different training set sizes compared to the target models. Finally, we consider settings where the data available to the adversary differs statistically from target-model training data. For Berka, four scenarios are explored. In the first, adversary and target model data is drawn from disjoint accounts. Second, the adversary observes data from a later time period, while target models are trained on earlier transactions. Third, adversary data is corrupted such that half of the categorical and numerical features are replaced with uniform draws across their respective ranges. Finally, independent sampling from one-way marginal distributions is simulated to constructed attacker data. For Diabetes, the final two scenarios are again used to corrupt adversary data. Appendix \ref{attack_mismatch_details} provides more details about the shadow model and dataset mismatch experiments, including quantifying the statistical divergence of the adversary datasets.

\section{Main Results} \label{main_results}

The experiments in this section are structured to systematically probe the relationship between privacy leakage and training setup, attacker knowledge, and compute power. Where appropriate, the connection, or lack thereof, between MIA success and DCR is also exhibited. Unless otherwise specified, when varying different aspects, such as training steps, all other settings remained fixed to the default values discussed in Section \ref{methodology} and Appendices \ref{experiment_details}--\ref{ensemble_appendix}.

\begin{figure}[ht!]
\begin{subfigure}{0.3175\textwidth}
\includegraphics[width=0.99\linewidth]{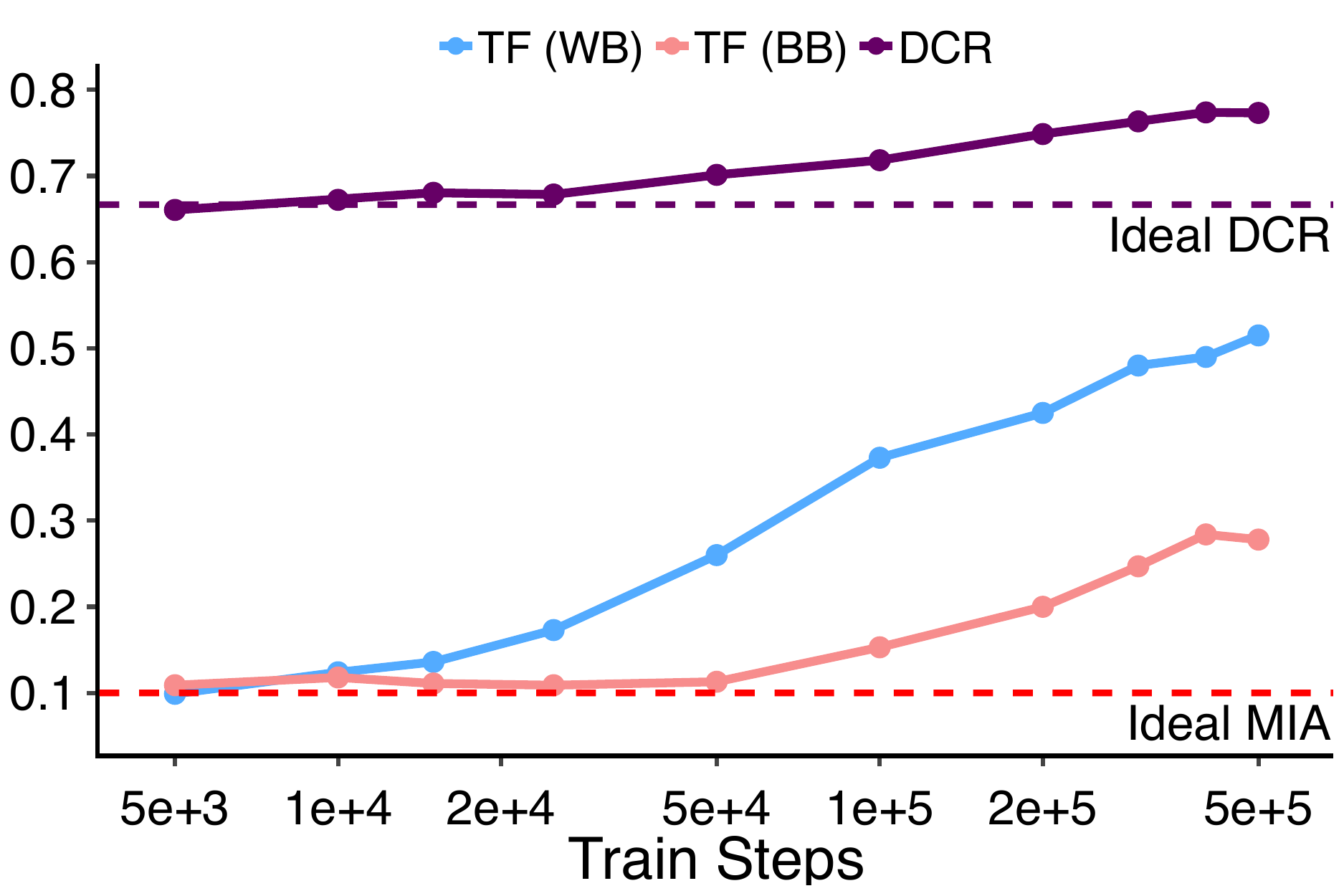}
\end{subfigure}
\begin{subfigure}{0.3175\textwidth}
\includegraphics[width=0.99\linewidth]{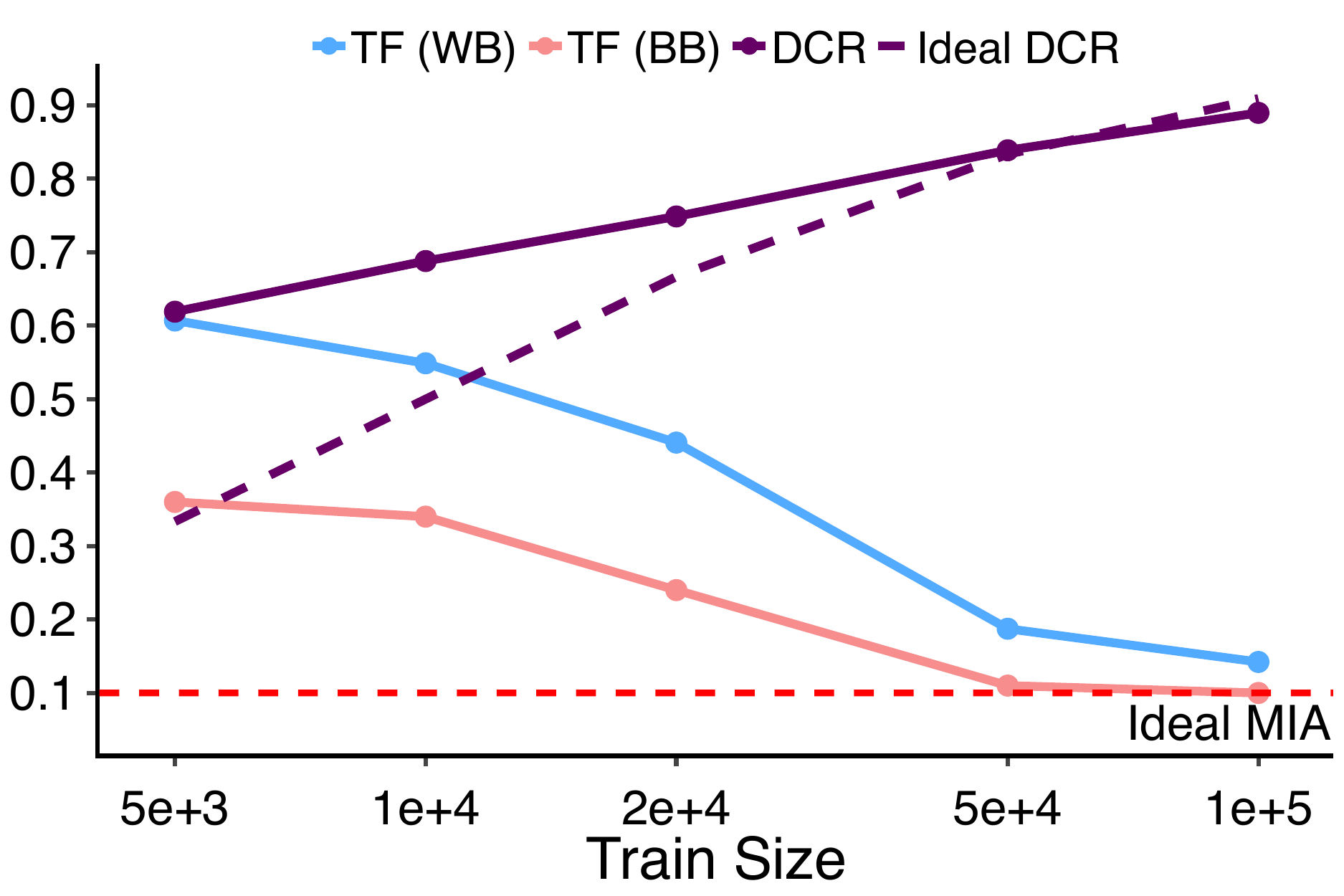}
\end{subfigure}
\raisebox{-1ex}{
\begin{subfigure}{0.35\textwidth}
\includegraphics[width=0.99\linewidth]{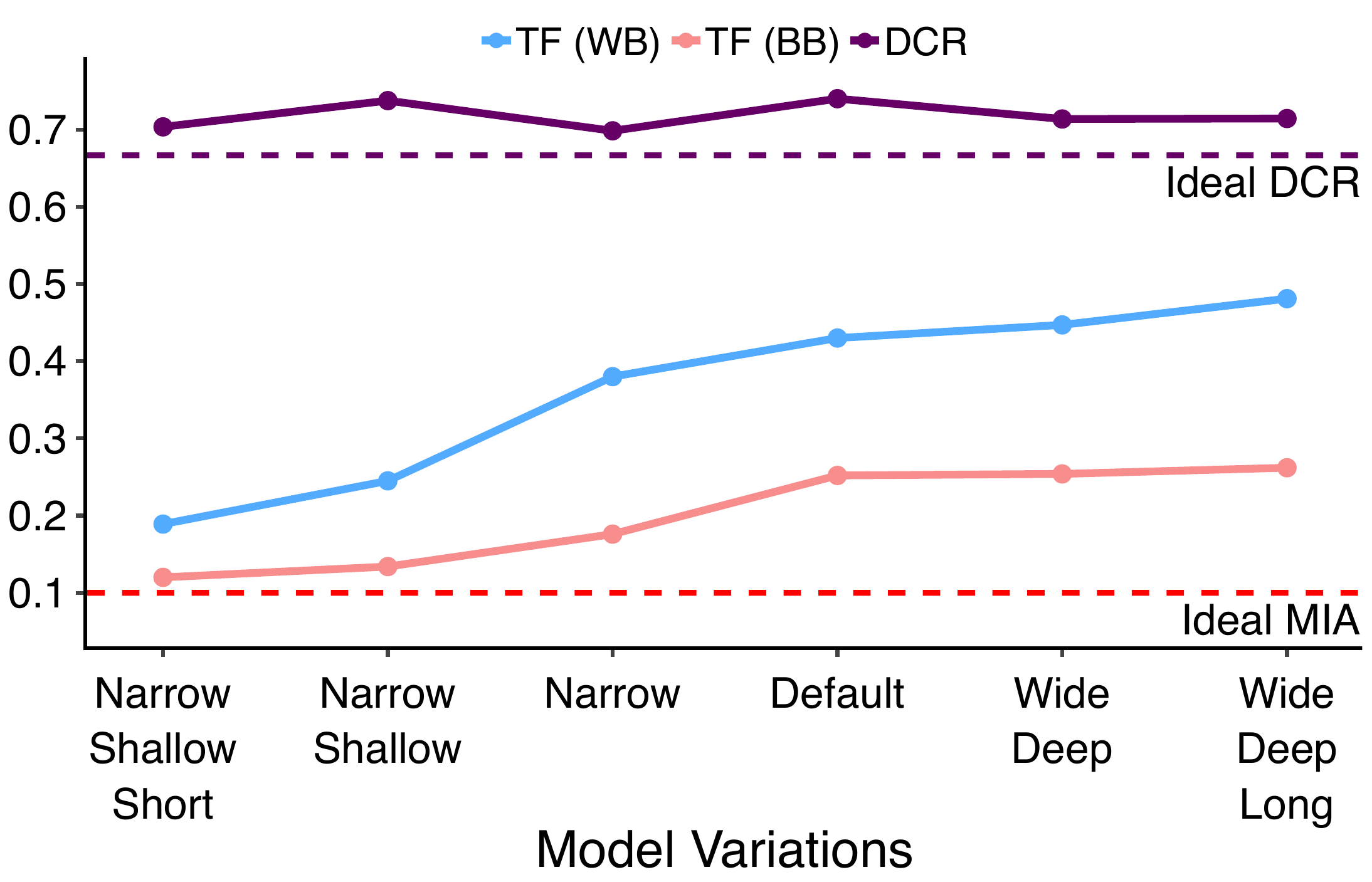}
\end{subfigure}}
\begin{subfigure}{0.3175\textwidth}
\includegraphics[width=0.99\linewidth]{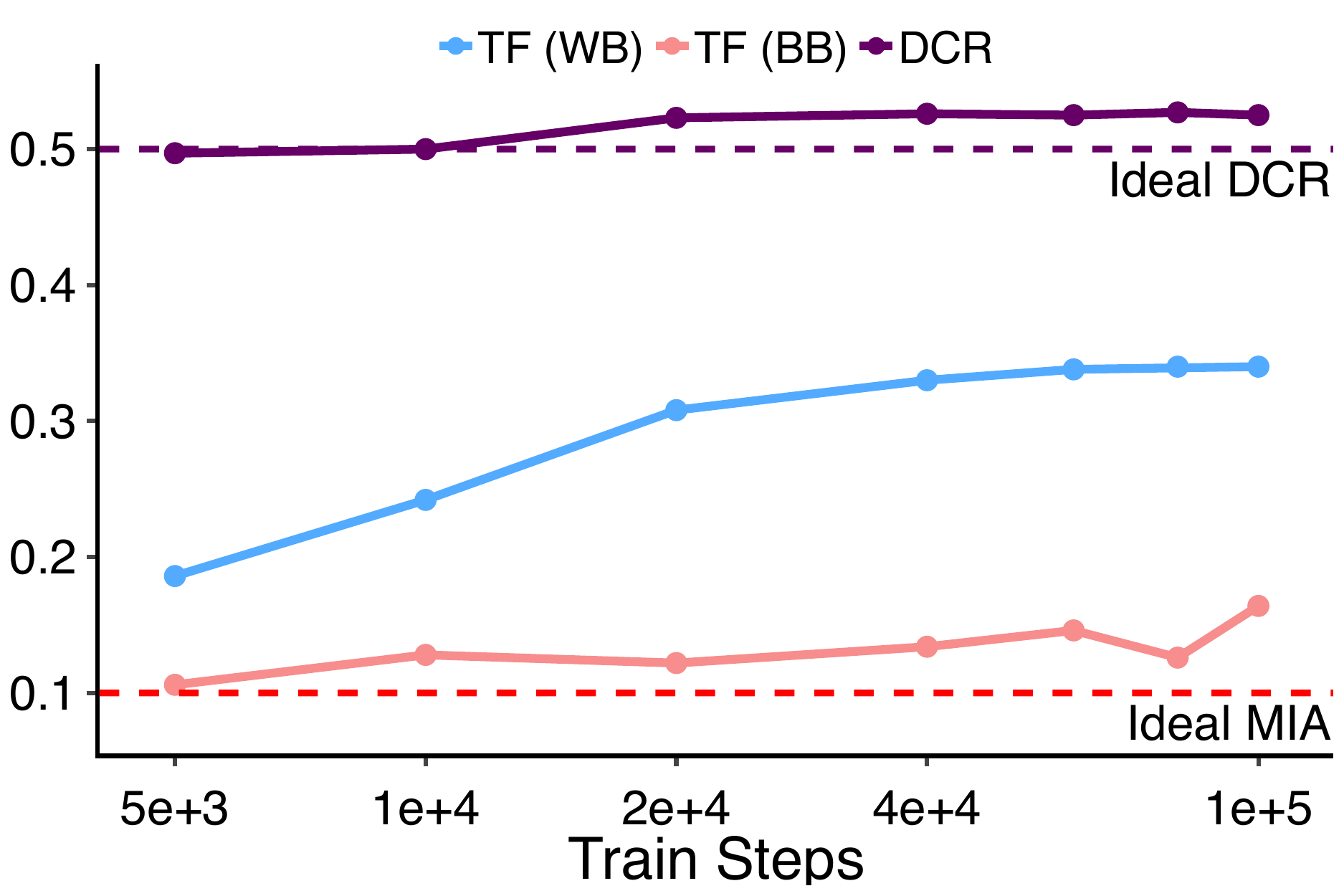}
\end{subfigure}
\begin{subfigure}{0.3175\textwidth}
\includegraphics[width=0.99\linewidth]{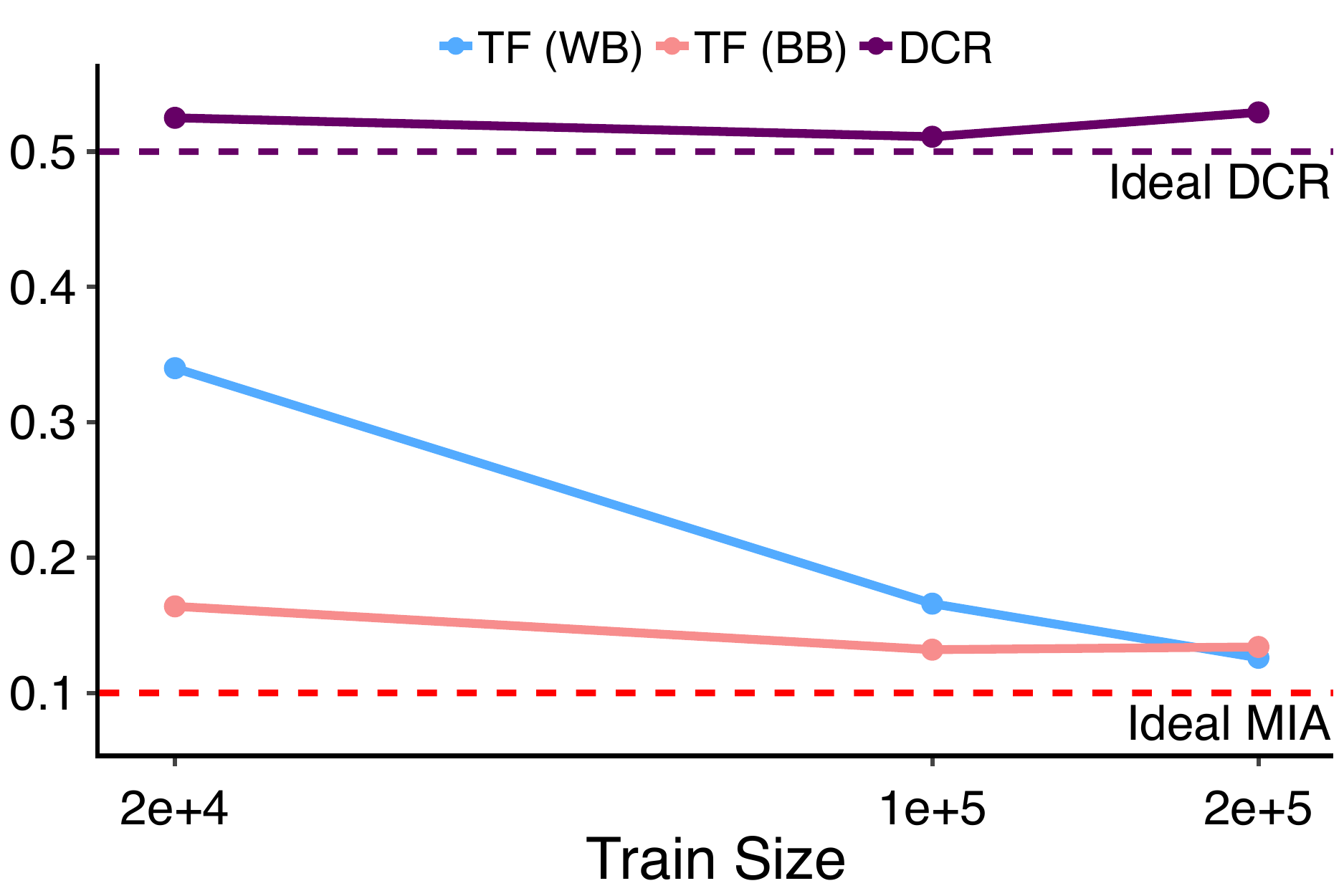}
\end{subfigure}
\raisebox{-1ex}{
\begin{subfigure}{0.35\textwidth}
\includegraphics[width=0.99\linewidth]{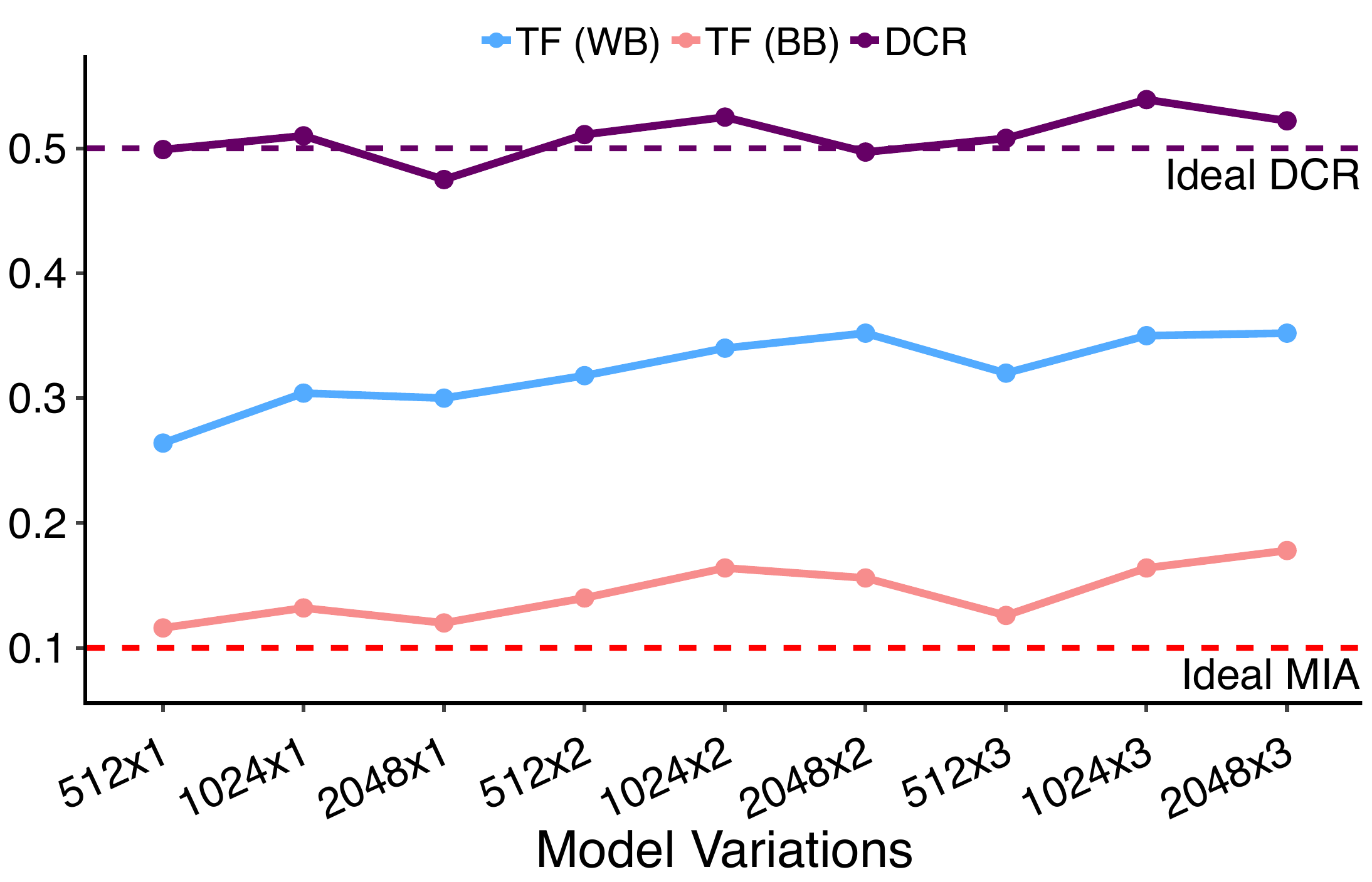}
\end{subfigure}}
\begin{subfigure}{0.3175\textwidth}
\includegraphics[width=0.99\linewidth]{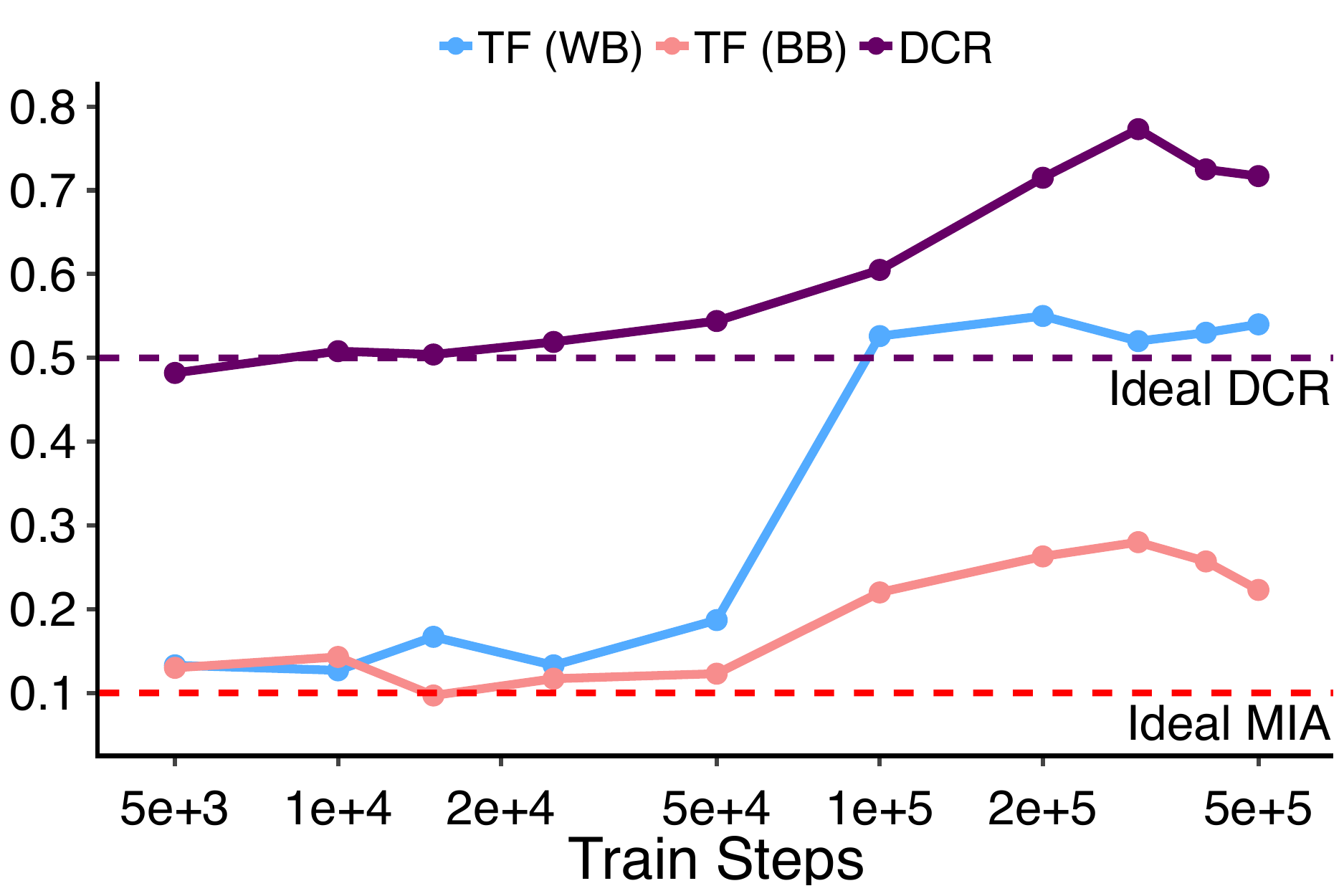}
\end{subfigure}
\begin{subfigure}{0.3175\textwidth}
\includegraphics[width=0.99\linewidth]{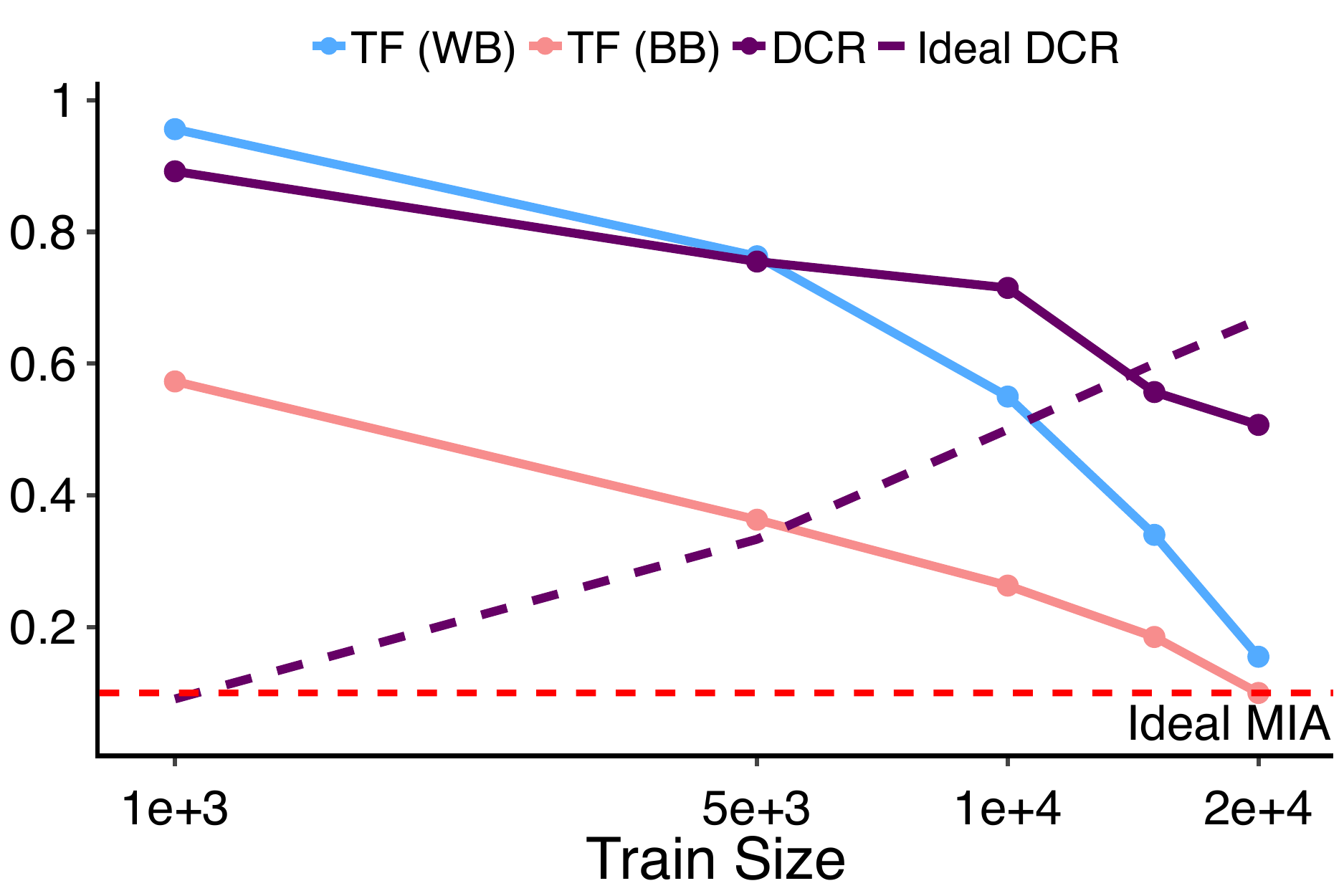}
\end{subfigure}
\raisebox{-1ex}{
\begin{subfigure}{0.35\textwidth}
\includegraphics[width=0.99\linewidth]{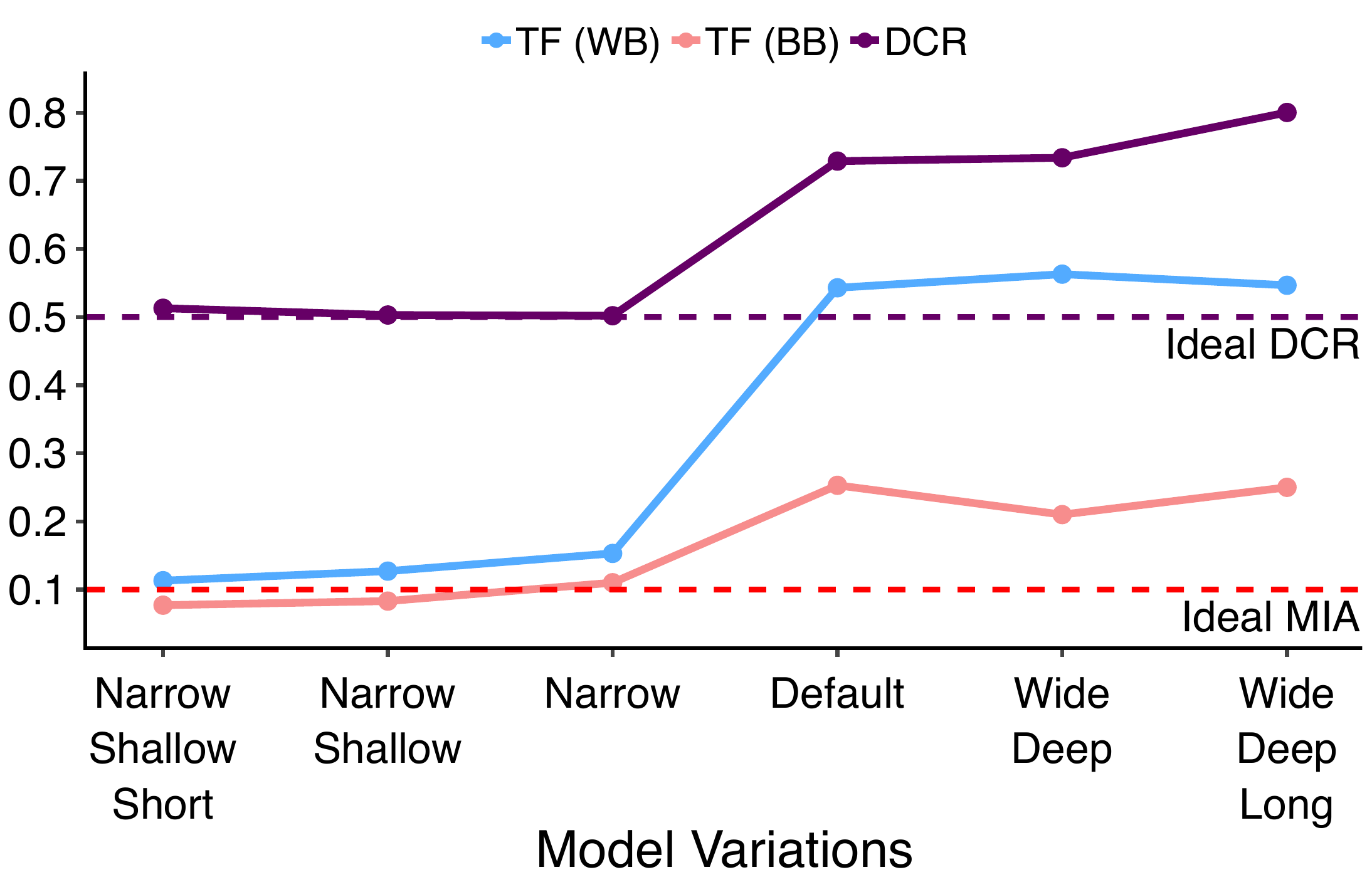}
\end{subfigure}}
\begin{subfigure}{0.3175\textwidth}
\includegraphics[width=0.99\linewidth]{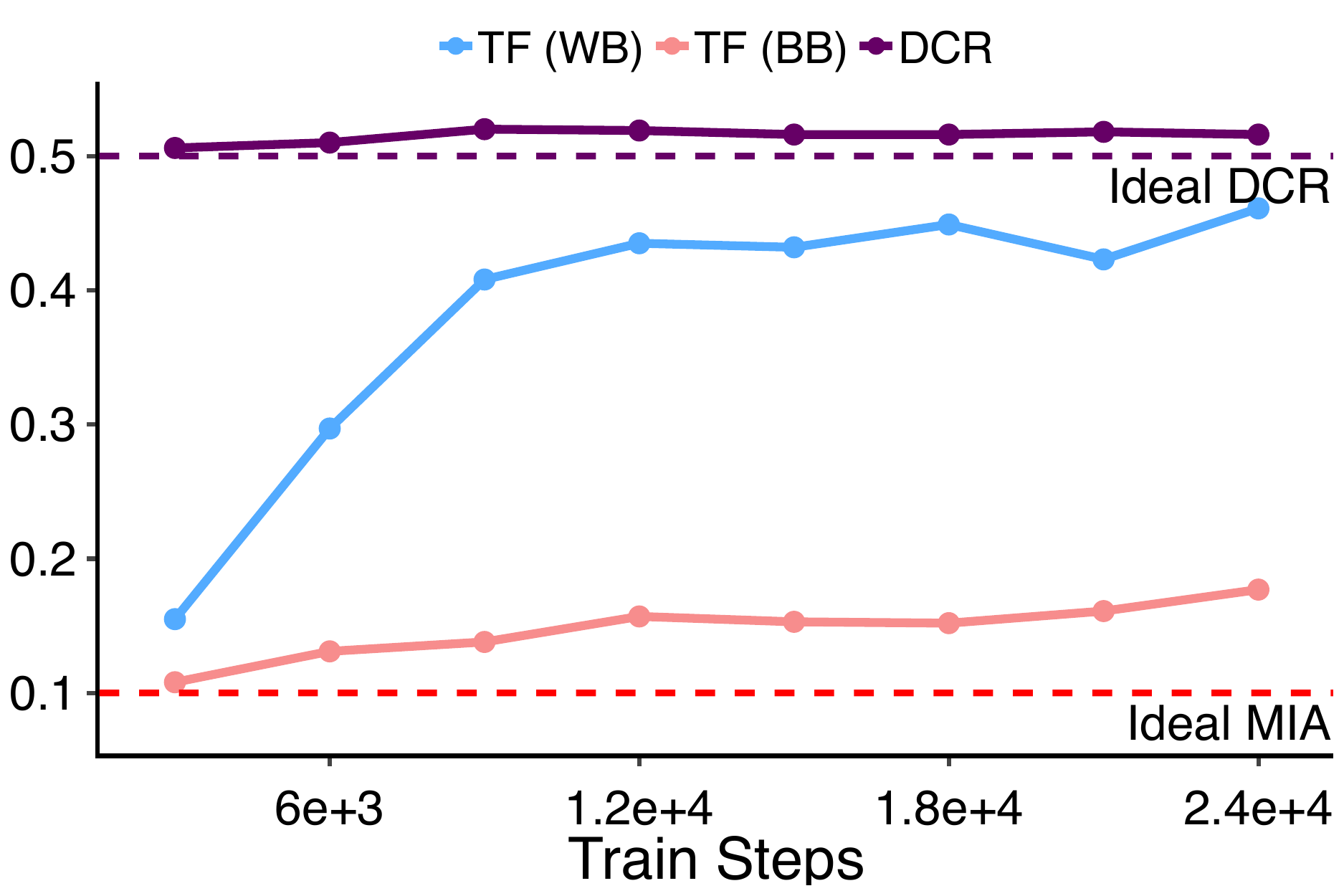}
\end{subfigure}
\begin{subfigure}{0.3175\textwidth}
\includegraphics[width=0.99\linewidth]{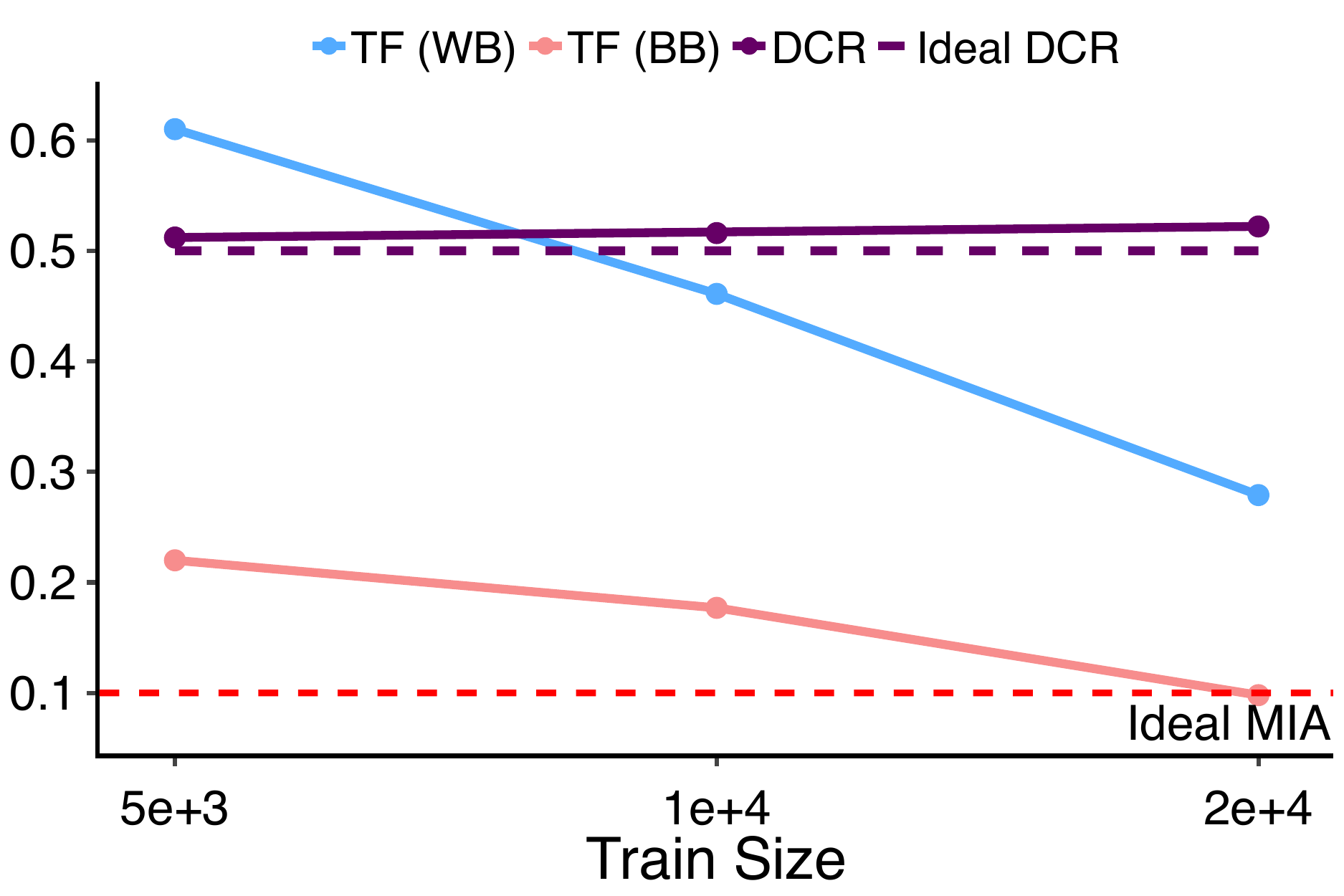}
\end{subfigure}
\raisebox{-1ex}{
\begin{subfigure}{0.35\textwidth}
\includegraphics[width=0.99\linewidth]{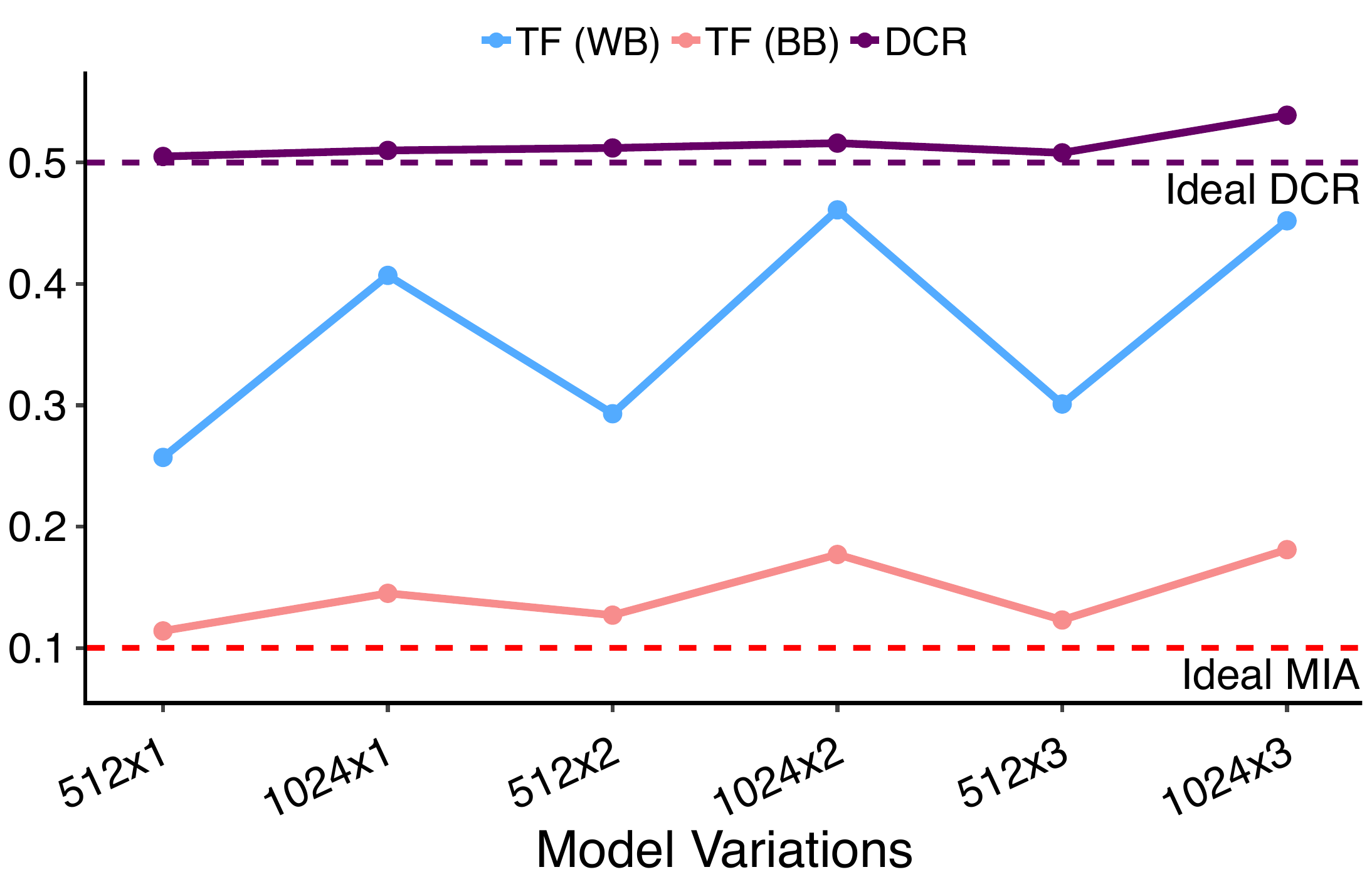}
\end{subfigure}}
\caption{Variations in training steps, training dataset size, and architecture vs. TF MIA success and DCR in WB and BB settings for Berka (Rows 1, 2) and Diabetes (Rows 3, 4) across ClavaDDPM (Rows 1, 3) and TabDiff (Rows 2, 4) target models.} \label{fig:tf_training_variations}
\end{figure}

\subsection{Training Steps, Training Size, and Model Variations} \label{steps_size_model}

In this section, the number of training steps, training dataset size, and diffusion architecture are modified for both ClavaDDPM and TabDiff models. The influence of such variations on privacy leakage, measured via WB and BB TF attacks, is quantified. Despite previous research suggesting that diffusion models are more robust to memorization \citep{Bonnaire2025WhyDiffusionDontMemorize}, the number of training steps and amount of training data have a stark impact on privacy leakage. Increasing training iterations coincide with a nearly monotonic increase in MIA success for both TDMs in Figure \ref{fig:tf_training_variations} (left). Figure \ref{fig:tf_training_variations} (center) shows that larger training datasets dramatically reduce MIA success across all models and datasets. For ClavaDDPM, these relationships are generally captured by DCR, where it diverges from and converges to ideal values when increasing training steps or dataset sizes, respectively. However, it fails to capture the scale of rising leakage for the WB TF attack with increasing training steps for both datasets. On the other hand, DCR is quite poorly calibrated for TabDiff, hovering near the ideal value in all settings, despite significant privacy leakage and dramatic changes.

Figure \ref{fig:tf_training_variations} (right) exhibits attack success as a function of architecture changes. In line with previous work on simpler ML models, TDMs incorporating more parameters and longer training runs induce rising MIA success. Furthermore, the DCR metric demonstrates significant unreliability for both TDMs. With ClavaDDPM, DCR largely reflects attack success for Diabetes but materially fails to do so in the case of Berka. A similar pattern is seen for other pseudo-privacy metrics in Appendix \ref{other_privacy_heuristics}. The larger number of training points and smaller set of columns in Berka potentially prevents simple distance-based output-space collapse in the generation process of ClavaDDPM, despite the potential for leakage clearly being present. The ablation studies in Appendix \ref{ensemble_ablation_results}, where none of the three Ensemble modules on their own produce significant MIA success, also suggests this. For TabDiff, DCR captures almost none of the observed leakage variations for either dataset, in spite of rising MIA success. In nearly all TabDiff experiments, DCR demonstrates poor sensitivity, signaling that pseudo-privacy measurements carry even less utility for this model compared to ClavaDDPM.

\begin{figure}[ht!]
\centering
\begin{subfigure}{0.3175\textwidth}
\includegraphics[width=0.99\linewidth]{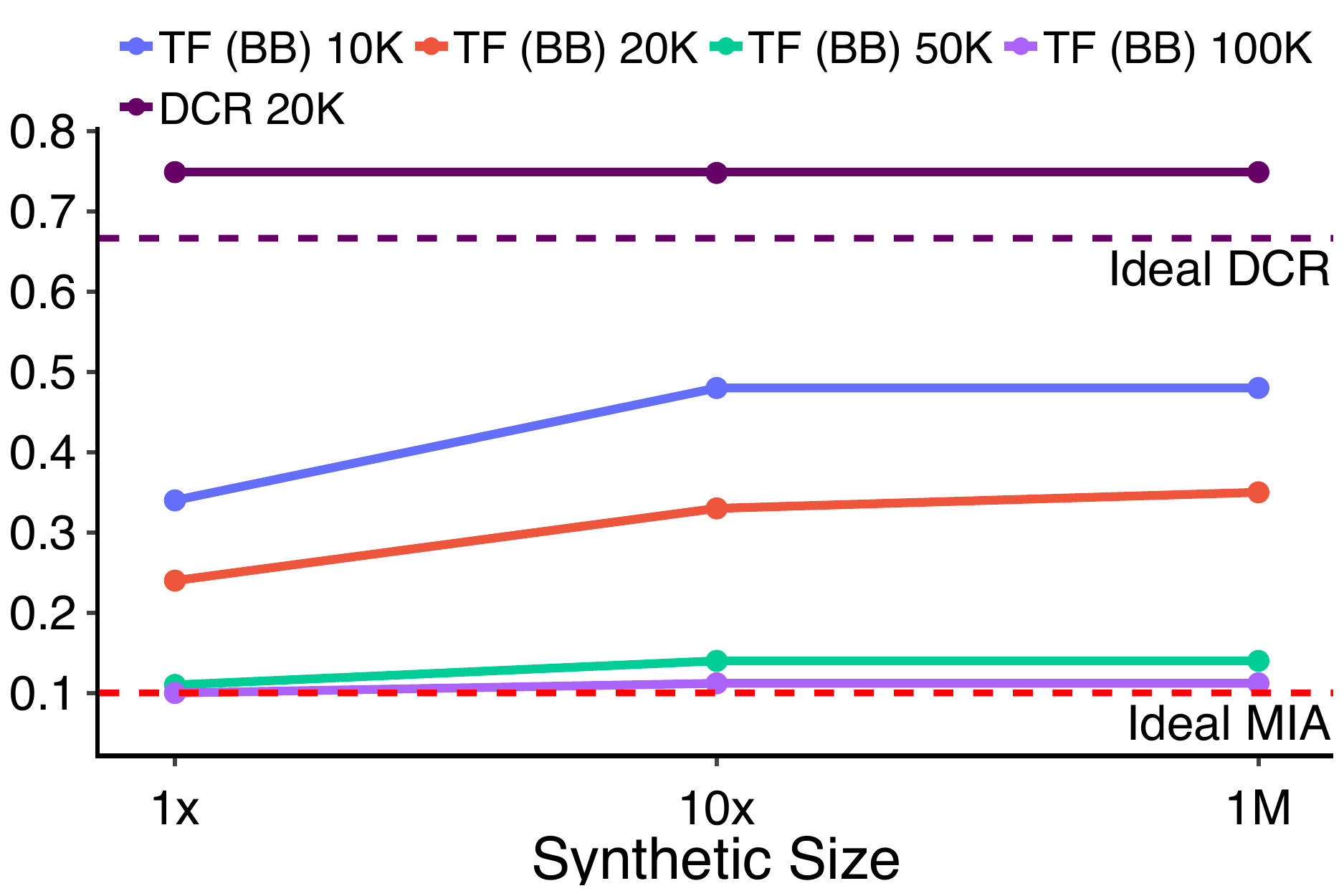}
\end{subfigure}
\begin{subfigure}{0.3175\textwidth}
\includegraphics[width=0.99\linewidth]{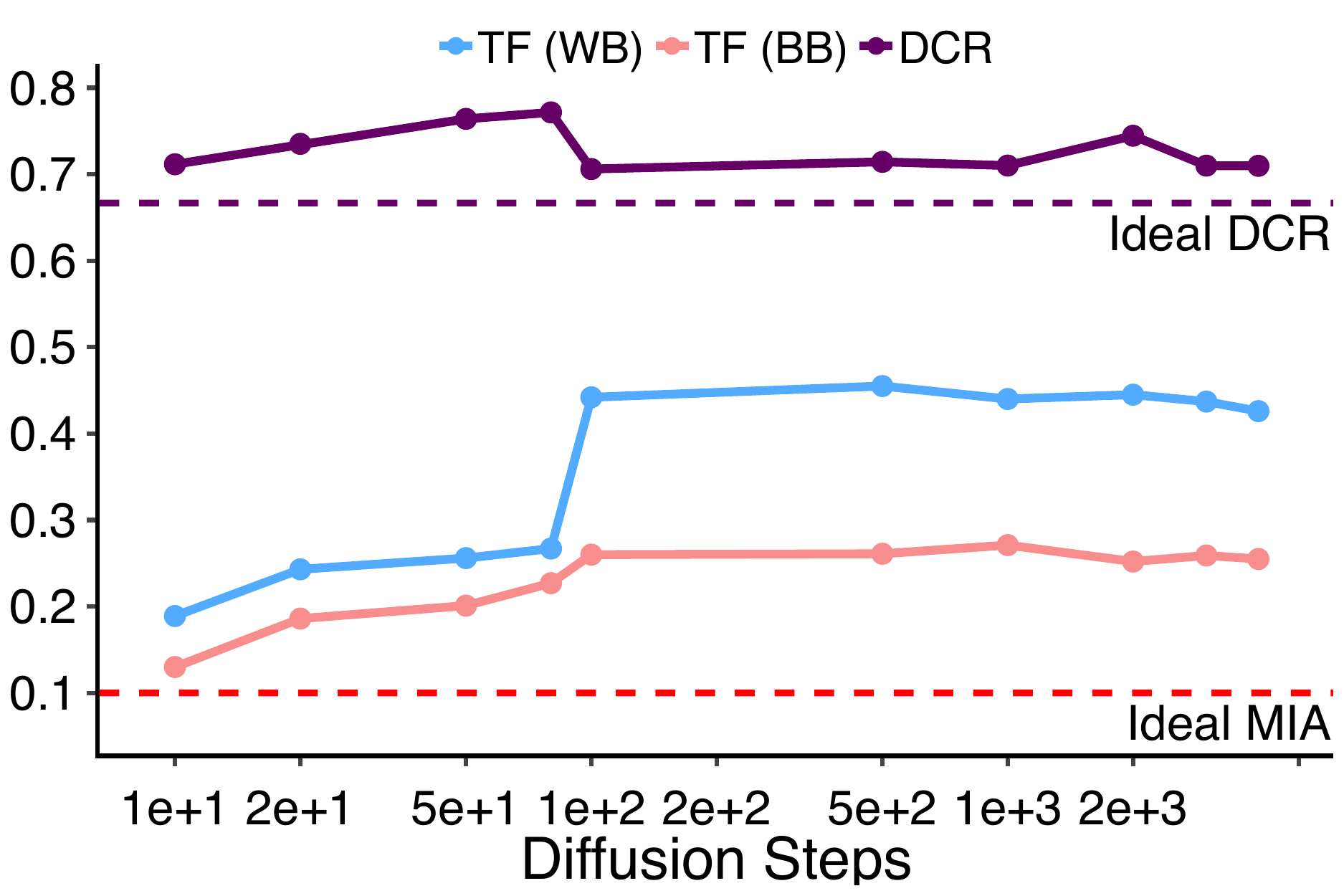}
\end{subfigure}
\begin{subfigure}{0.3175\textwidth}
\includegraphics[width=0.99\linewidth]{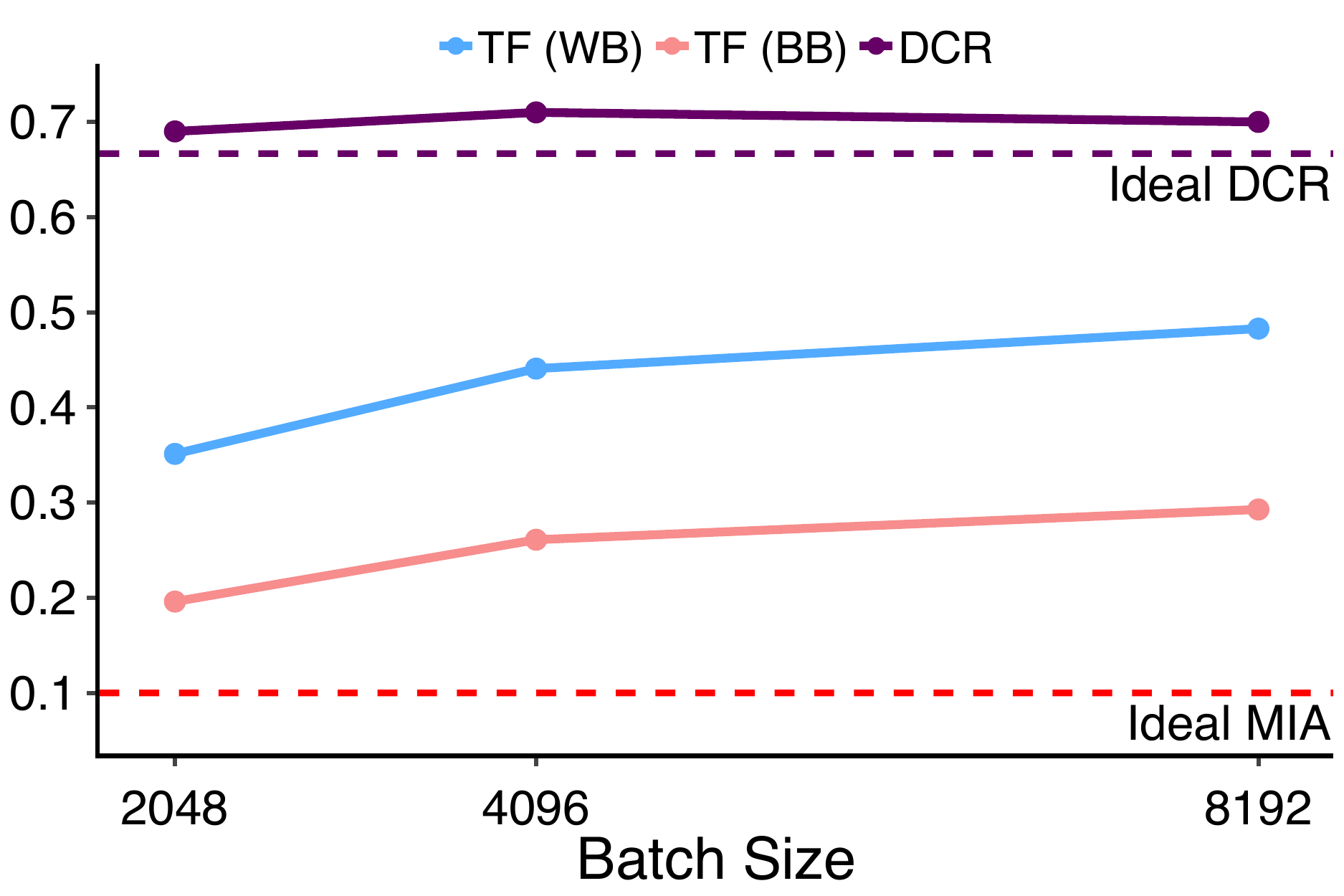}
\end{subfigure}
\begin{subfigure}{0.3175\textwidth}
\includegraphics[width=0.99\linewidth]{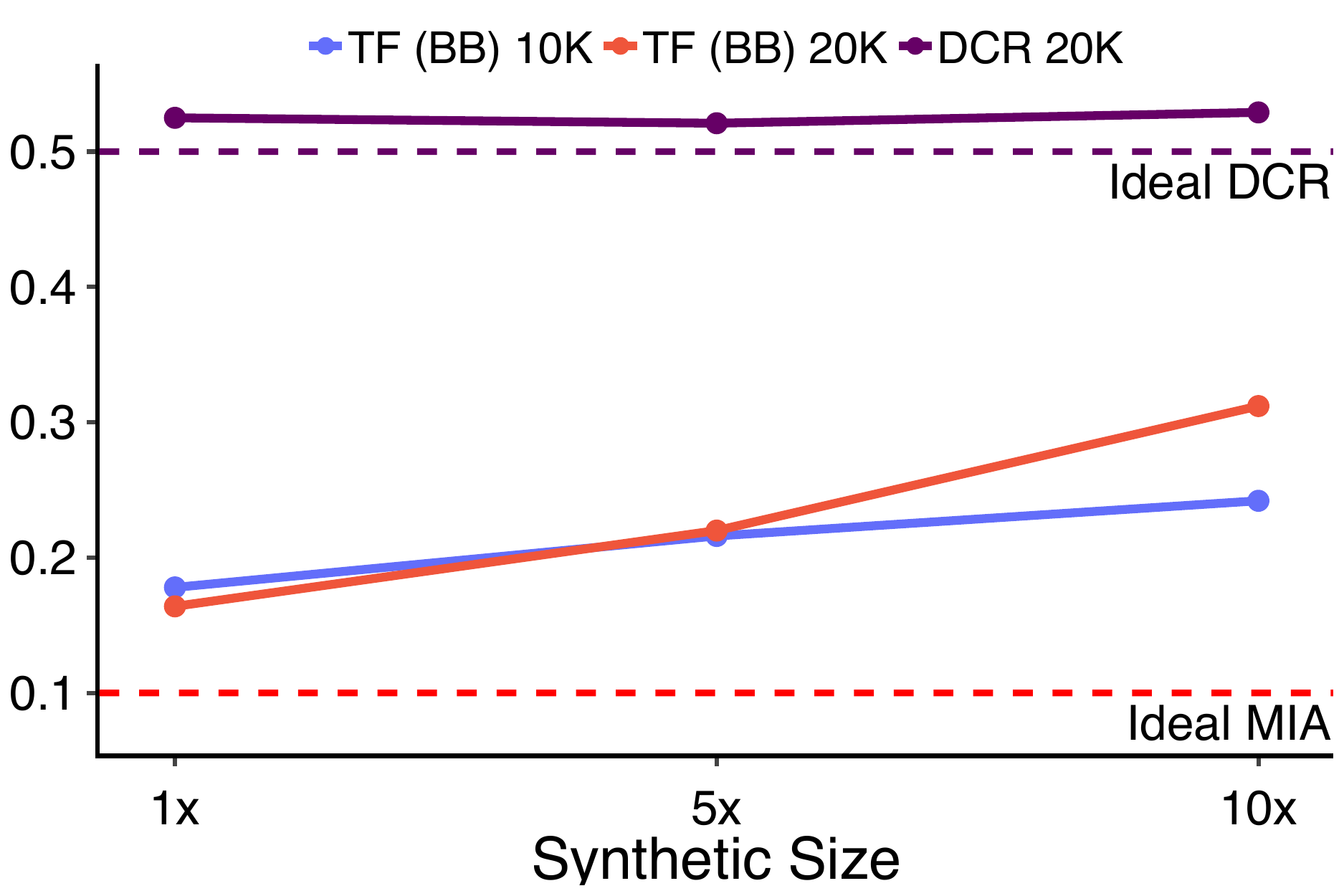}
\end{subfigure}
\begin{subfigure}{0.3175\textwidth}
\includegraphics[width=0.99\linewidth]{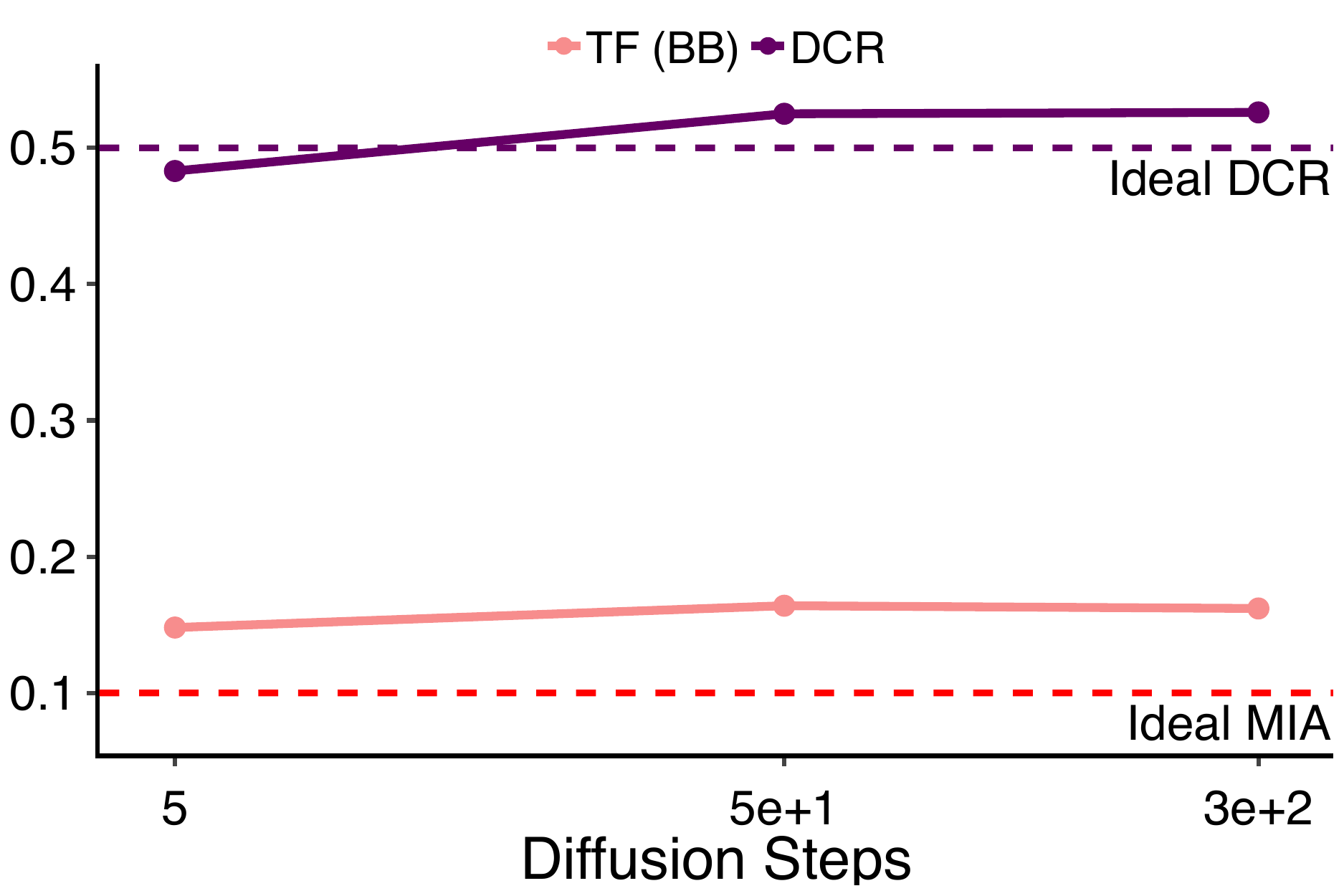}
\end{subfigure}
\begin{subfigure}{0.3175\textwidth}
\includegraphics[width=0.99\linewidth]{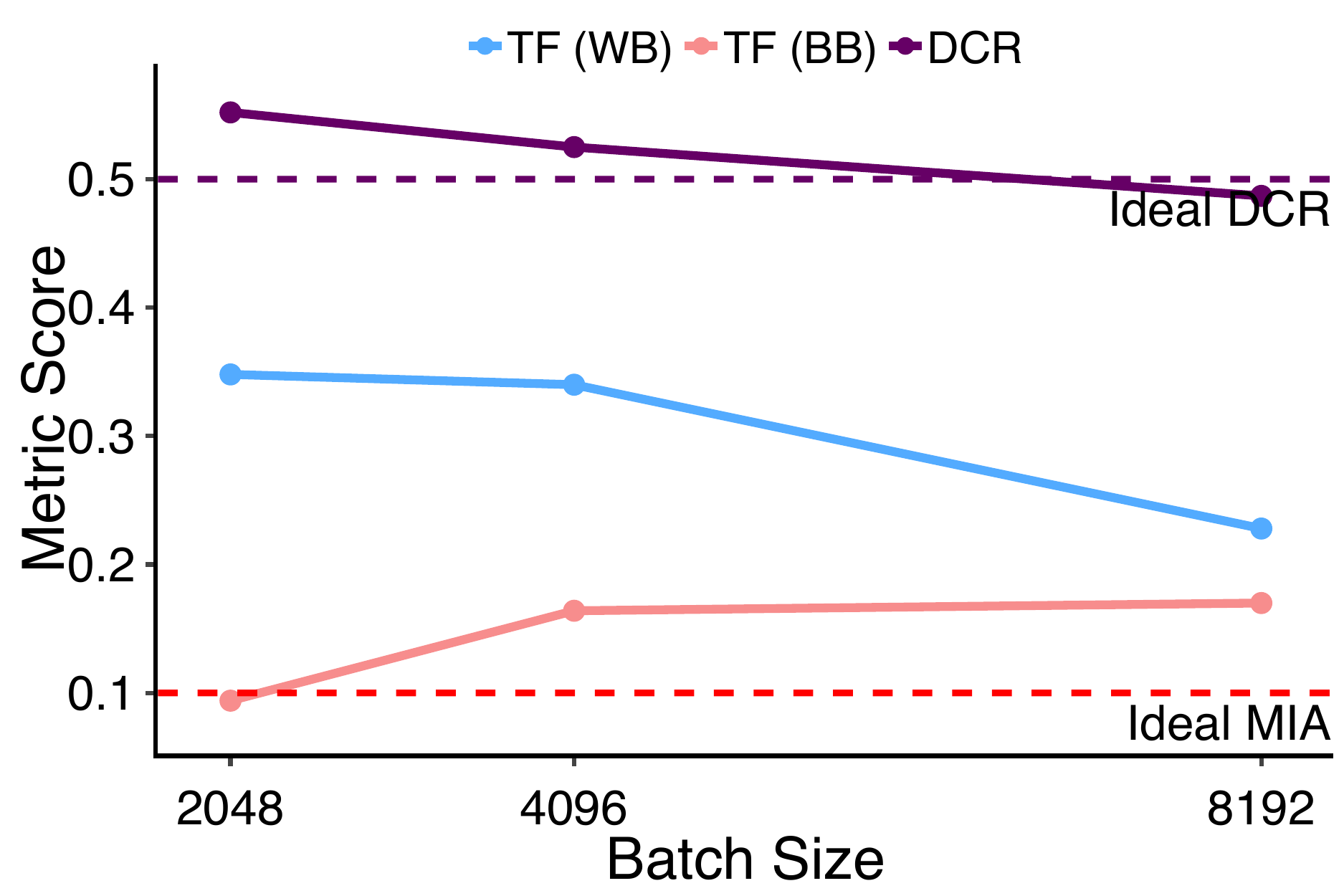}
\end{subfigure}
\begin{subfigure}{0.3175\textwidth}
\includegraphics[width=0.99\linewidth]{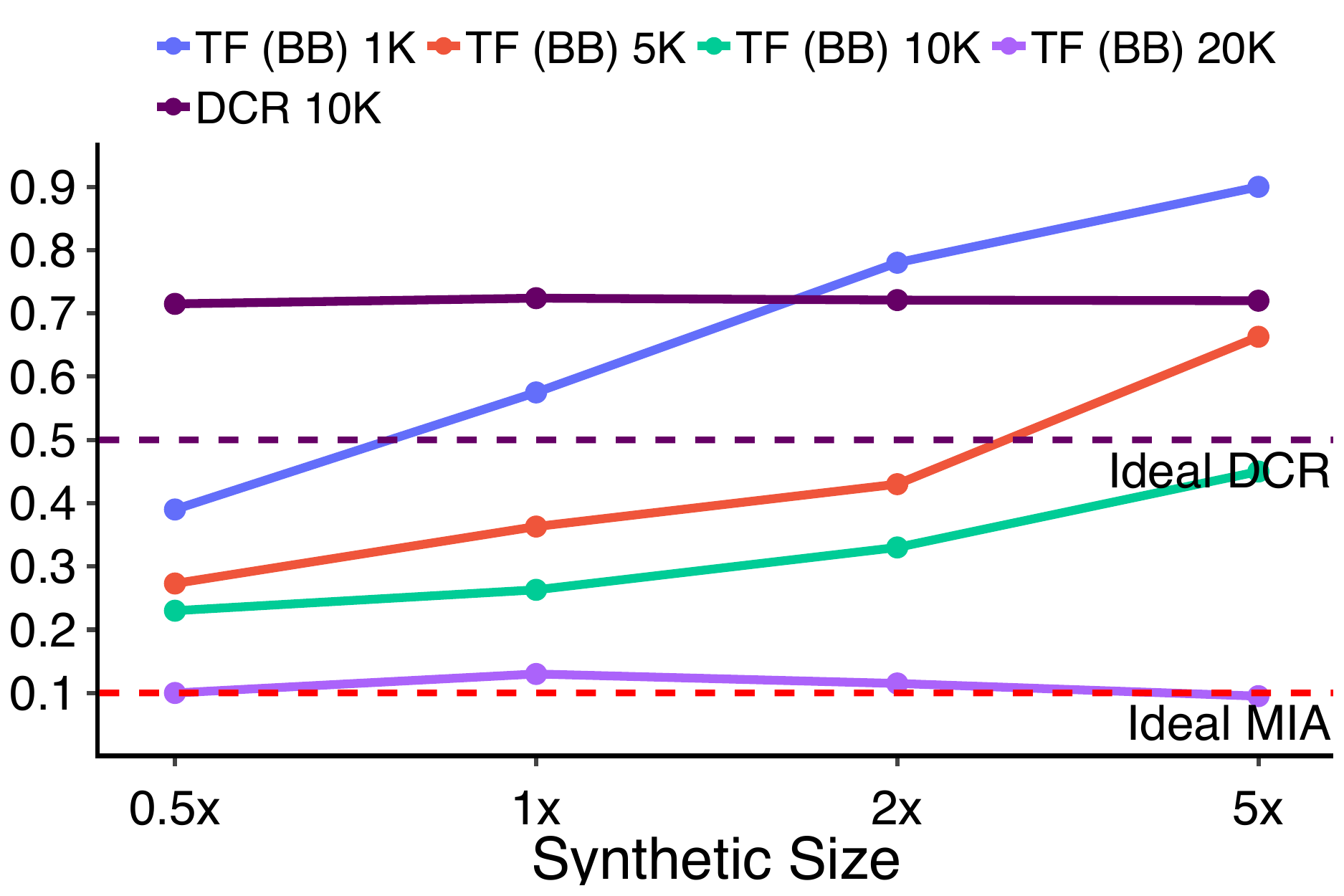}
\end{subfigure}
\begin{subfigure}{0.3175\textwidth}
\includegraphics[width=0.99\linewidth]{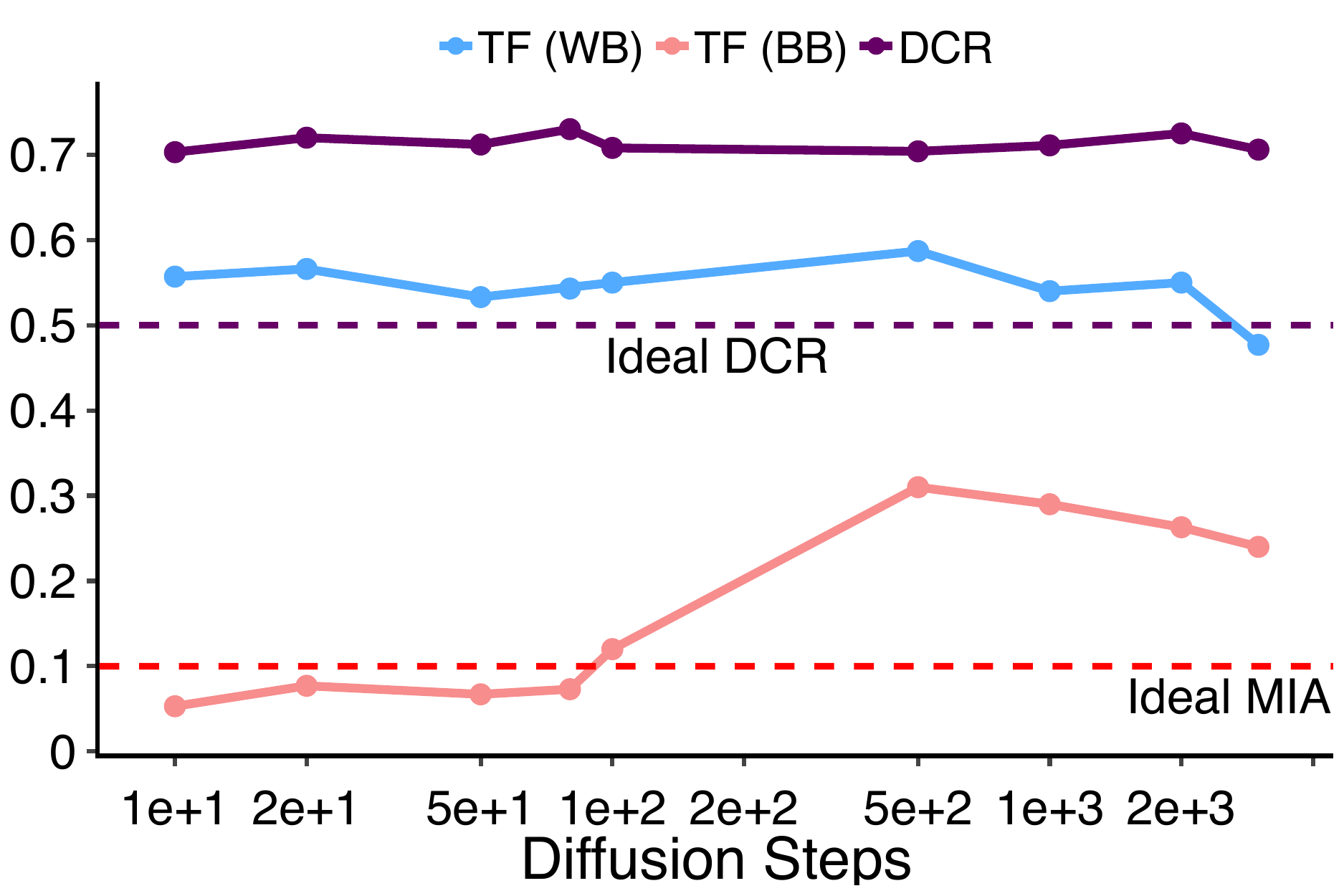}
\end{subfigure}
\begin{subfigure}{0.3175\textwidth}
\includegraphics[width=0.99\linewidth]{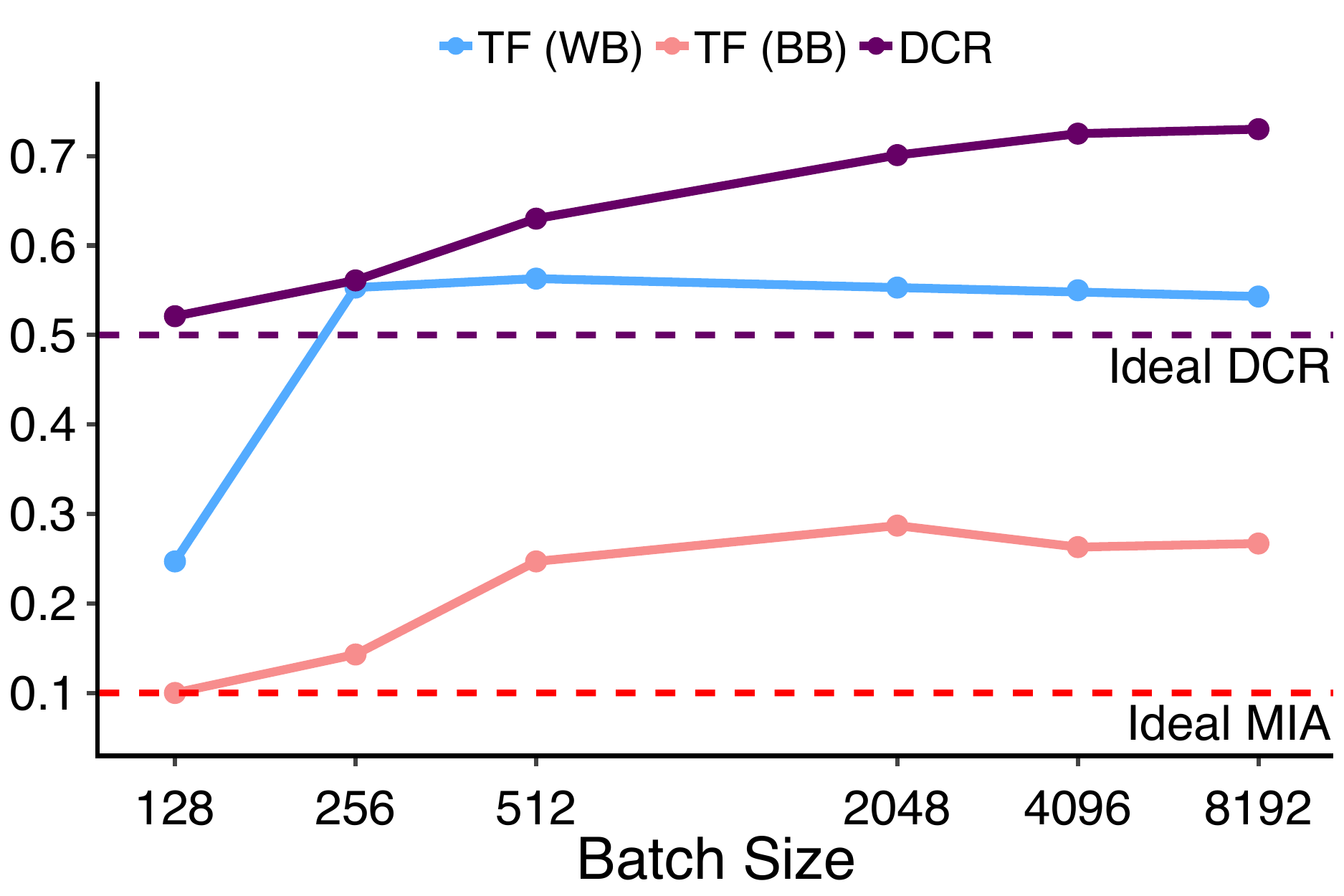}
\end{subfigure}
\begin{subfigure}{0.3175\textwidth}
\includegraphics[width=0.99\linewidth]{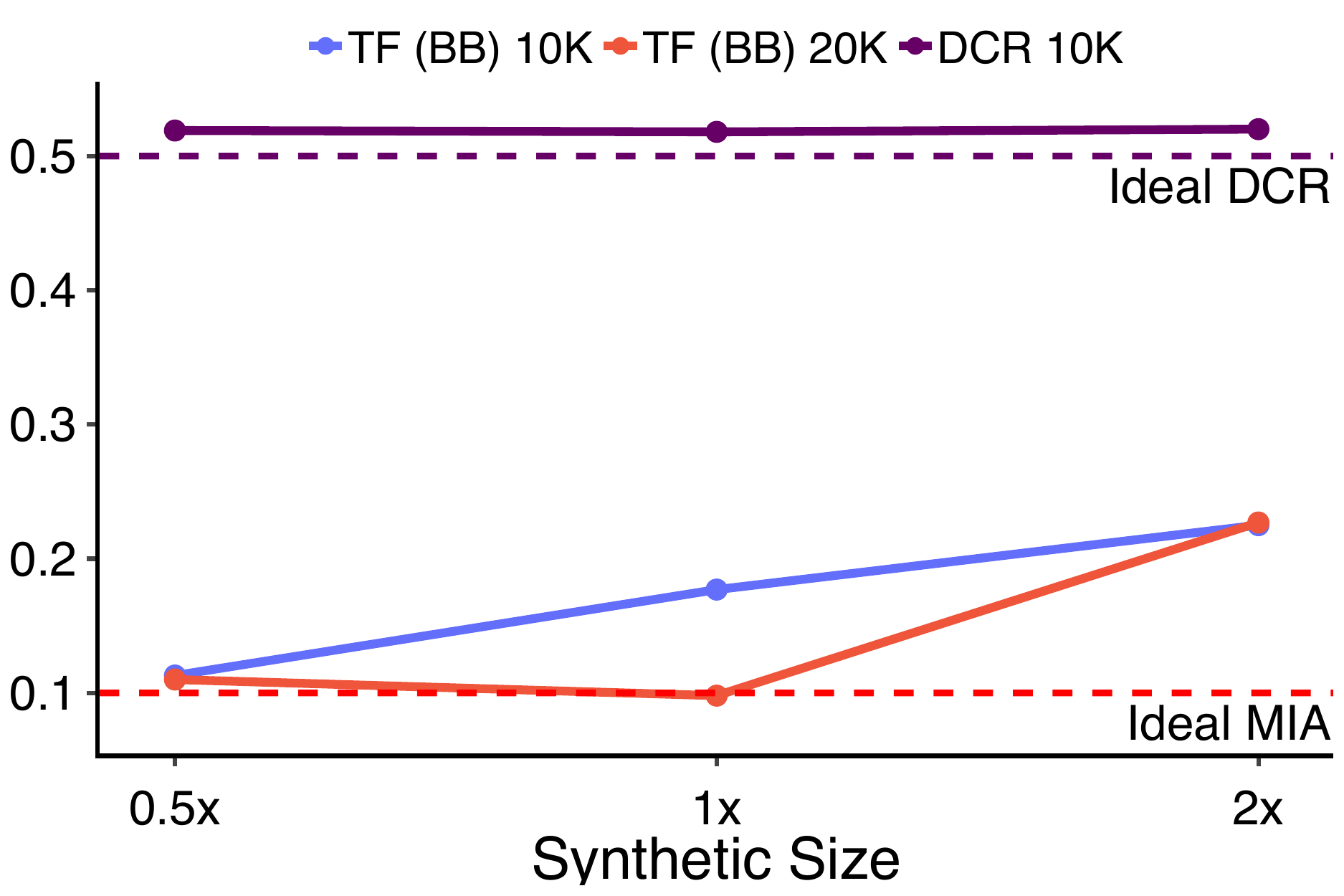}
\end{subfigure}
\begin{subfigure}{0.3175\textwidth}
\includegraphics[width=0.99\linewidth]{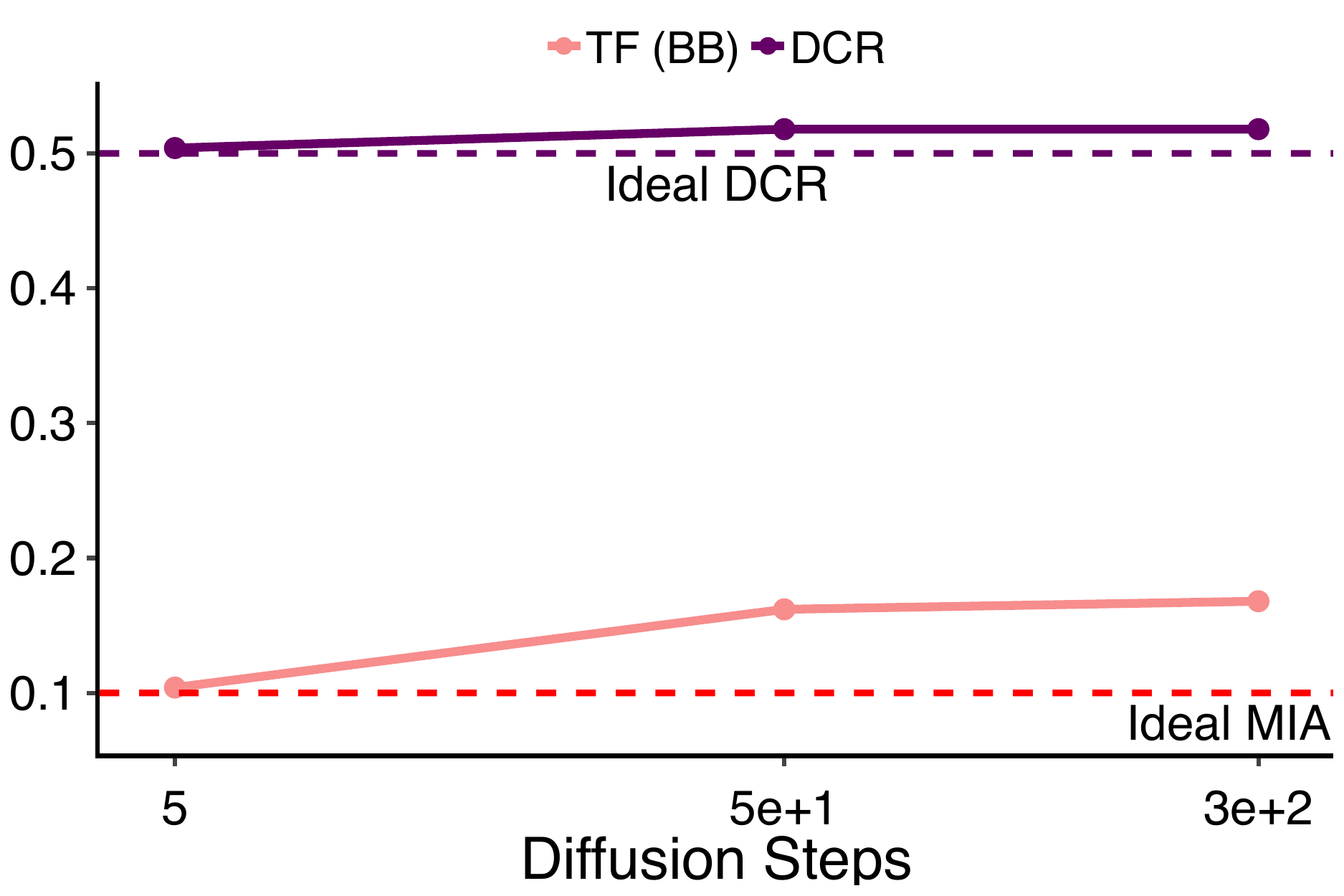}
\end{subfigure}
\begin{subfigure}{0.3175\textwidth}
\includegraphics[width=0.99\linewidth]{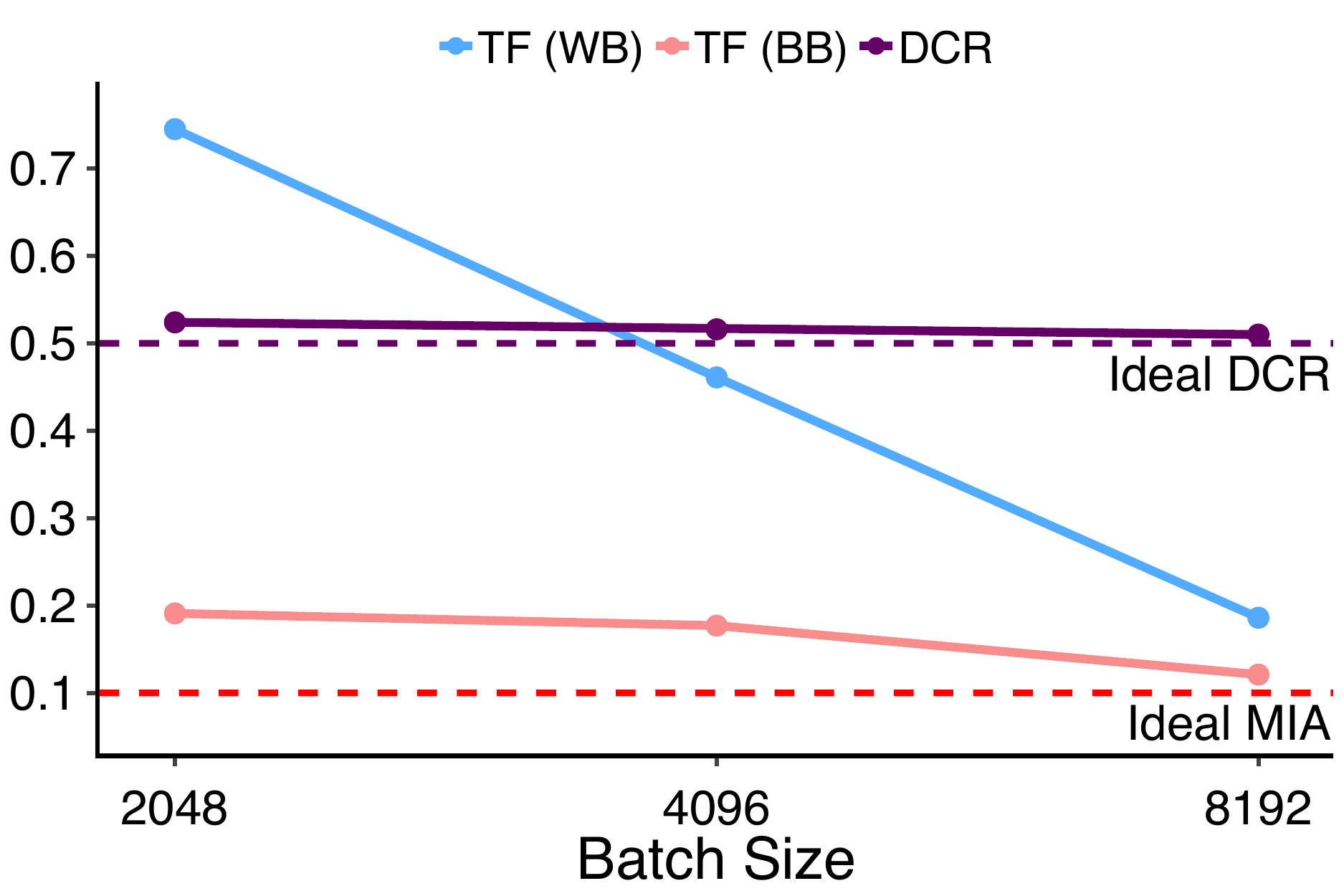}
\end{subfigure}
\caption{Variations in synthesis ratios, batch size, and  diffusion steps vs. TF MIA success and DCR in WB and BB settings for Berka (Rows 1, 2) and Diabetes (Rows 3, 4) across ClavaDDPM (Rows 1, 3) and TabDiff (Rows 2, 4) target models.} \label{fig:tf_synthesis_variations}
\end{figure}

\subsection{Synthetic Ratio, Diffusion Steps, and Batch Size} \label{diffusion_batch_ratio}

Experiments in this section vary the number of diffusion timesteps, training batch size, and ratios of synthetic data to training data for both ClavaDDPM and TabDiff. Figure \ref{fig:tf_synthesis_variations} (left) shows privacy leakage as the ratio of generated synthetic data to training data rises for the BB TF attack. In all cases, MIA success is strongly tied to this ratio, but the scaling also depends on the training set size and TDM. For example, in smaller data regimes, ClavaDDPM exhibits a drastic increase in privacy risk when generating $10\times$ more synthetic data than the size of its training set, whereas models trained on larger datasets do not experience an increase of comparable magnitude. This relationship is less clear for TabDiff, but generating more synthetic data uniformly increases leakage, regardless of training set size. In this setting, DCR remains deficient in capturing the marked increases in privacy leakage when synthesizing increasingly large quantities of data. This is likely because the models' learned data-space distribution for generating synthetic samples is static, regardless of how many points either model is asked to produce after training, hampering DCR sensitivity.

As seen in Figure \ref{fig:tf_synthesis_variations} (center), increasing the number of diffusion timesteps also typically increases leakage. Note that, as a continuous-time diffusion model, diffusion step discretization only occurs at inference time for TabDiff and, thus, only alters BB attack success. The effect is more pronounced for ClavaDDPM, likely due to it influencing both training dynamics and data generation. However, using more steps does moderately elevate leakage for TabDiff. For both datasets and TDMs, DCR is misleading, though it is worse for ClavaDDPM. For example, with Berka and ClavaDDPM, it does rise for the initial set of values, but then drops, suggesting that any risk may have modulated, though it did not. For Diabetes and TabDiff, MIA success rises, despite DCR hovering near ideal values.

The only lever for which the consequences of adjustment significantly diverge between the two models is batch size. In Figure \ref{fig:tf_synthesis_variations} (right), MIA success climbs with increasing batch size for ClavaDDPM, intuitively matching previous work suggesting regularizing effects from small batches \citep{pmlr-v119-smith20a}. In this regime, DCR is essentially flat for Berka, missing rising privacy risk. For the Diabetes dataset, DCR does rise and appears well-calibrated to BB TF success rate. However, there is an early jump in WB success rate for this dataset, which is maintained thereafter. This step change is not reflected by DCR or any other pseudo-privacy metric in Appendix \ref{other_privacy_heuristics}. On the other hand, for TabDiff, TF attack success generally drops with increasing batch size, especially for Diabetes. Some work does suggest that well-calibrated training can elicit good generalization with larger batch sizes \citep{10817779}, but the contrast with ClavaDDPM is interesting. While the phenomenon is consistent between datasets, the mechanism behind this divergence is an important focus of future work.

\begin{figure}[ht!]
\centering
\begin{subfigure}{0.3175\textwidth}
\includegraphics[width=0.99\linewidth]{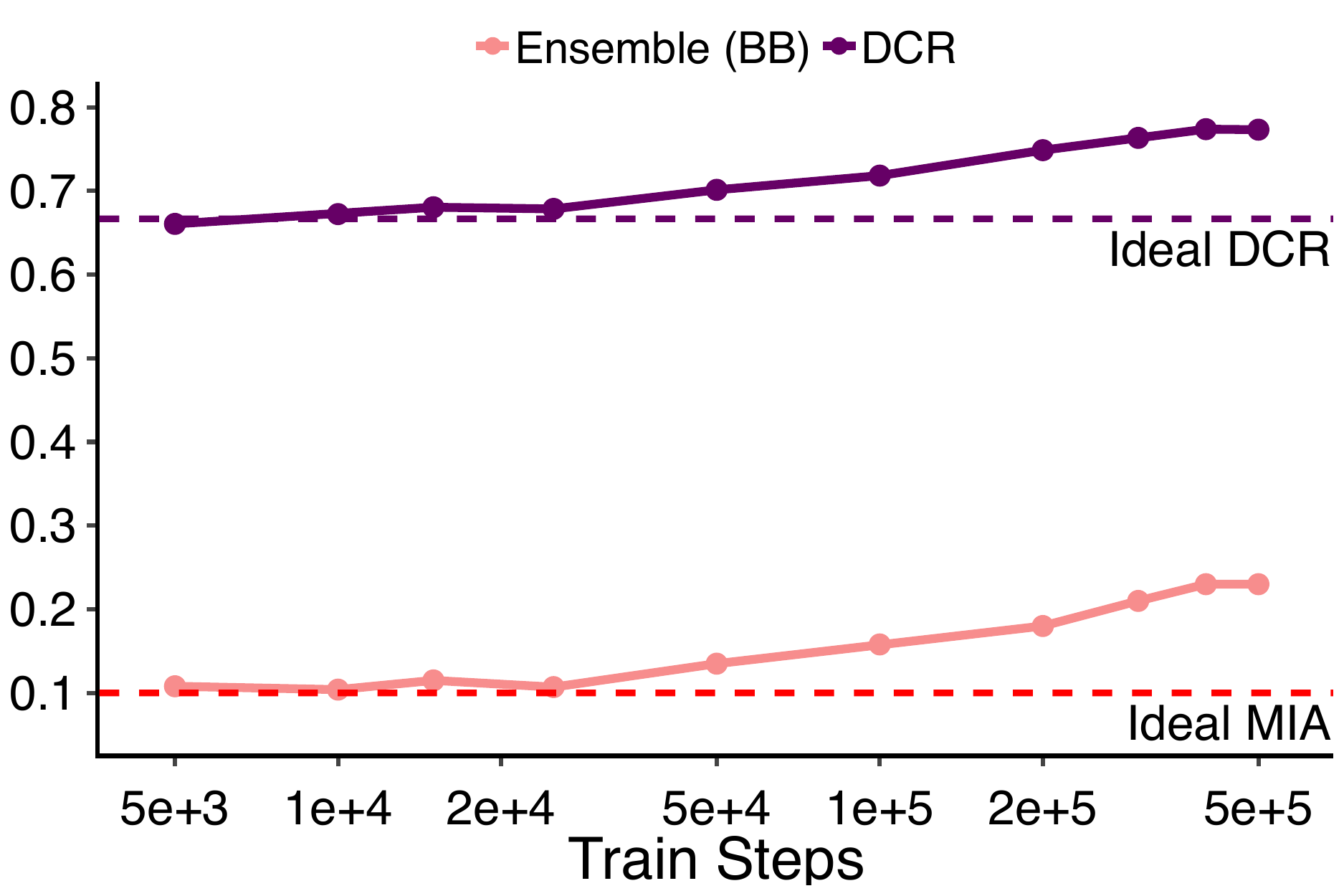}
\end{subfigure}
\begin{subfigure}{0.3175\textwidth}
\includegraphics[width=0.99\linewidth]{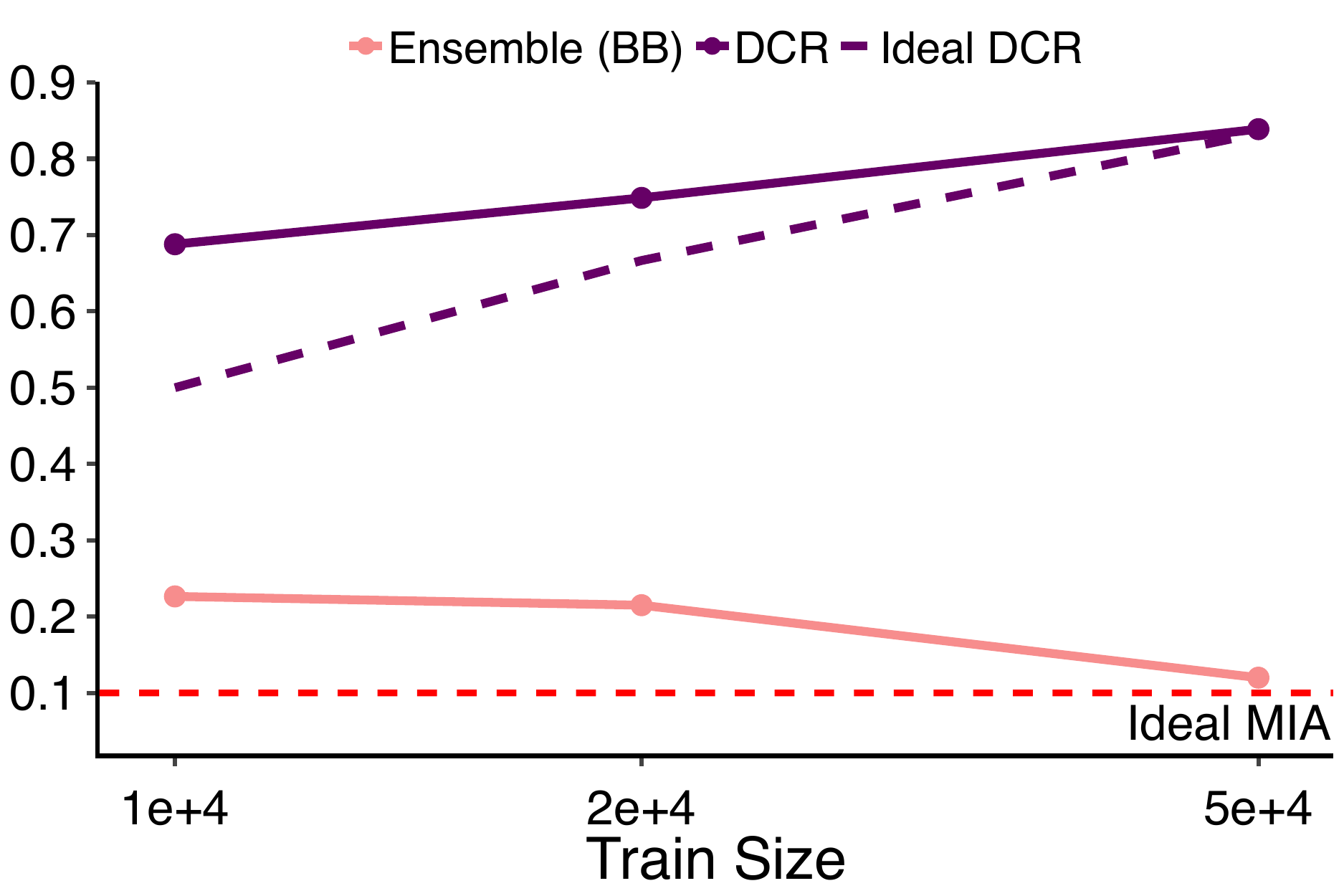}
\end{subfigure}
\begin{subfigure}{0.3175\textwidth}
\includegraphics[width=0.99\linewidth]{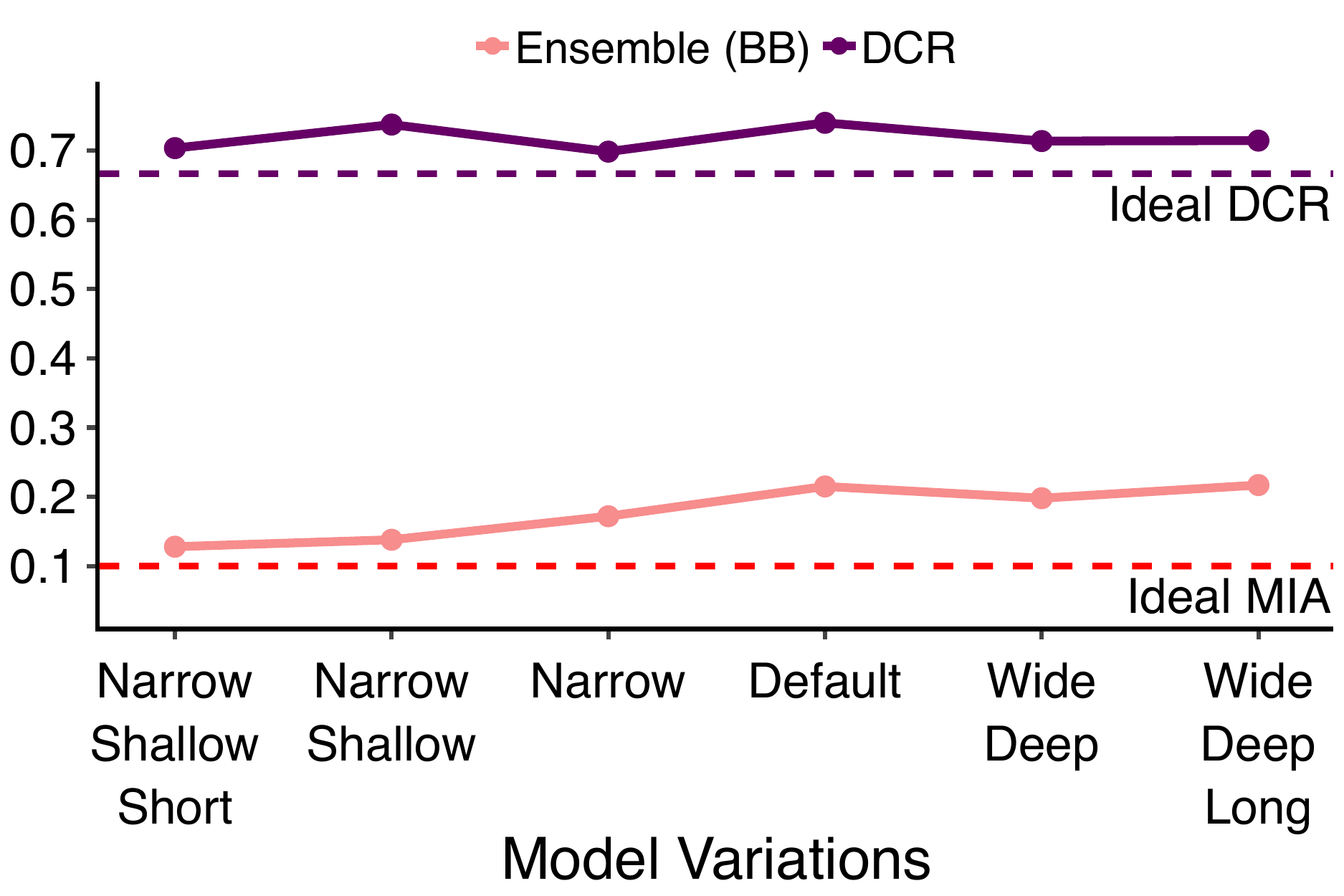}
\end{subfigure}
\begin{subfigure}{0.3175\textwidth}
\includegraphics[width=0.99\linewidth]{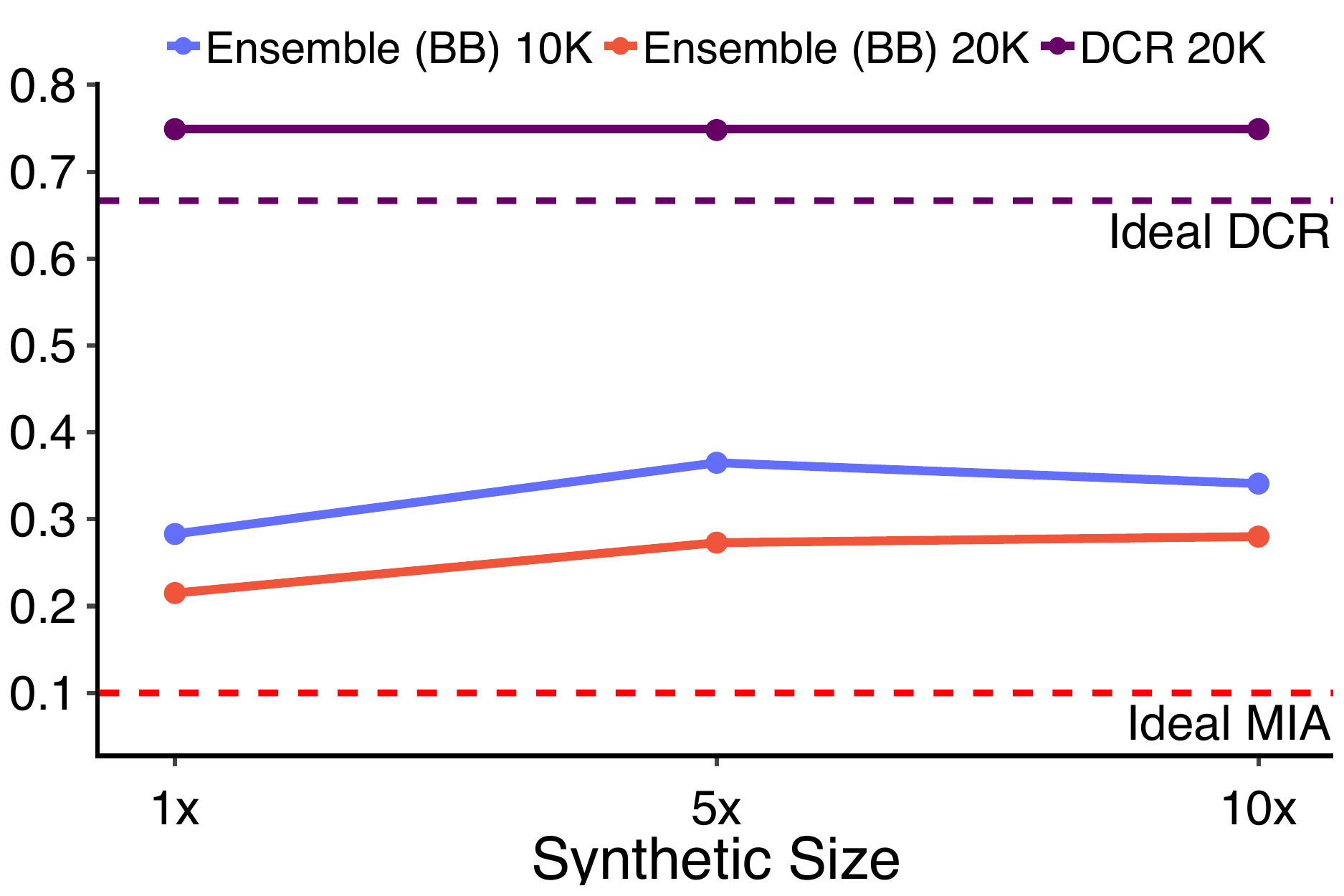}
\end{subfigure}
\begin{subfigure}{0.3175\textwidth}
\includegraphics[width=0.99\linewidth]{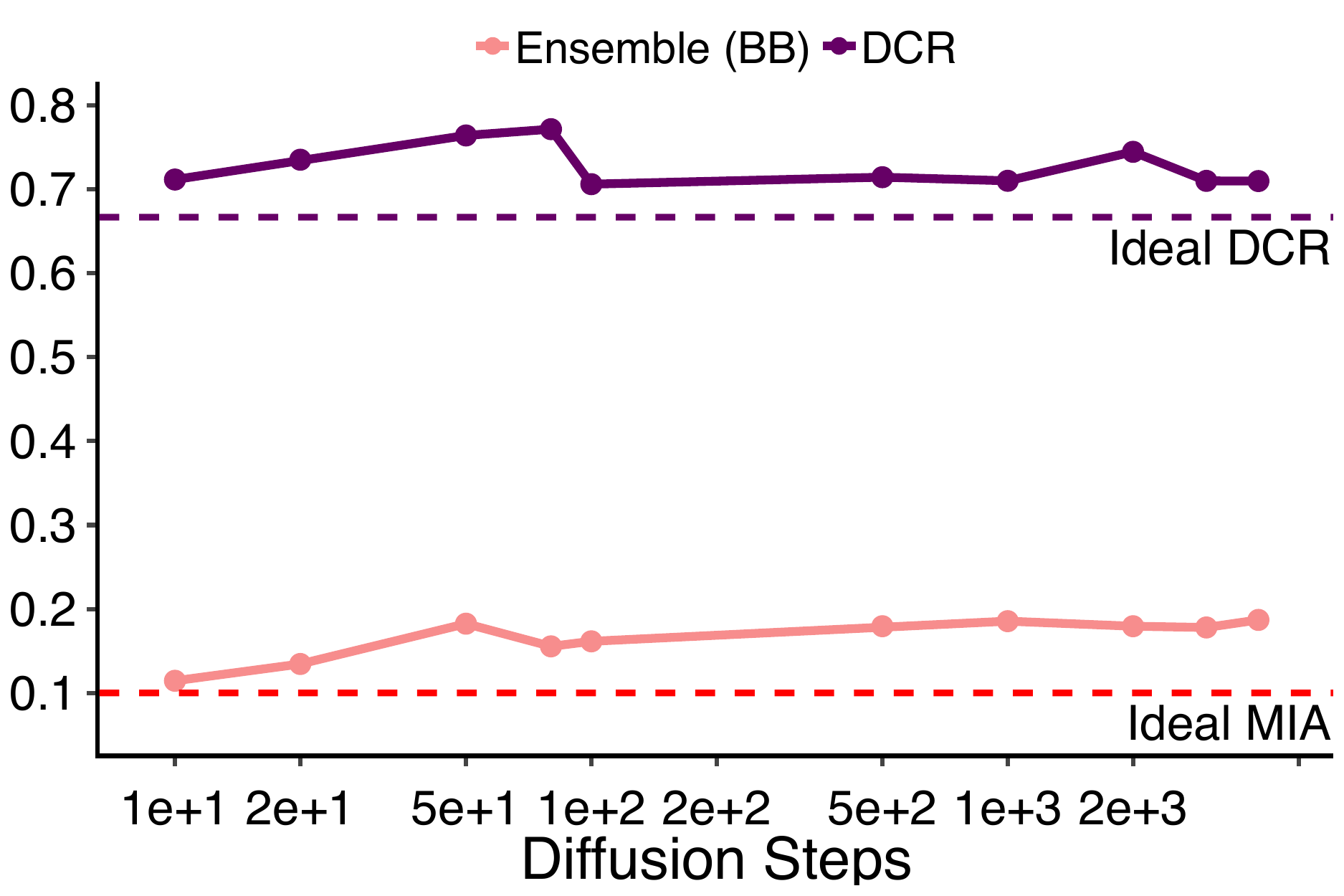}
\end{subfigure}
\begin{subfigure}{0.3175\textwidth}
\includegraphics[width=0.99\linewidth]{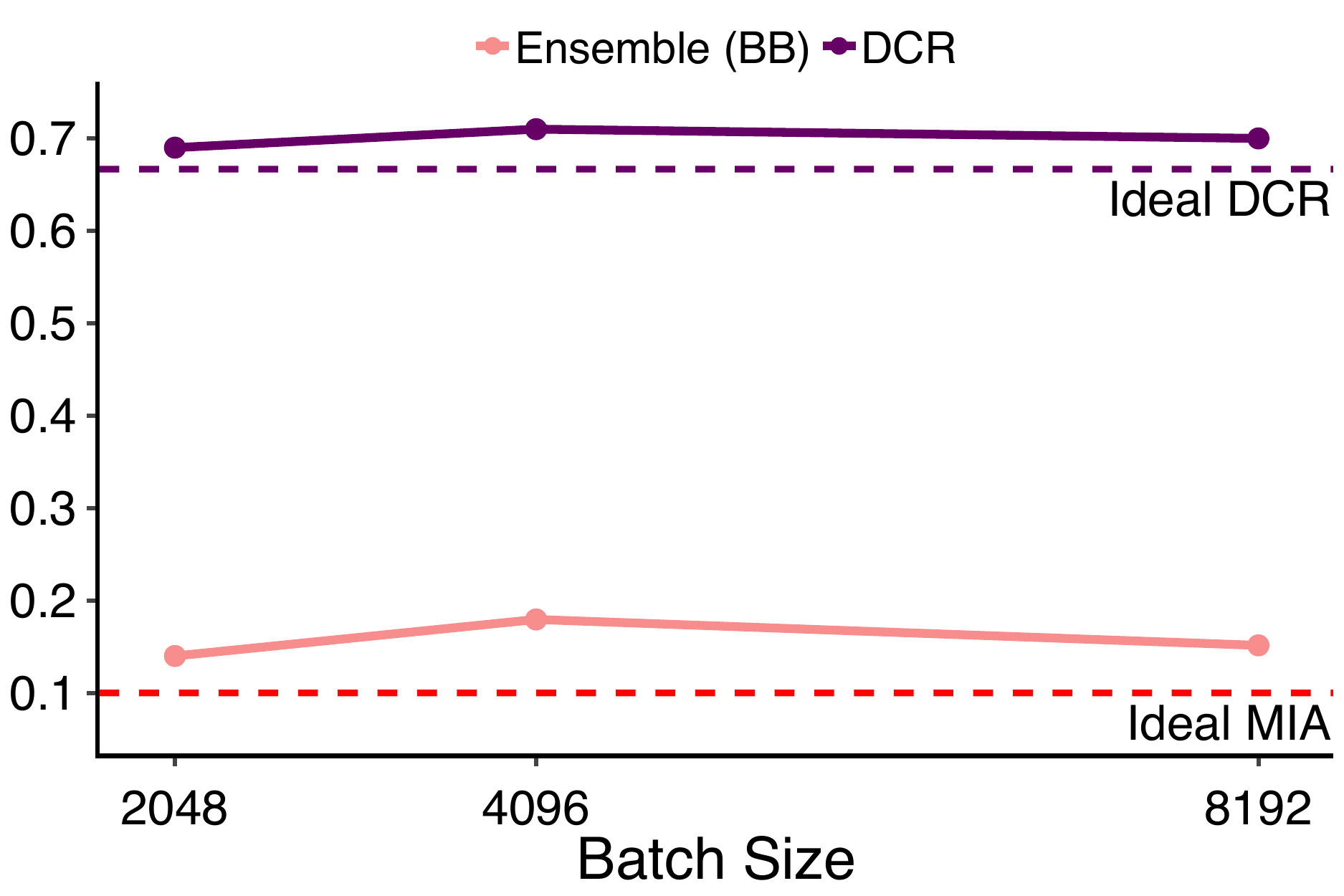}
\end{subfigure}
\begin{subfigure}{0.3175\textwidth}
\includegraphics[width=0.99\linewidth]{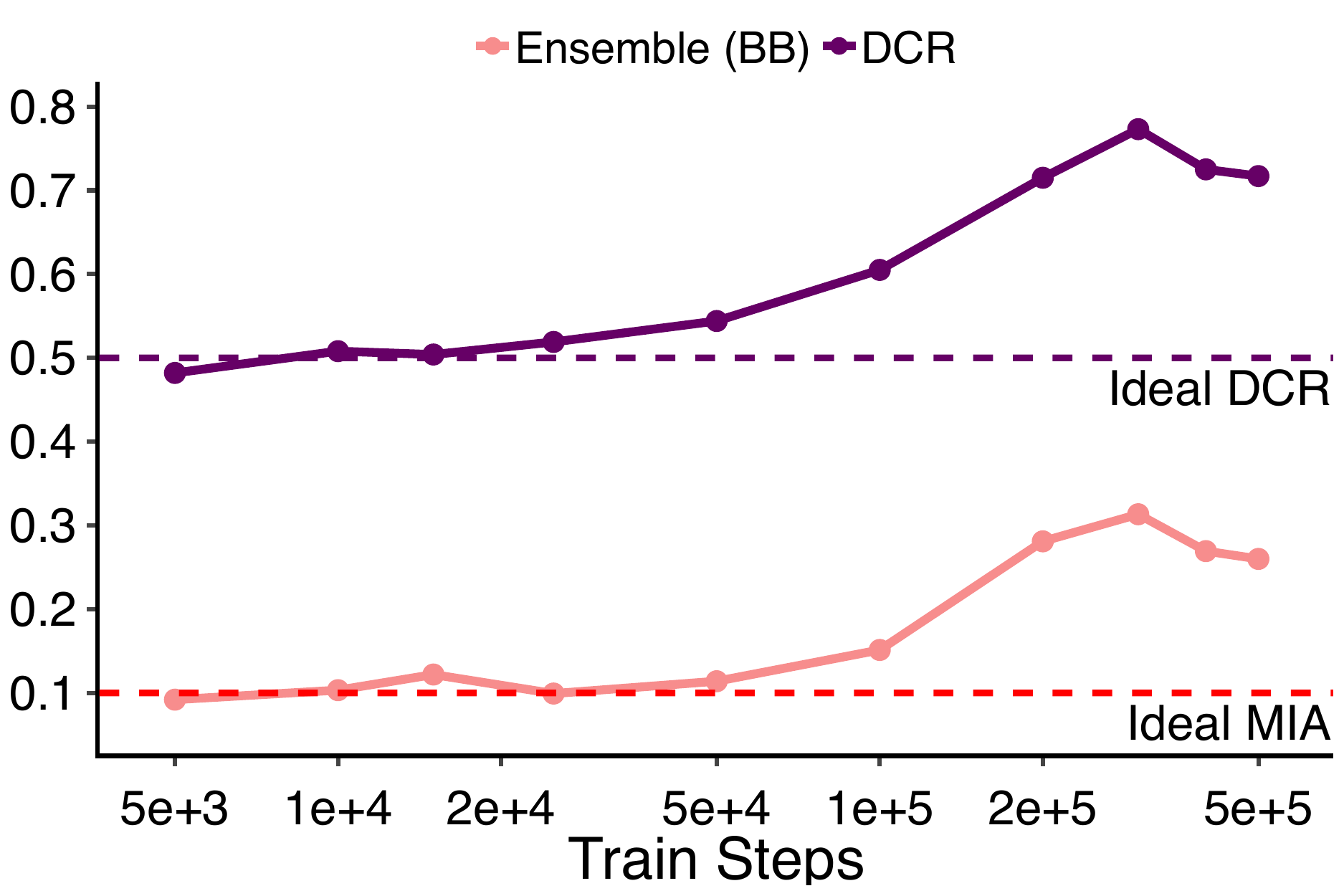}
\end{subfigure}
\begin{subfigure}{0.3175\textwidth}
\includegraphics[width=0.99\linewidth]{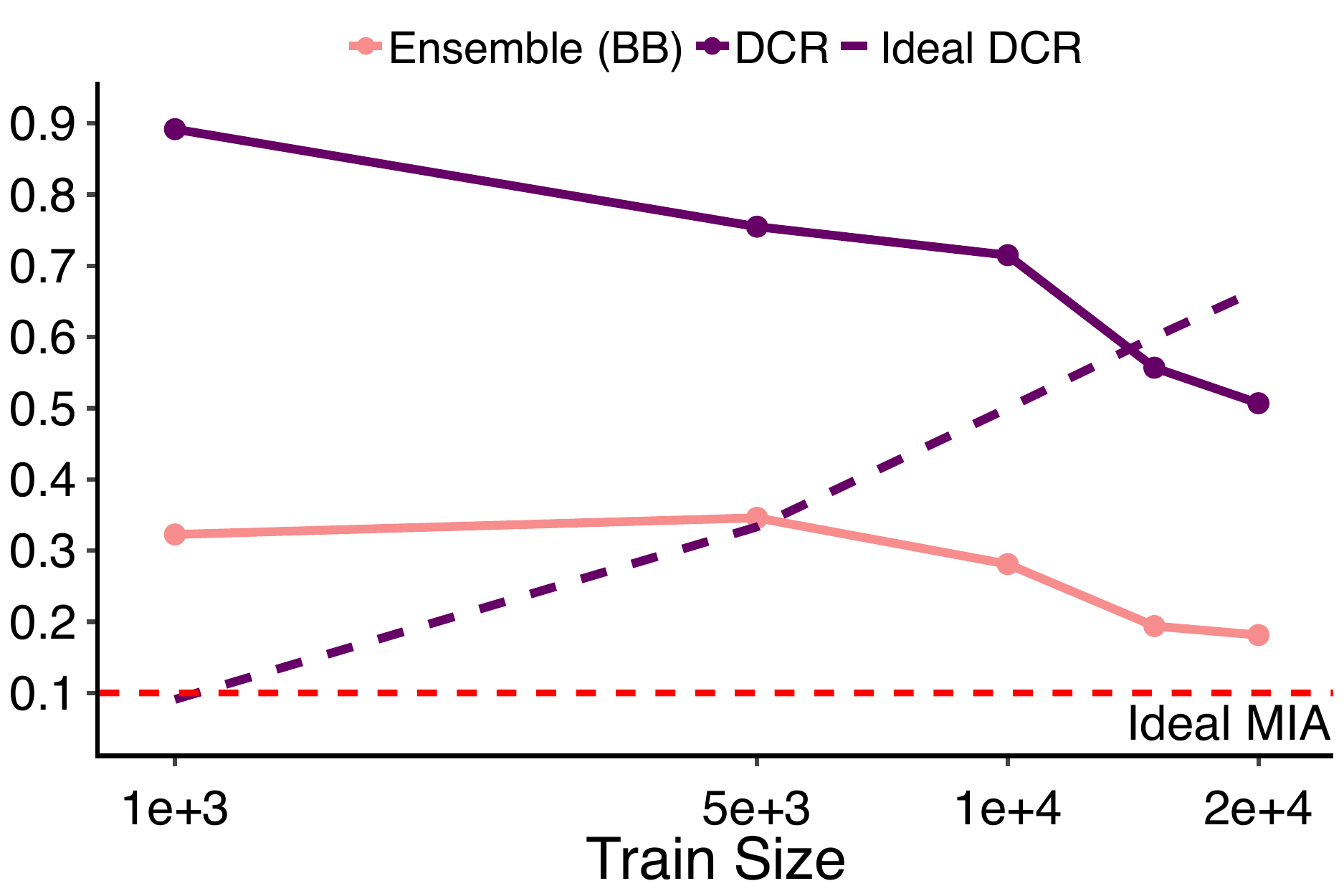}
\end{subfigure}
\begin{subfigure}{0.3175\textwidth}
\includegraphics[width=0.99\linewidth]{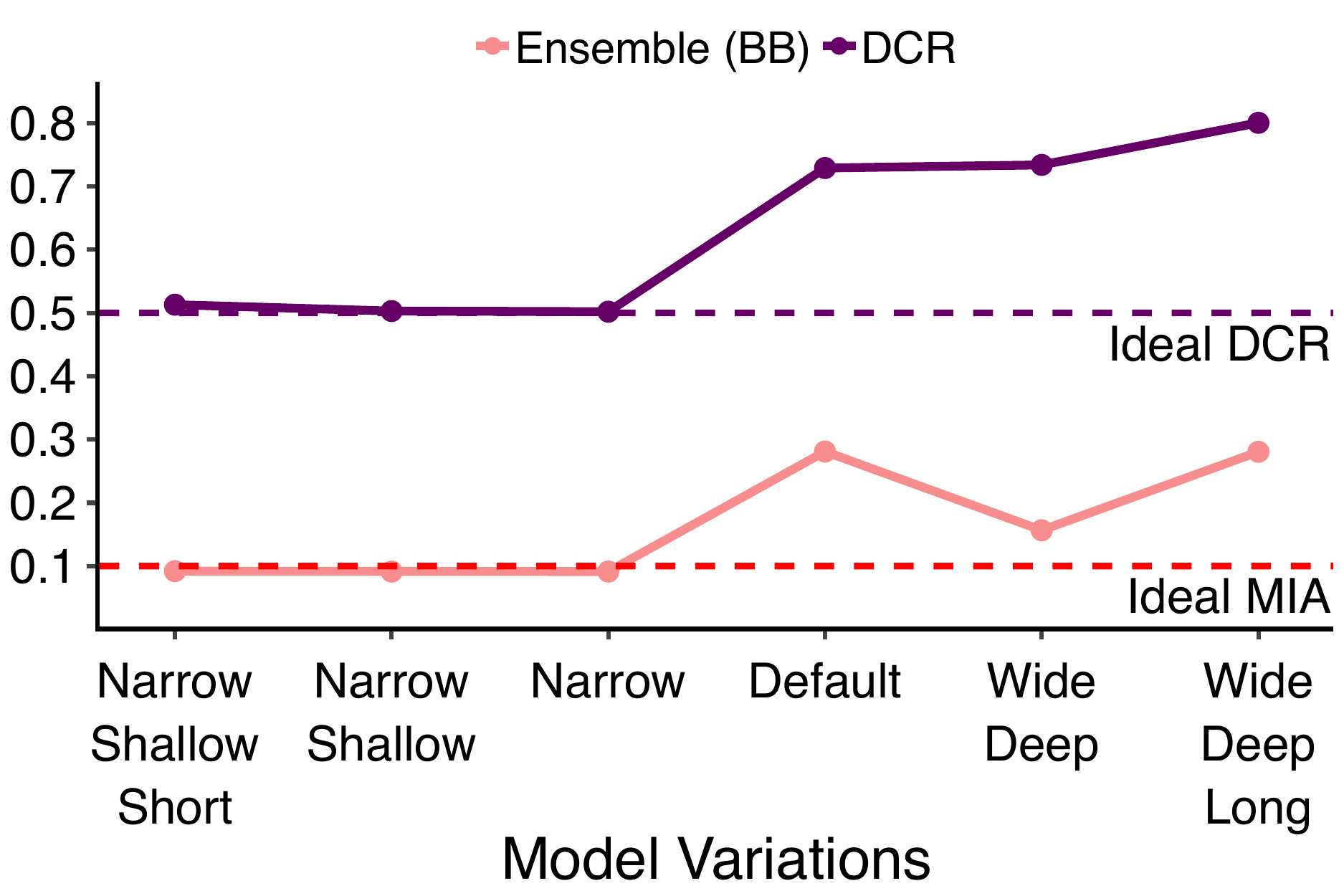}
\end{subfigure}
\begin{subfigure}{0.3175\textwidth}
\includegraphics[width=0.99\linewidth]{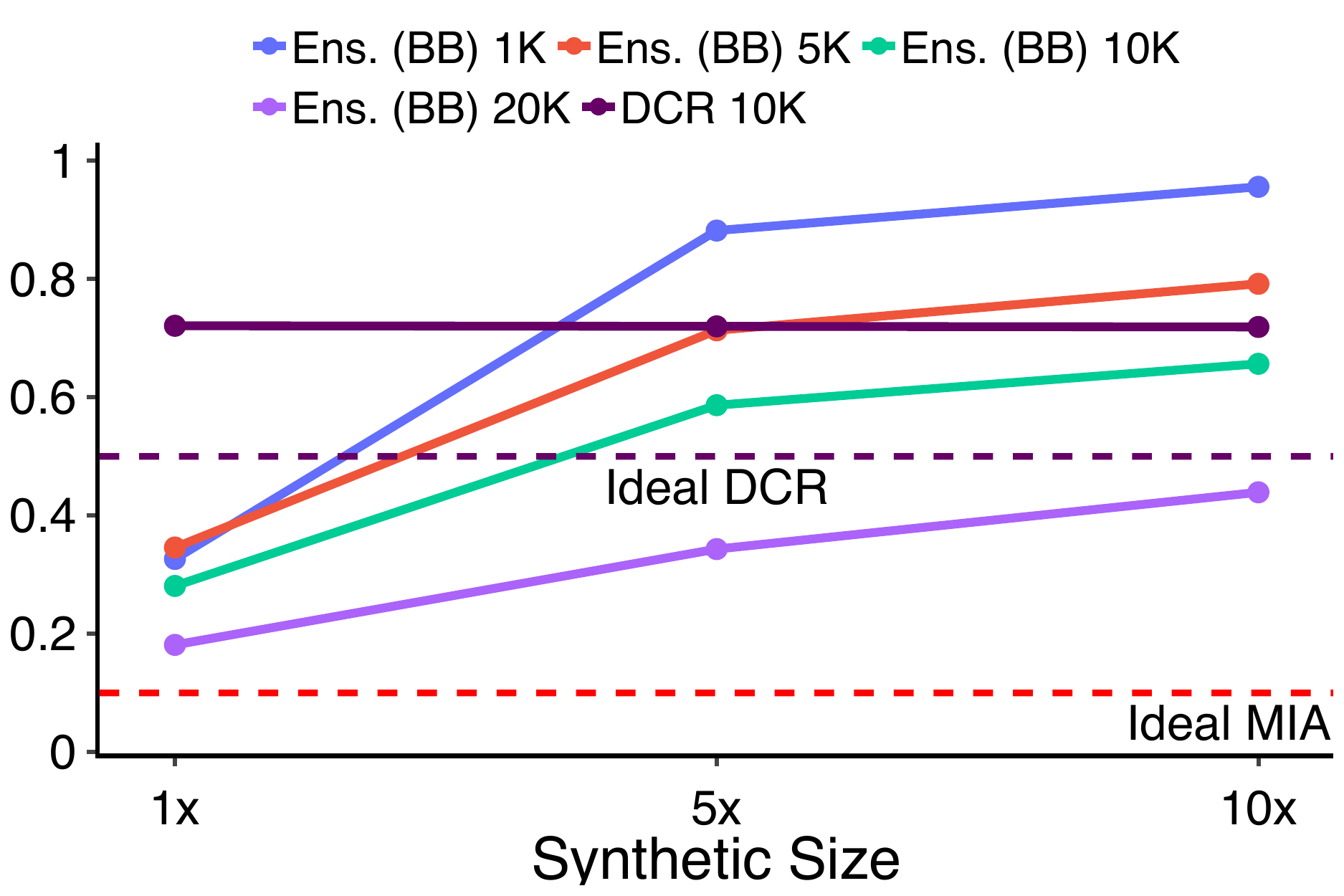}
\end{subfigure}
\begin{subfigure}{0.3175\textwidth}
\includegraphics[width=0.99\linewidth]{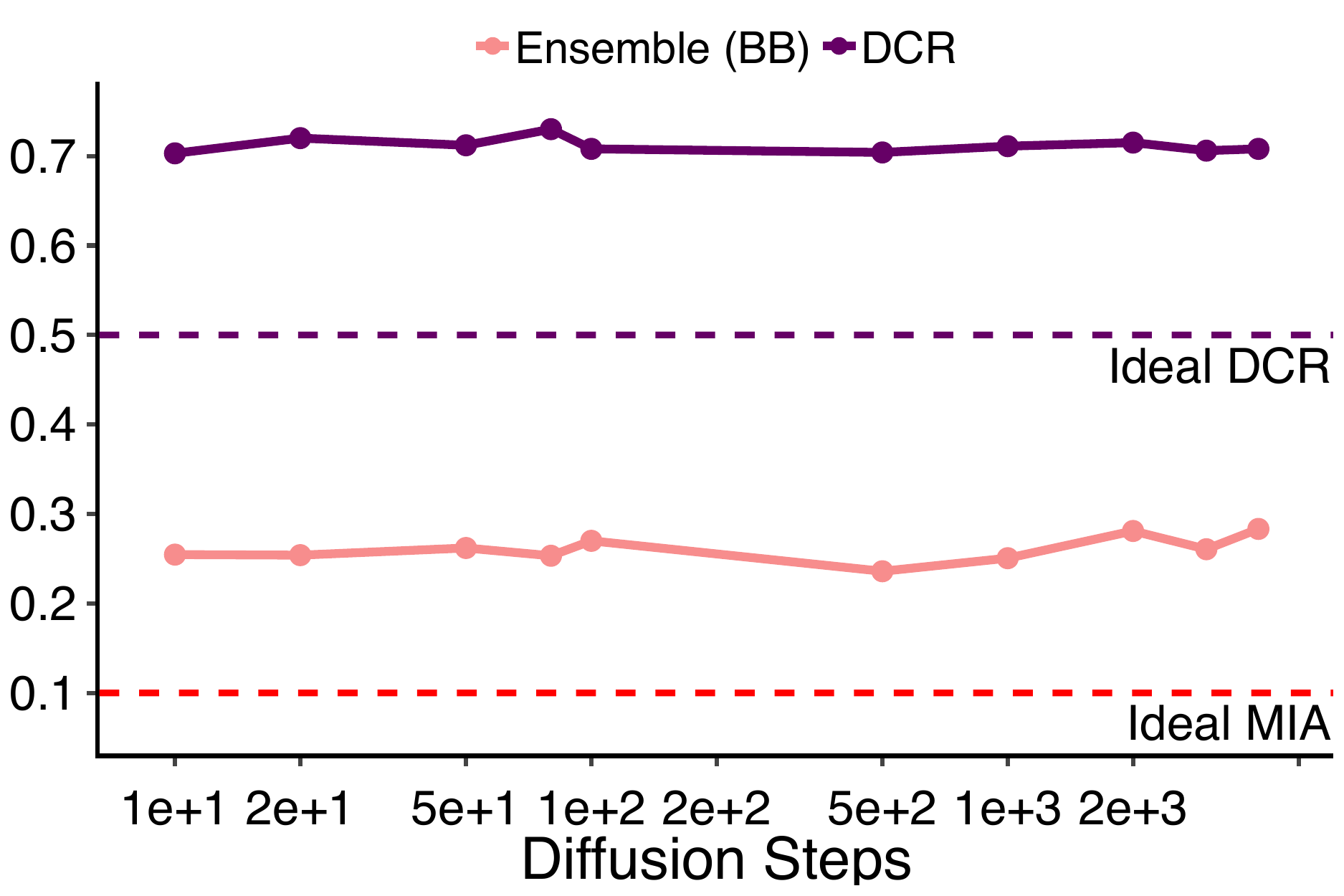}
\end{subfigure}
\begin{subfigure}{0.3175\textwidth}
\includegraphics[width=0.99\linewidth]{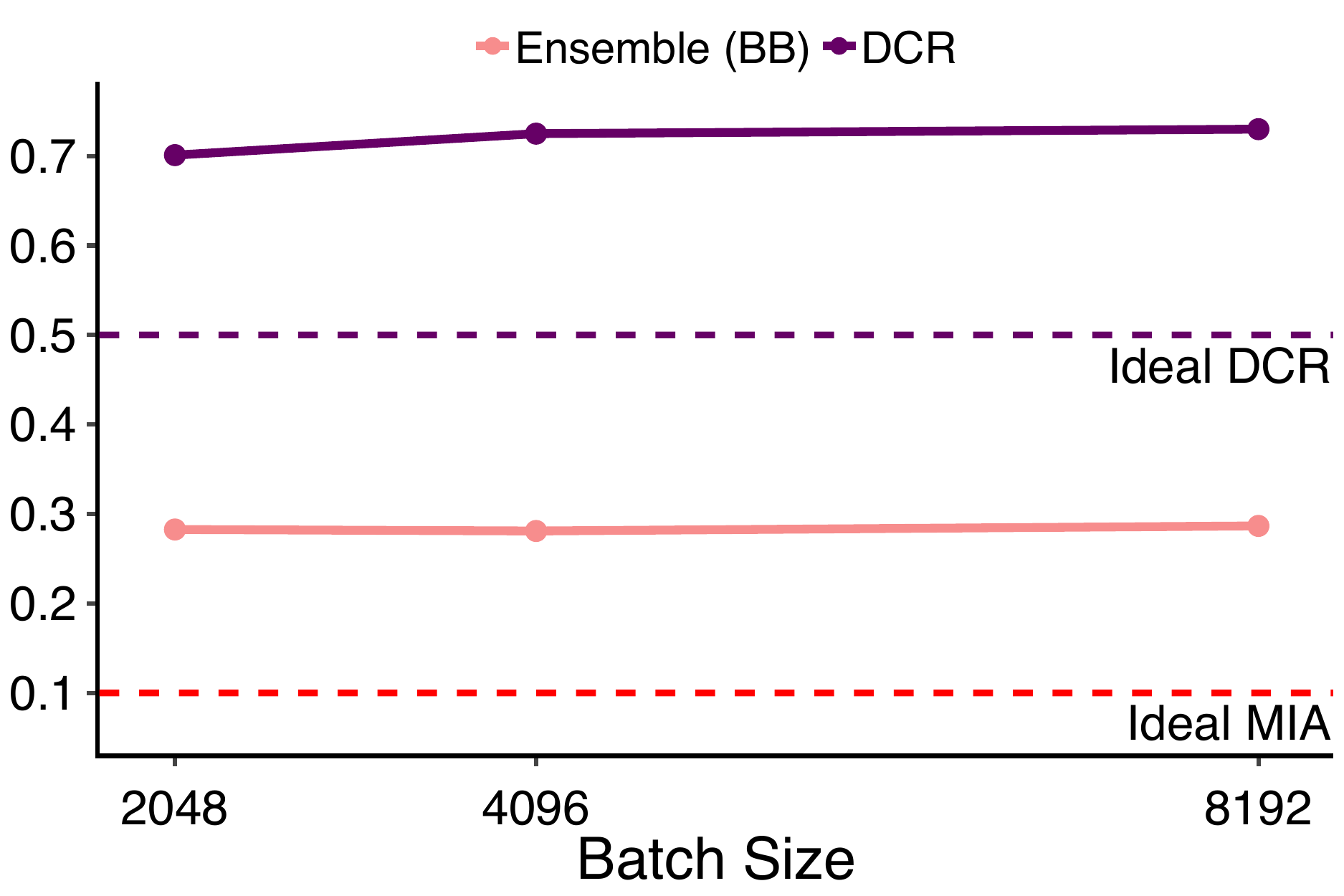}
\end{subfigure}
\caption{Variations in training steps, train size, architecture, synthesis ratios, diffusion steps, and batch size vs. Ensemble MIA success and DCR for Berka (Rows 1, 2) and Diabetes (Rows 3, 4).} \label{fig:ensemble_variations}
\end{figure}

\subsection{Generalization to Other Attacks} \label{black_box_generalization}

A natural question is whether the patterns observed in the previous sections are also visible for designs beyond the TF attack. To test this, we apply the Ensemble attack to ClavaDDPM and explore the same levers. The results in Figure \ref{fig:ensemble_variations} demonstrate that the relationship of training steps, dataset size, and model variations to privacy leakage is also present in the Ensemble attack results. Here, DCR is more correlated with MIA success. However, it still fails for the Berka model variation experiments. The synthetic-to-train dataset ratios experiments also replicate the relationships of the TF attacks and further underscore the unreliability of DCR leakage estimates in this setting. Batch size has limited impact on Ensemble attack success, though both datasets maintain non-trivial success rates. This disconnect is potentially due to batch size more strongly influencing model loss landscapes and internal representations, on which the TF attack relies, compared to output-space structures, where the Ensemble approach operates \citep{Keskar1, Granziol1}.

The growth in Ensemble MIA success with respect to changes in diffusion steps is more tempered than the TF attack, in somewhat better alignment with DCR, but does modestly rise for both datasets while DCR does not. This moderation is likely due to the mechanism of loss reconstruction in the TF attack on ClavaDDPM, which is heavily influenced by the timestep scale of the diffusion model, whereas the Ensemble method operates solely on properties of the generated data. Recall that the Ensemble technique is a BB and data-driven approach with a non-trivial reliance on distance-, density-, and canary-based measures. As such, the results demonstrate that, while DCR and other pseudo-privacy measures provides a flawed estimate of privacy risk, see Appendix \ref{other_privacy_heuristics}, more sophisticated use of distance-based metrics can produce more meaningful and reliable privacy leakage evaluations. Regardless, many of the drivers of privacy leakage are consistent between the attacks, and the results emphasize the benefit of modeling distinct attack vectors.

\subsection{Attacker Resources, Knowledge, and Data} \label{resources_knowledge_data}

\begin{figure}[ht!]
\begin{subfigure}{0.325\textwidth}
\includegraphics[width=0.99\linewidth]{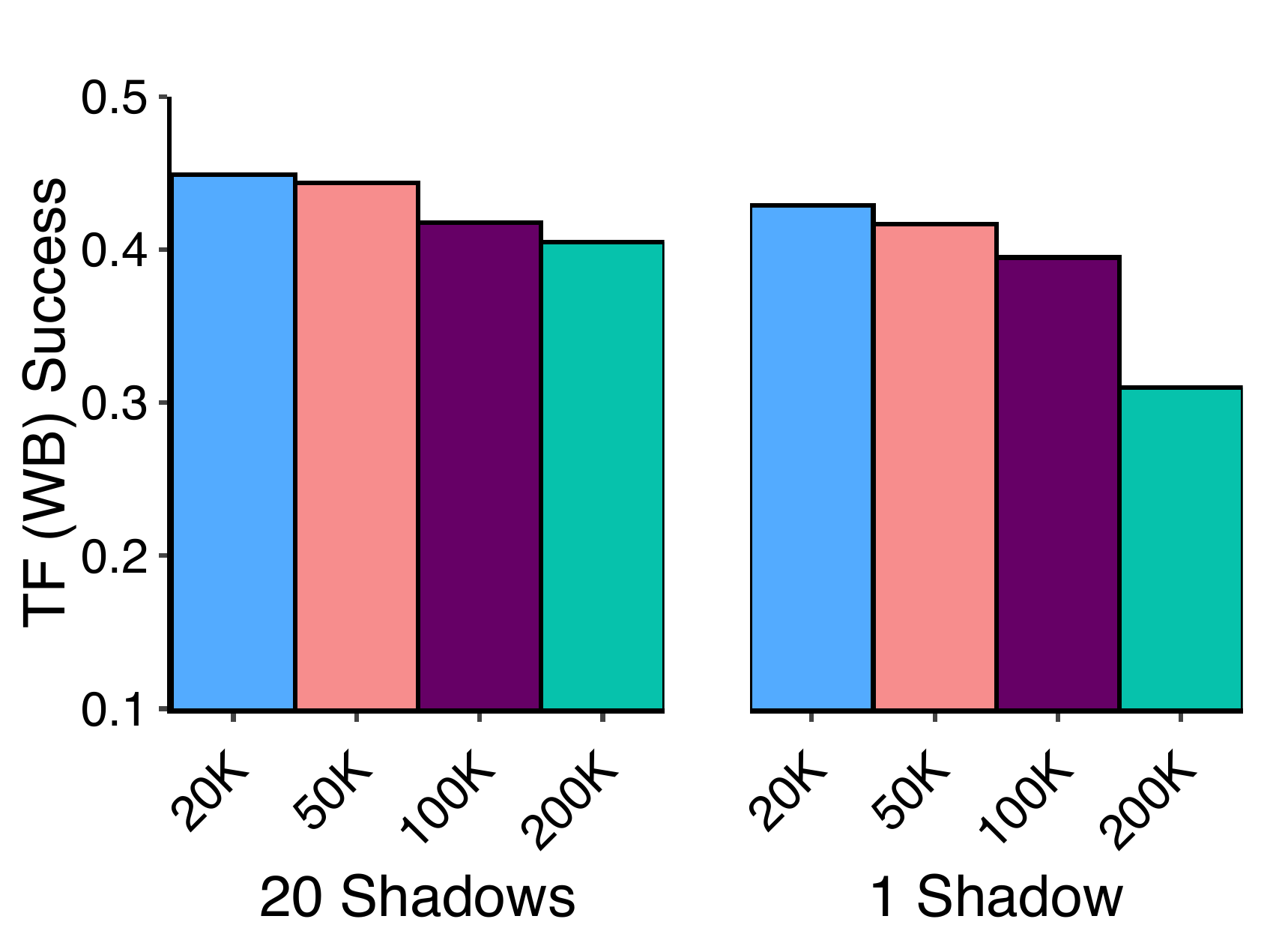}
\end{subfigure}
\raisebox{-0.5ex}{
\begin{subfigure}{0.325\textwidth}
\includegraphics[width=0.99\linewidth]{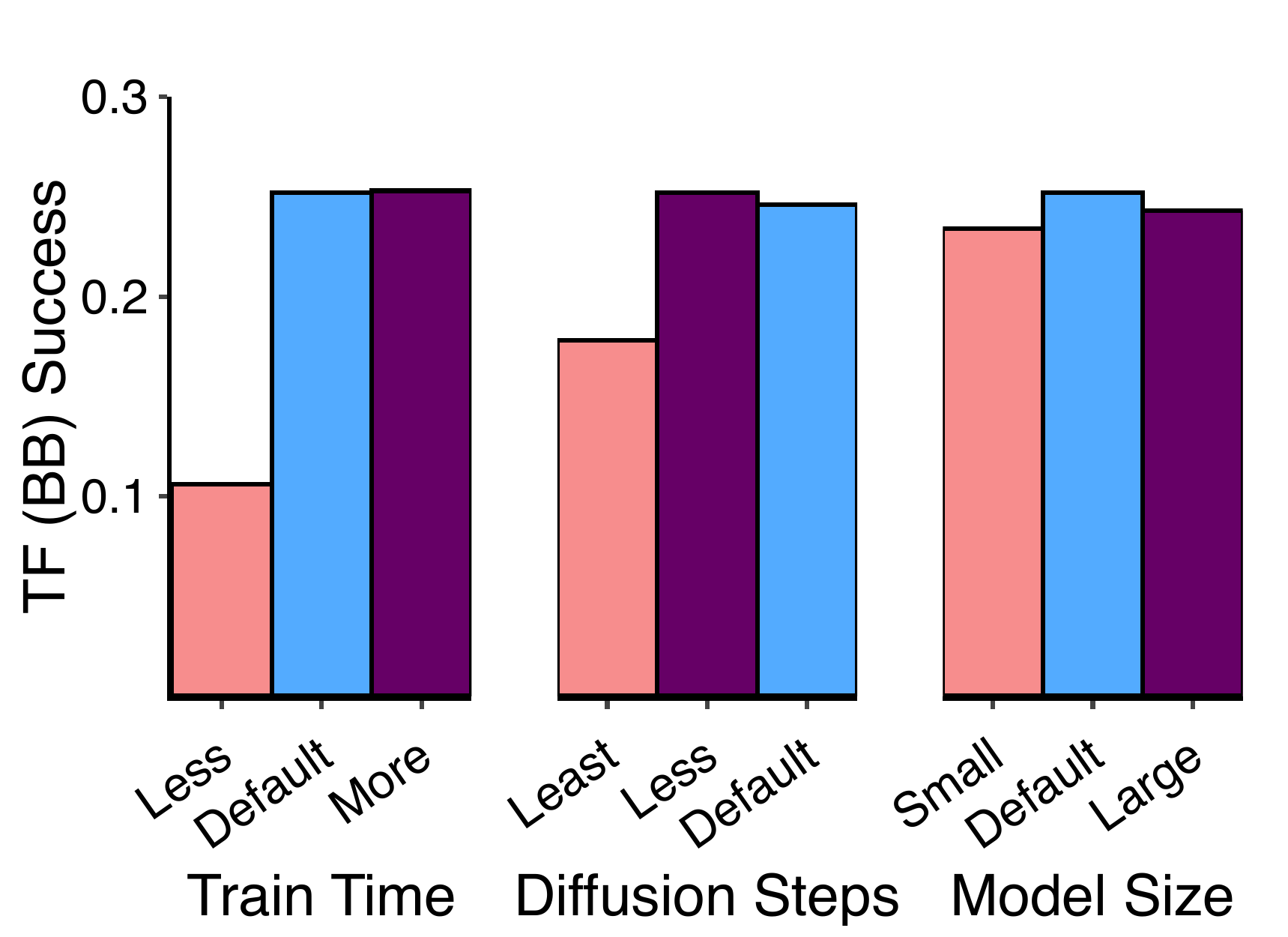}
\end{subfigure}}
\begin{subfigure}{0.325\textwidth}
\includegraphics[width=0.99\linewidth]{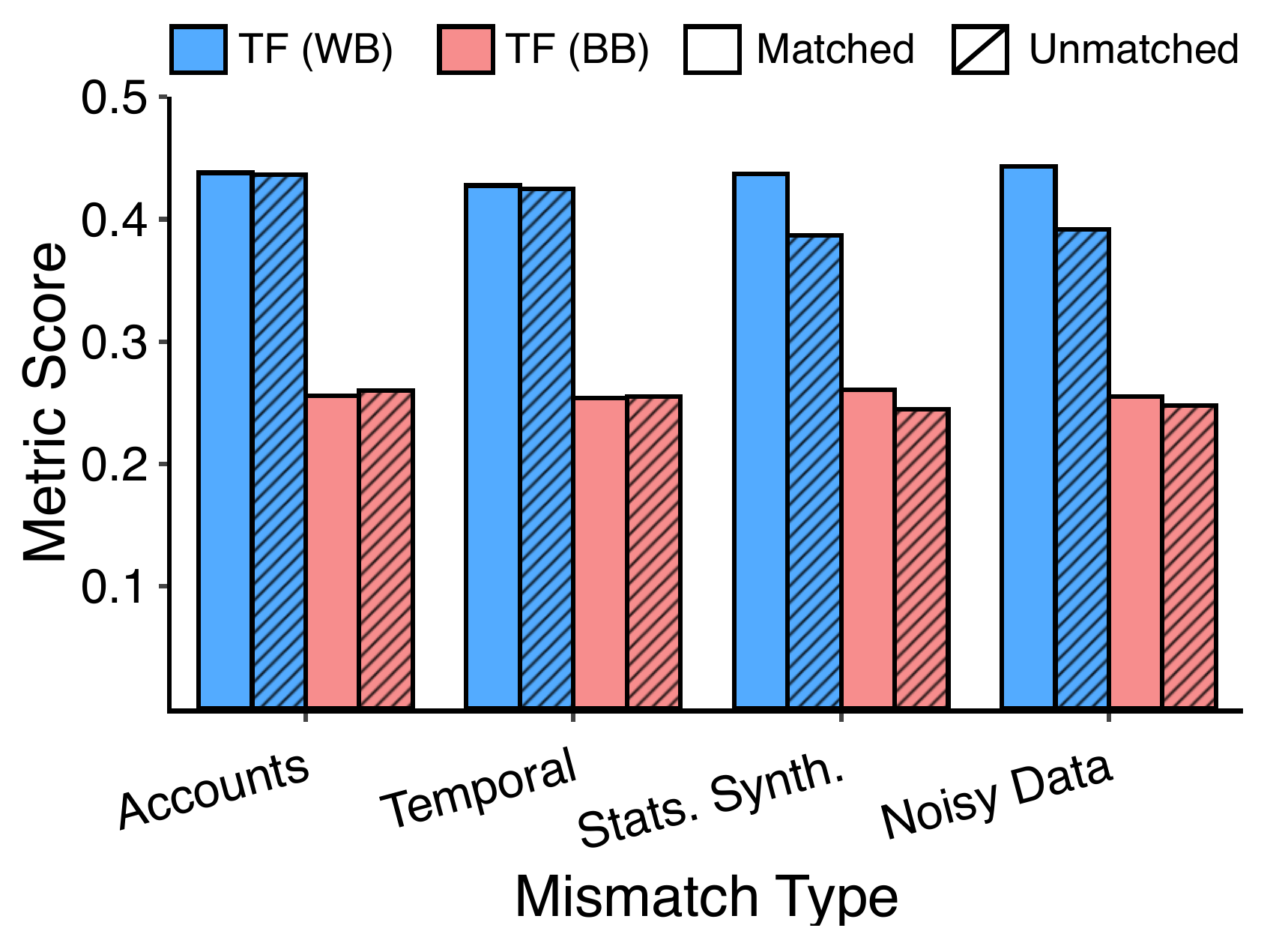}
\end{subfigure}
\begin{subfigure}{0.325\textwidth}
\includegraphics[width=0.99\linewidth]{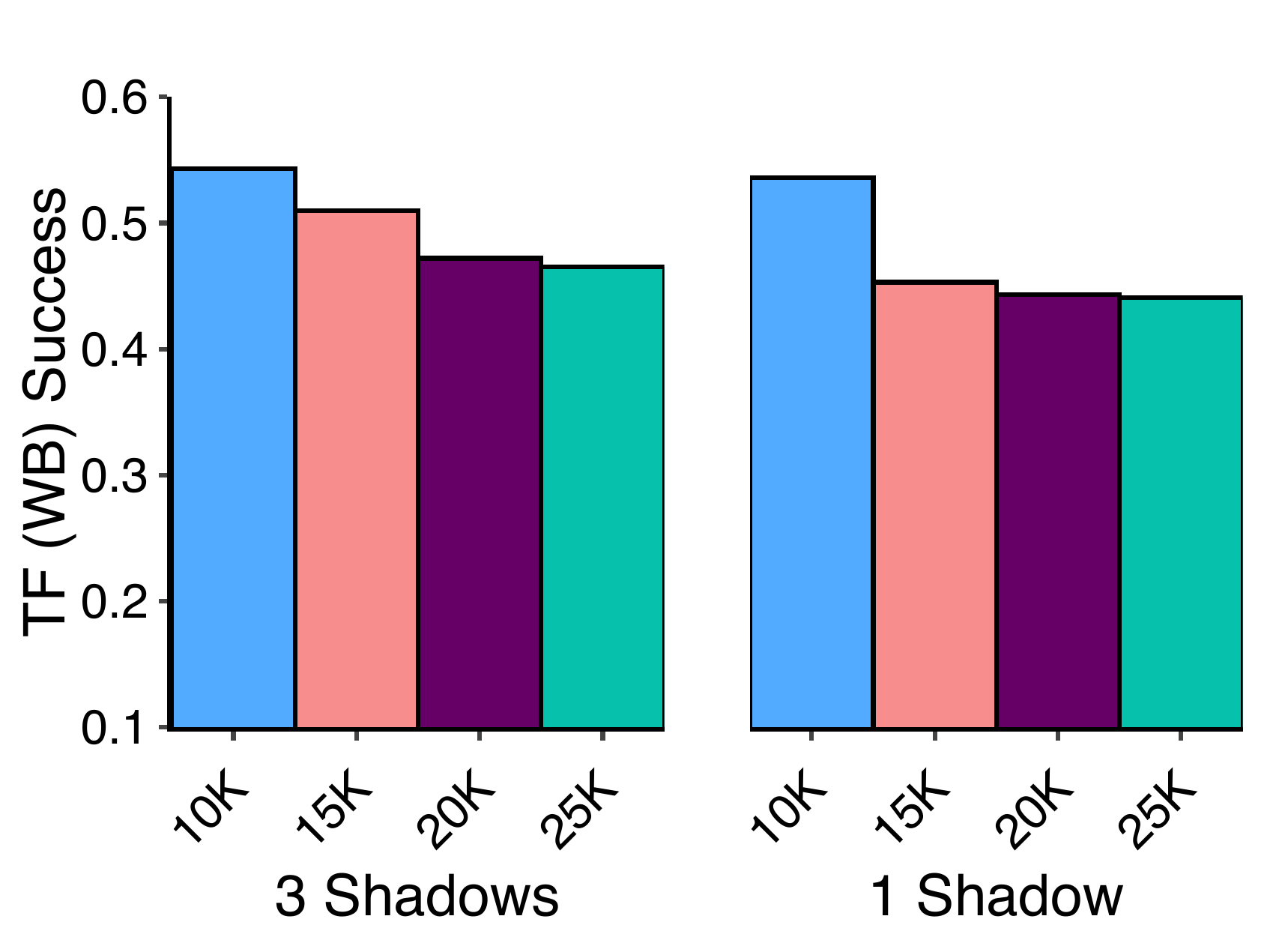}
\end{subfigure}
\raisebox{-0.5ex}{
\begin{subfigure}{0.325\textwidth}
\includegraphics[width=0.99\linewidth]{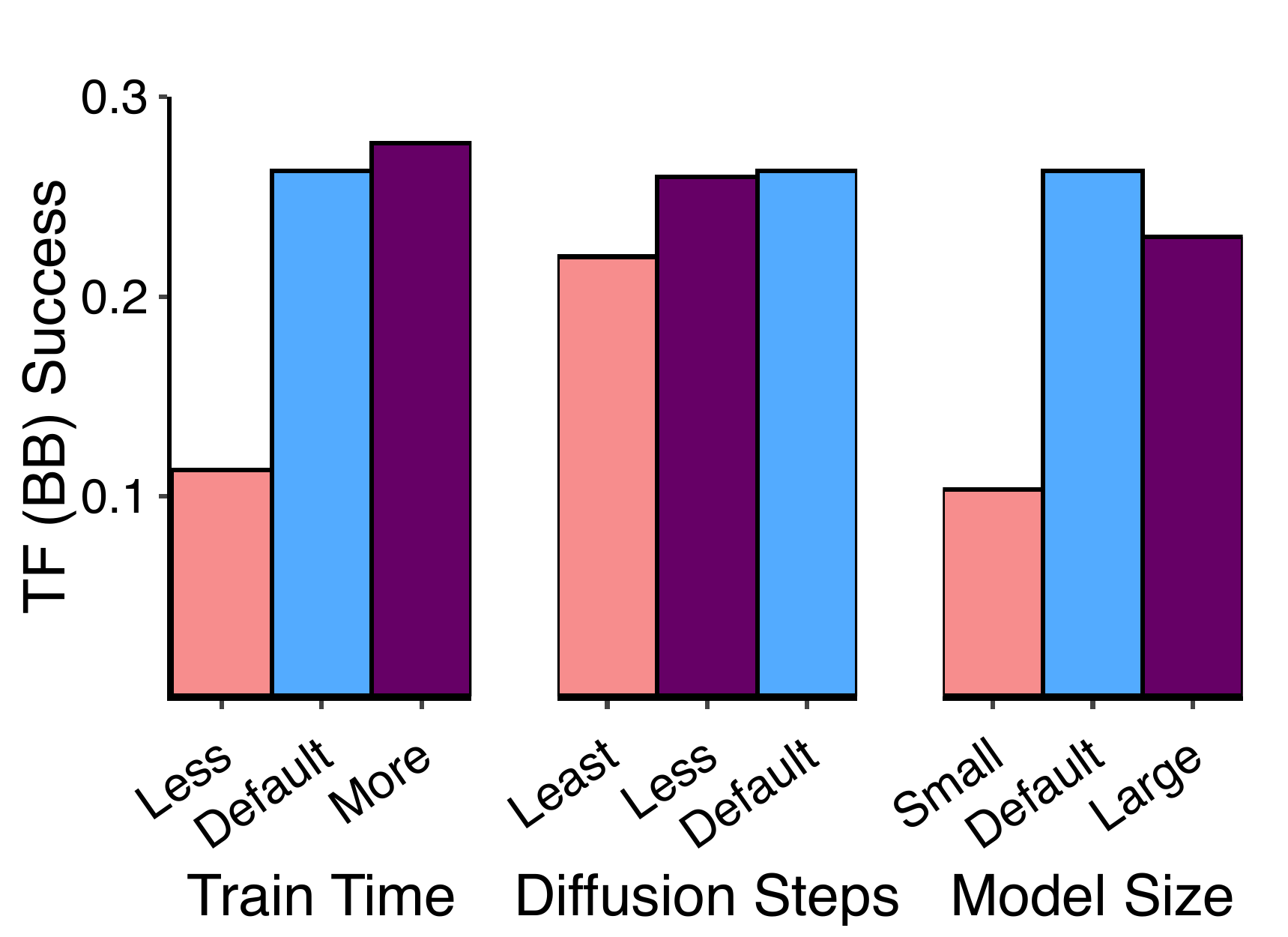}
\end{subfigure}}
\raisebox{-0.5ex}{
\begin{subfigure}{0.325\textwidth}
\includegraphics[width=0.99\linewidth]{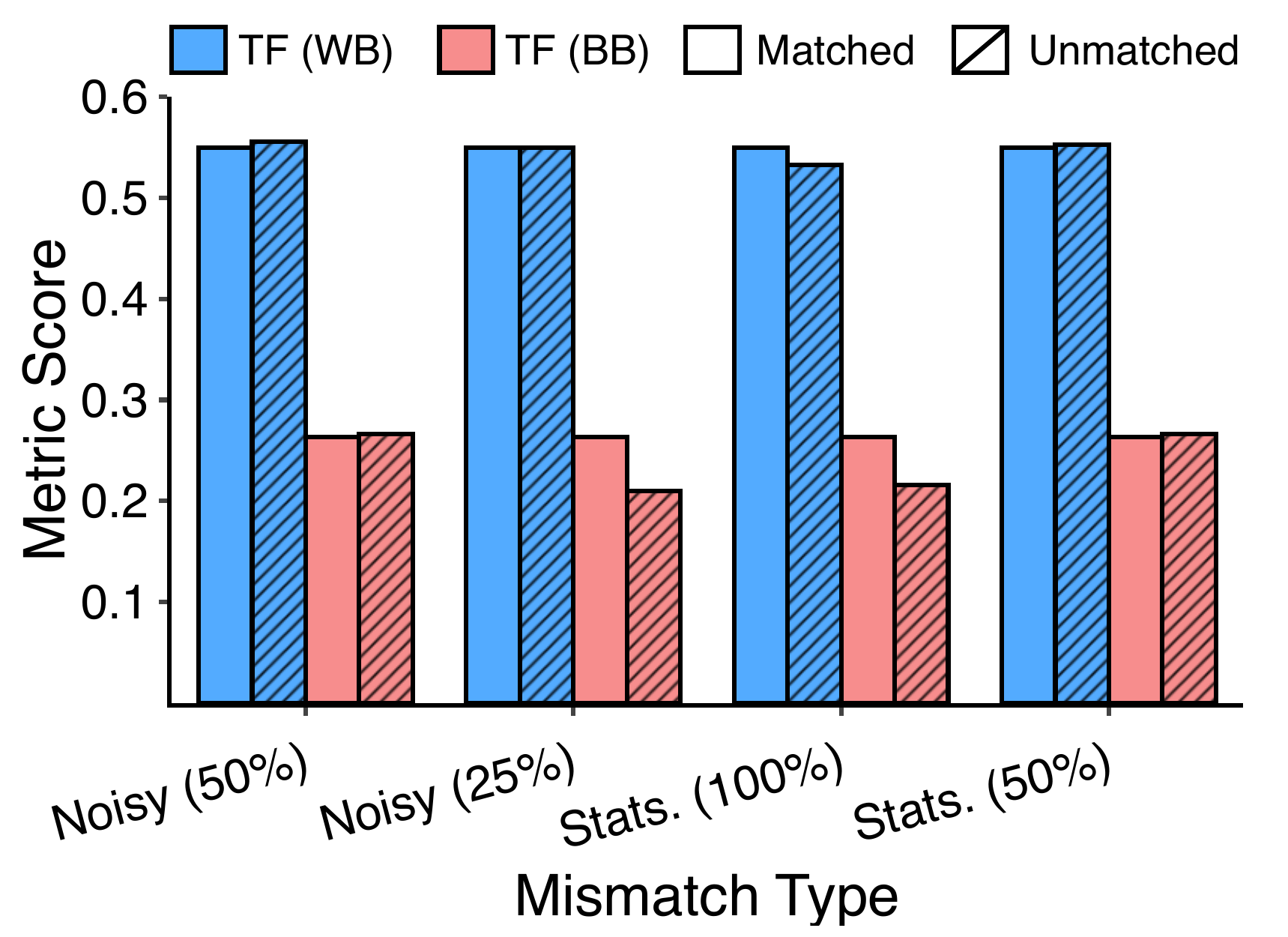}
\end{subfigure}}
\caption{Variations in attacker compute (left), shadow model mismatch (center) and shadow-model data mismatch (right) vs. WB and BB TF MIA success for Berka (Row 1) and Diabetes (Row 2).} \label{fig:attacker_variations_tf}
\end{figure}

The privacy risks observed in previous sections accrue under the assumption that attackers have access to significant compute, perfect knowledge of training procedures, and a pool of IID reference data. Here, we consider whether such risks persist for TDMs beyond these ideal conditions using ClavaDDPM as the target model. Figure \ref{fig:attacker_variations_tf} displays the impact of variations in attacker compute power and imperfect knowledge or data access for the TF attack. The attack is surprisingly robust to reductions in the number of trained shadow models. For example, training a single such model reduces WB MIA success for both datasets by less than $0.03$, contrasting previous work \citep{Salem2019ML-Leaks, Hu2022MIASurvey}. This implies that adversaries need not apply heavy compute to construct successful TF attacks. MIA success is also resilient to shadow-model training mismatches. When trained on more data (Figure \ref{fig:attacker_variations_tf}, left) than the target, attack success remains high. Using larger shadow models than the target also preserves MIA success. When using a smaller shadow model, the attack remains effective for Berka, but does collapse for Diabetes. Training shadow models for significantly fewer iterations or with many fewer timesteps does noticeably reduce attack effectiveness. However, training such models for more iterations or with only marginally misaligned timesteps has little impact. The final set of experiments show that the TF attack is effective even when an attacker's data distribution differs in substantial ways. The BB version of the attack shows, at most, only small degradations across all scenarios. The WB attack is marginally impacted by the statistical synthesis and noisy data modifications, but remains quite strong.

\begin{figure}[ht!]
\begin{subfigure}{0.33\textwidth}
\includegraphics[width=0.99\linewidth]{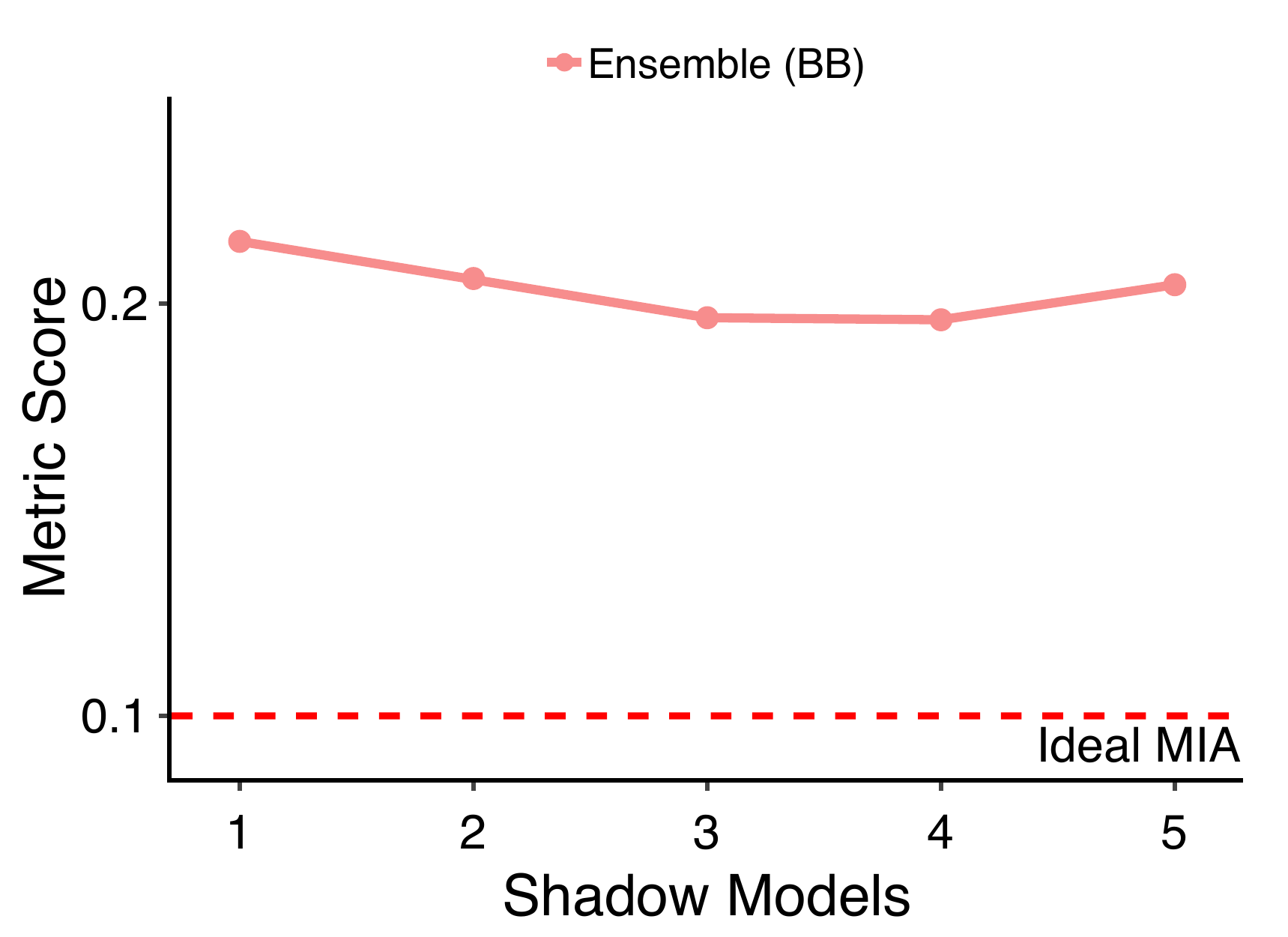}
\end{subfigure}
\begin{subfigure}{0.33\textwidth}
\includegraphics[width=0.99\linewidth]{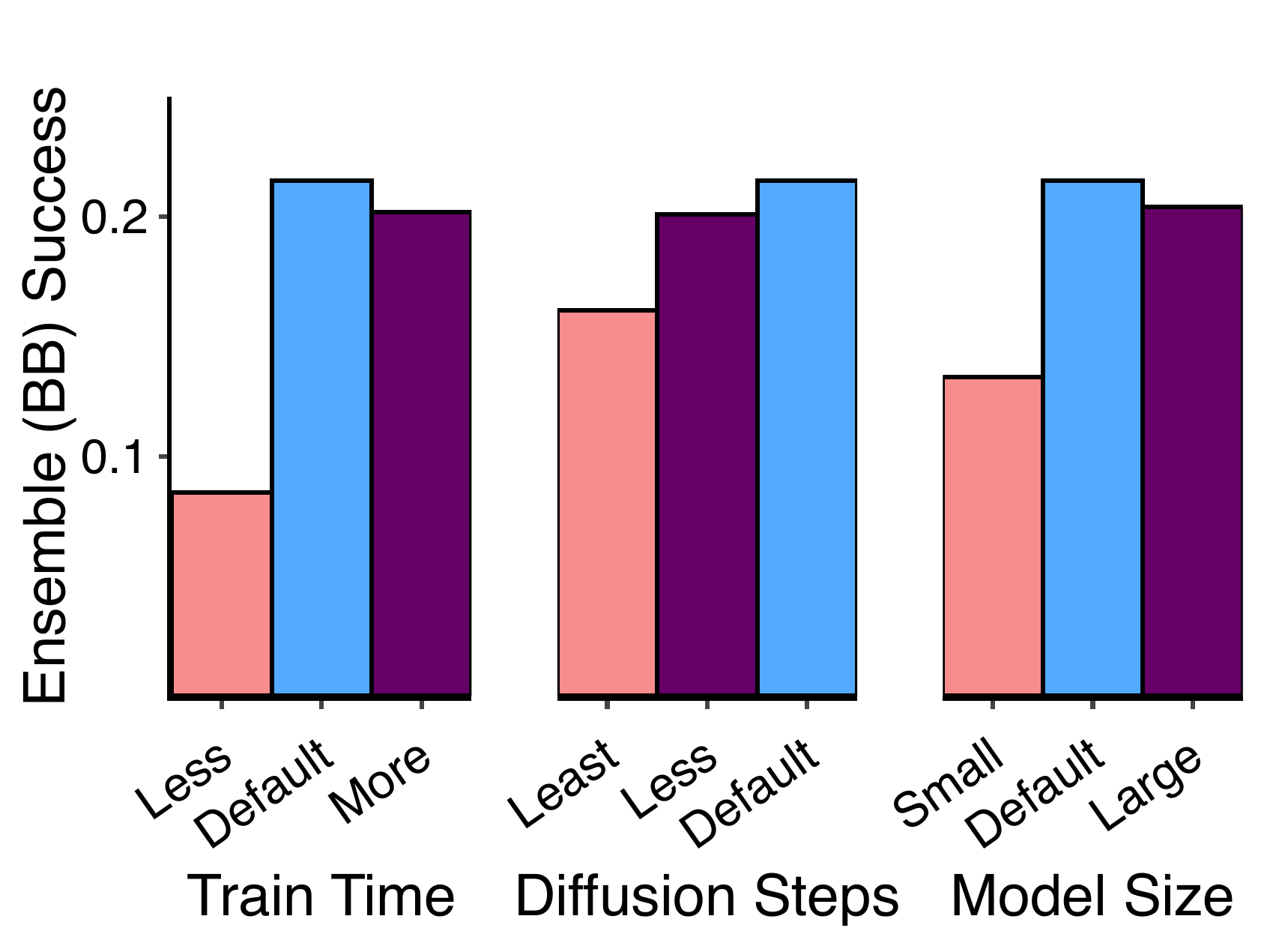}
\end{subfigure}
\begin{subfigure}{0.33\textwidth}
\includegraphics[width=0.99\linewidth]{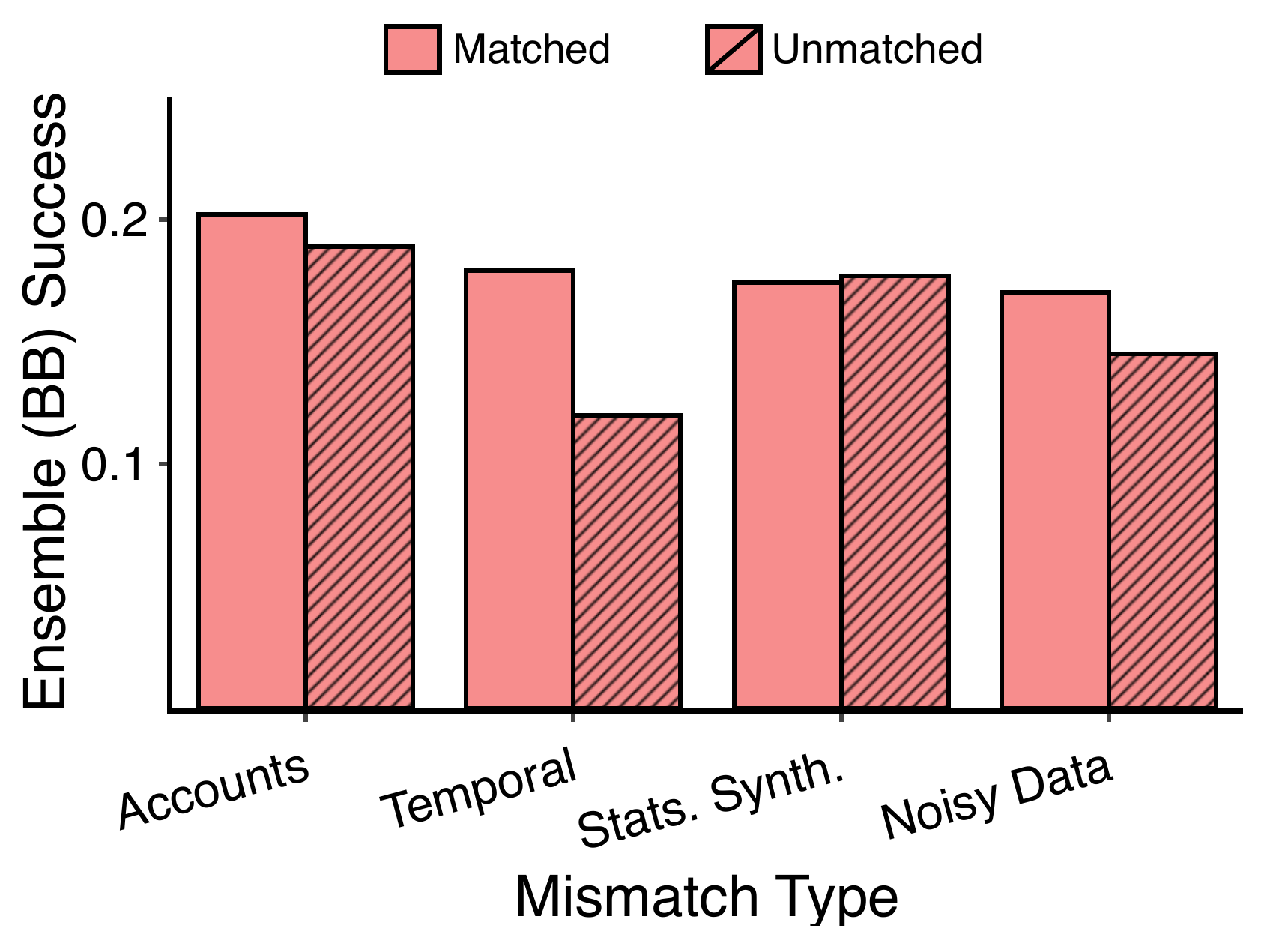}
\end{subfigure}
\begin{subfigure}{0.33\textwidth}
\includegraphics[width=0.99\linewidth]{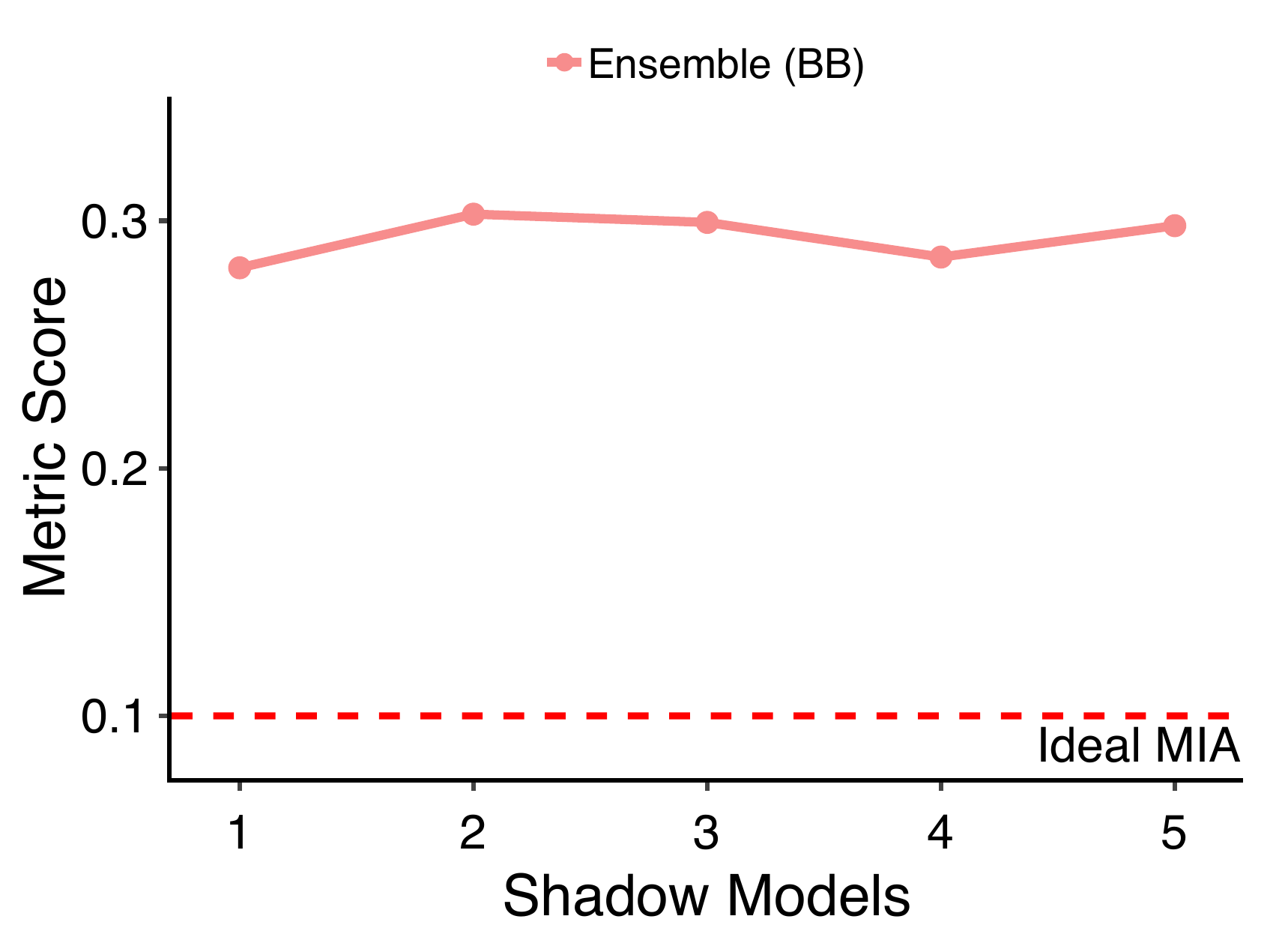}
\end{subfigure}
\begin{subfigure}{0.33\textwidth}
\includegraphics[width=0.99\linewidth]{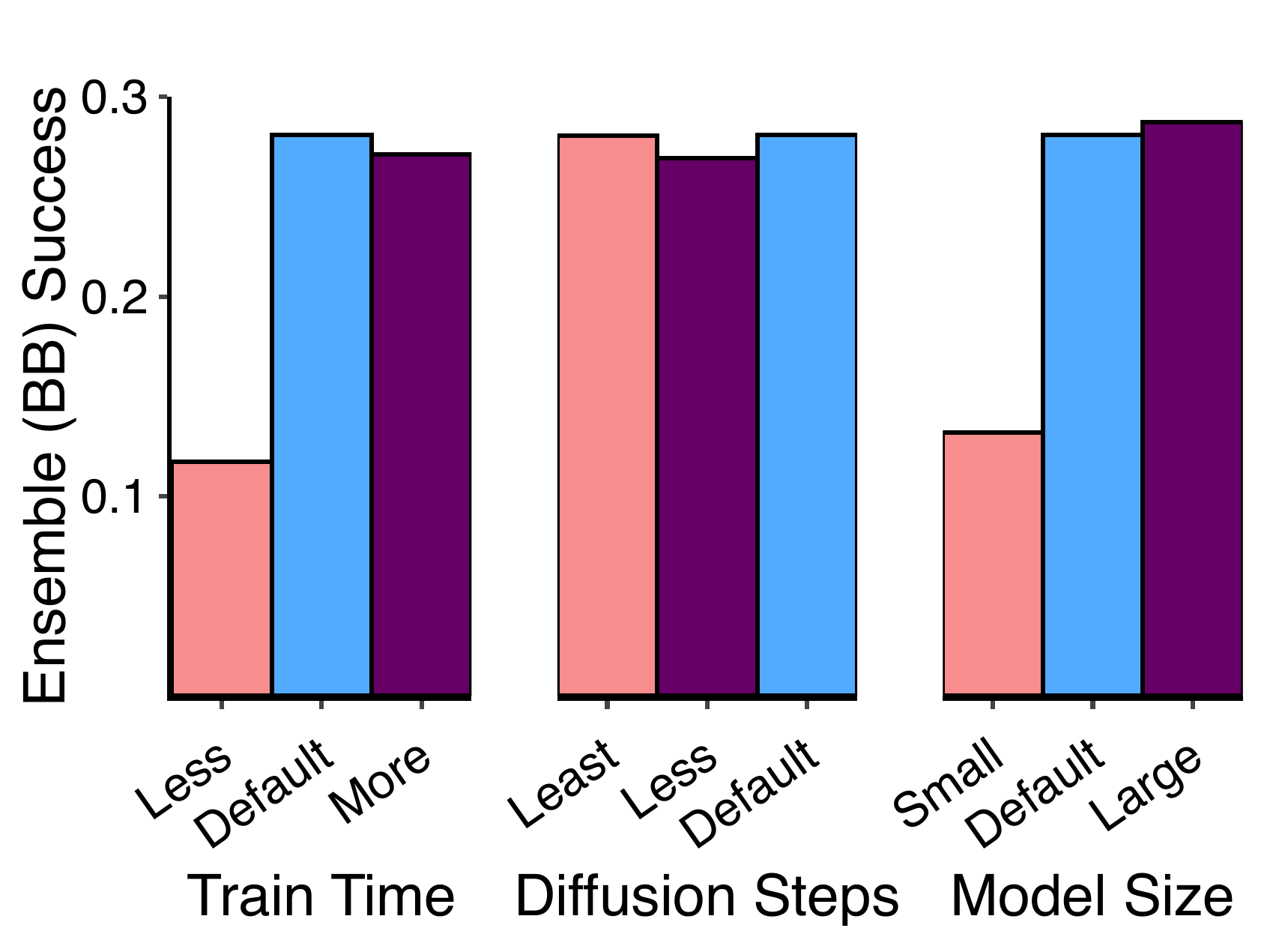}
\end{subfigure}
\begin{subfigure}{0.33\textwidth}
\includegraphics[width=0.99\linewidth]{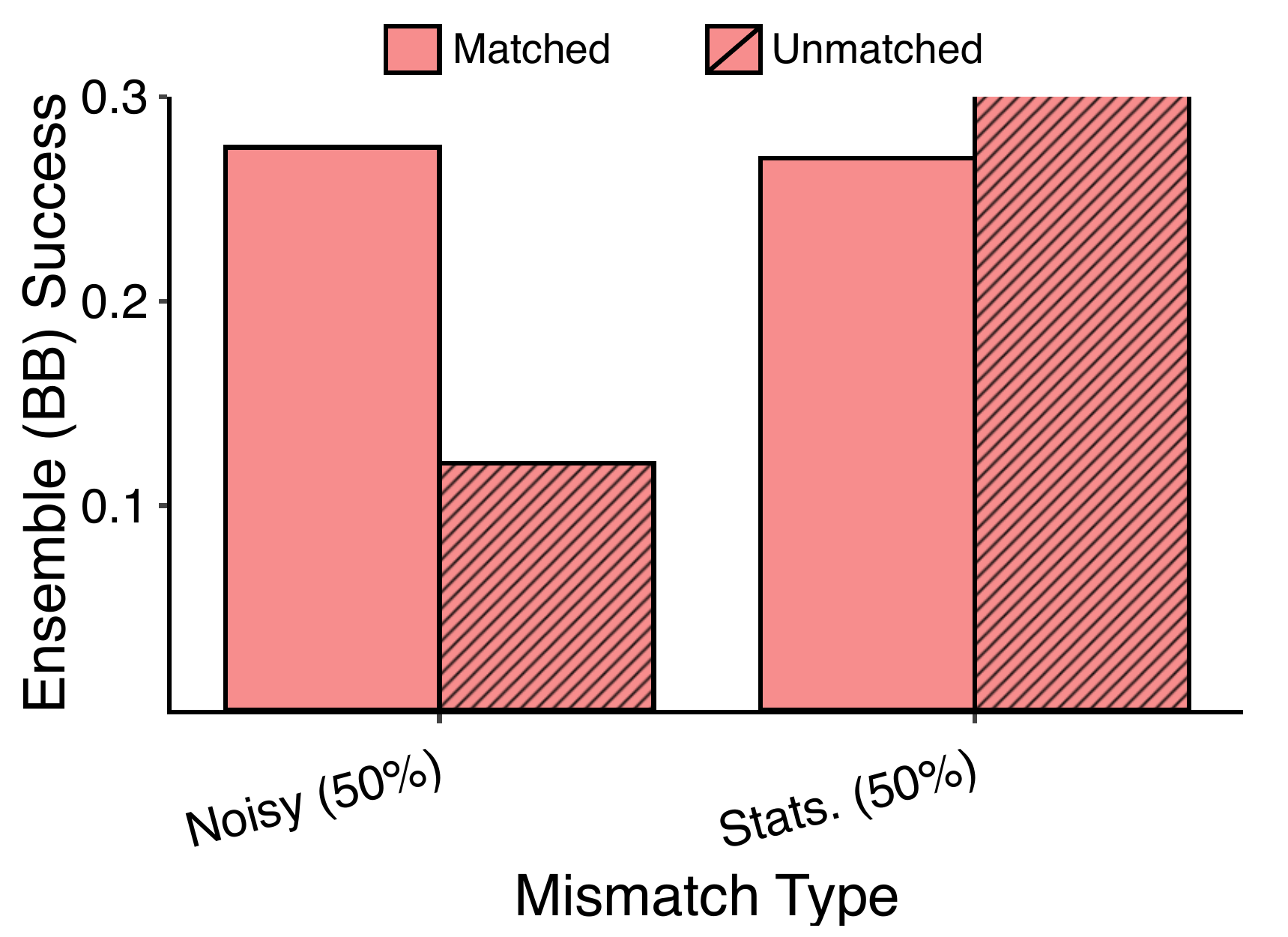}
\end{subfigure}
\caption{Variations in attacker compute (left), shadow model mismatch (center) and shadow-model data mismatch (right) vs. Ensemble MIA success for Berka (Row 1) and Diabetes (Row 2).} \label{fig:attacker_variations_ensemble}
\end{figure}

The results for similar variations in attacker compute, target training knowledge, and reference data access for the Ensemble attack are shown in Figure \ref{fig:attacker_variations_ensemble}. The default configuration for the attack constructs a single shadow model for meta-classifier training. The results demonstrate that this setup is sufficient and attackers need not train large collections of shadow models to produce successful MIAs. Similar to the TF attack, perfect knowledge of target-model training and architecture are not critical for Ensemble attack success. Longer training, fewer timesteps, or larger shadow models minimally degrade MIA success. Finally, for the Berka dataset, the Ensemble attack is robust to certain kinds of distribution divergence, especially disjoint accounts and statistically synthesized data. However, as a data-driven approach, it is more sensitive to temporal shifts and data noise. For Diabetes, the attack is resilient to certain kinds of data disagreement, but is impacted by heavily noised data. Additional results are discussed in Appendix \ref{ensemble_ablation_results}. Taken together, these results imply that successful attacks on TDMs may be constructed well beyond idealized scenarios.

\section{Conclusions, Limitations, and Future Work}

In this work, factors that influence privacy leakage in data-space TDMs are quantified through controlled experimentation. Such factors include training choices, synthesis volume, and limitations on attacker compute, knowledge, and data access. Through the lens of leading WB and BB MIAs specifically designed for TDMs, we show that access to large compute resources, perfect knowledge of hyperparameters and model architecture, and access to IID data are not prerequisites to constructing successful attacks. Moreover, DCR, and other widely used pseudo-privacy risk metrics, are shown to be ineffective gauges of privacy risk in many scenarios, going well beyond previous studies. A potential limitation of this work is the specific focus on data-space TDMs. The development of strong MIAs for latent-space TDMs like TabSyn remains an open problem \citep{cheng2025membershipinferencediffusionmodelsbasedsynthetic, tabsyn, MIDST2025}, precluding evaluation of the same levers considered here for such models. Future work will focus on developing such attacks and conducting a similarly extensive analysis of privacy leakage for these models. Further, we plan to apply the insights from this work to improve the utility-privacy tradeoffs in DP-training of diffusion models.

\subsection*{AI use statement}

Generative AI tools were not used in place of human researchers for any core tasks in this research, including those requiring disclosure as part of the ICLR 2027 guidelines. LLMs were sparsely used in the course of writing the manuscript to improve readability by exploring potential rephrasings and grammatical structures. No central conclusions, takeaways, or ideas are derived with the help of Generative AI tools. All writing elements have been thoroughly reviewed by the authors. We take full responsibility for the final content of this work, including text, claims, or artifacts.

\subsection*{Ethics statement}

All authors have read and respected the ICLR Code of Ethics through the course of this research. While the topic of this work concerns privacy leakage in ML models, no real privacy breach occurs in the experiments. We work with existing MIAs and use publicly available datasets, highlighting the impact of various training choices, attacker knowledge, and data access on privacy leakage. These experiments aim to inform practitioners of the risks associated with synthetic tabular data and tabular diffusion models. This includes how to better assess and control privacy leakage risks.

\subsection*{Reproducibility statement}

The paper includes extensive details to ensure that the datasets, preprocessing,
training setup, and experimental framework are fully understandable and reproducible for interested readers. The Methodology section in the main body and Appendices A--D, and G document the details associated with the main results and conclusions of this work. The code associated with the main results, including model training, attack construction, and evaluation will be made public along with this work. An anonymized version of the code used for the ClavaDDPM experiments and utility evaluation is here: \url{https://github.com/VectorInstitute/midst-toolkit}. For the TabDiff experiments, the repository associated with \citep{mahyar2026fermiexploitingrelationsmembership} was kindly shared with us by the authors. The datasets used in this work, namely the Berka and Diabetes datasets, are publicly accessible through Kaggle and the UCI Machine Learning Repository, respectively. The experiments in this paper produce a large quantity of synthetic data and model artifacts, including many shadow models in various forms. Due to the volume of these assets, they will not be publicly available, but will be made available upon request. Finally, the computational resources required to run the experiments reported in the main results of this paper are discussed in detail in Appendix H.



\bibliography{references}
\bibliographystyle{iclr2027_conference}

\appendix

\section{Training and Attacker Configuration Experiment Details} \label{experiment_details}

As noted in Section \ref{experiment_configurations}, the default configurations for ClavaDDPM are adapted from \citep{Pang2025ClavaDDPM}, while the default settings for TabDiff are drawn from \citep{mahyar2026fermiexploitingrelationsmembership}. 

\subsection{ClavaDDPM Default Configuration and Variations}

For ClavaDDPM, the default settings are as follows. The target models are DNN-based diffusion architectures with six layers and hidden dimension sequence [$512$, $1024$, $1024$, $1024$, $1024$, $512$]. Each model is trained for $200000$ steps with a batch size of $4096$. The number of diffusion timesteps is $2000$. For Berka, each target model is trained on $20000$ distinct data points and generates an equal number of synthetic points such that $\vert \mathcal{D}_t^i \vert = \vert \mathcal{D}_s^i \vert = 20000$. A total of $n_t = 10$ target models are trained, each with a unique challenge set of size $\vert \mathcal{C}_i \vert = 200$. Alternatively, for the Diabetes dataset, target-model training and synthesized datasets are comprised of $10000$ samples. There are three target models, each with $\vert \mathcal{C}_i \vert = 1000$. The focus of this work is record-level membership inference. As such, in constructing all datasets for the Berka and Diabetes experiments, any identifier columns are excluded and records are randomly sampled without replacement, including $\mathcal{D}_t^i$, $\mathcal{C}_i$, and others.

In the TF attack default setup, adversaries train $20$ shadow models using $20000$ points drawn without replacement from $\mathcal{D}_h$ for the Berka dataset. For Diabetes, seven shadow models are trained using $10000$ points each. When applying the Ensemble attack for ClavaDDPM, a single target shadow model is trained by default, and $16$ RMIA shadow models. The attack draws samples of $20000$ points without replacement from $\mathcal{D}_h$ for Berka and $10000$ samples for Diabetes to train the attack classifier.

In experiments modifying the diffusion model architecture, changes for each setting are as follows. The ``narrow, shallow, short'' setting uses a DNN with layers and dimensions [256, 512, 512, 256] and trains for $100000$ iterations. The ``narrow, shallow'' setting uses the aforementioned DNN but trained for $200000$ iterations. For the ``wider, deeper, long'' configuration, the DNN is increased to have layers and dimensions [1024, 2048, 2048, 2048, 2048, 2048, 2048, 1024] and trains for $300000$ steps. The ``wider, deeper'' experiment uses the larger DNN, but returns to $200000$ training steps. 

\subsection{TabDiff Default Configuration and Variations}

The TabDiff target models are composed of transformer-based diffusion models. In the default setting, these models incorporate two transformer blocks, each with a DNN hidden dimension of $1024$. Each model uses a training batch size of $4096$. While TabDiff is a continuous-time diffusion model during training, inference is performed with discrete timesteps. The default number of diffusion steps is $50$. For Berka, each target model is trained on $20000$ points for $20000$ epochs and generates an equal number of synthetic points. In each experiment, $n_t=5$ target models are trained, each with a unique challenge set of size $\vert \mathcal{C}_i \vert = 2000$. For the Diabetes dataset, target-model training and synthetically generated datasets have $10000$ samples. By default, these models are trained for $8000$ epochs. There are $3$ target models, each with $\vert \mathcal{C}_i \vert = 1000$. For the default TF attack setup, adversaries train $10$ shadow models using $20000$ points drawn without replacement from $\mathcal{D}_h$ for Berka. For Diabetes, seven shadow models are trained using $10000$ points each. Similar to the ClavaDDPM experiments, all membership inference is record-level identification. For the Diabetes experiments, all datasets are constructed by randomly sampling records without replacement, as in the ClavaDDPM setup. For the Berka dataset, the partitions, challenge point compositions, and shadow model training sets are identical to that used in \citep{mahyar2026fermiexploitingrelationsmembership}. This includes preprocessing procedures.

When modifying the TabDiff model architecture, the number of training steps is held constant for each dataset. Two dimensions of variation are considered. The number of repeated transformer blocks (depth) is varied from $D=1$ to $D=3$. For each depth, the size of the DNN latent dimension (width) is varied from $W=512$ to $W=2048$ for Berka. Due to the smaller dataset size of Diabetes, only $W=512$ and $W=1024$ are considered. In the figures, these variations are denoted as $W \times D$ along the x-axis.

\subsection{Attack Resources, Knowledge, and Data Experimental Details: ClavaDDPM} \label{attack_mismatch_details}

For experiments incorporating mismatches between target and shadow model setups for ClavaDDPM, shorter training times correspond to just $5000$ steps, while longer training consumes $300000$ steps. When varying diffusion timesteps, the model with the ``least'' steps uses only $10$ and the ``less'' setting has $100$. The small shadow models are DNN-based architectures with layers and dimensions [256, 512, 512, 256], and the large models are [1024, 2048, 2048, 2048, 2048, 2048, 2048, 1024]. For Figure \ref{fig:attacker_variations_tf} (left), where experiments induce mismatches between the training data volume of the shadow and target models, the default setting corresponds to the left-hand bars with a shadow-model training dataset size of $20000$ for Berka and $10000$ for Diabetes. The training dataset size for the shadow models is increased, but the target model dataset is held constant, along with all other settings and hyperparameters.

For the Berka dataset, when simulating imperfect adversary data access via disjoint accounts, accounts, and their corresponding transactions, are randomly assigned to the attacker or target model datasets such that there is no overlap. When segmenting transactions temporally, target models are trained on transactions occurring on or before 1997-04-10 and adversary data is drawn from transactions occurring thereafter. The scenario of statistics-based synthesis considers the setting where the adversary only possesses knowledge of population-level marginal feature distributions and can sample from them. To simulate this, we randomly permute the rows of all columns independently. This preserves the marginal distribution of each column but removes all inter-feature dependencies. 

\begin{table}[ht!]
\caption{Comparison of the statistical distributions of the training data for target model and the adversary data used for training the shadow models under different distribution mismatch scenarios for the Berka dataset. ``Corr.\ $\Delta$'' is the Frobenius-norm difference between correlation matrices.  ``MI $\Delta$'' is the Frobenius-norm difference between mutual-information matrices (categorical features only). ``\% Diff.'' is the fraction of feature columns whose empirical marginals differ significantly ($\alpha=0.05$). For numerical columns, distributions are compared using a Kolmogorov-Smirnov test, while categorical columns are compared using Total Variation Distance with significance established via a permutation test ($1000$ permutations).}
\centering
\begin{tabular}{lccc}
\toprule
Mismatch Type &
Corr.\ $\Delta$ &
MI $\Delta$ &
\% Diff. \\
\midrule
Disjoint Accounts
  & 0.010 & 0.080 & 25.0 \\
Disjoint Time Windows
  & 0.142 & 0.029 & 87.5 \\
\midrule
Statistics-Based Synthesis
  & 2.100 & 1.518 & 0.0 \\
Noisy Adversary Data
  & 2.000 & 1.449 & 100.0 \\
\bottomrule
\end{tabular}
\label{tab:berka_distribution_mismatch}
\end{table}

Finally, the noisy adversary data scenario assumes the adversary has significantly less information. That is, they know the unique values in each column but have no knowledge of the population-level marginal feature distributions. To simulate this, we replace each numerical column with draws from a uniform distribution over that column’s observed range, and each categorical column with uniform draws over its observed categories. These perturbations induce substantial changes to the correlation structure, as reflected in the large correlation and MI matrix differences in Table \ref{tab:berka_distribution_mismatch}. 

For the Diabetes dataset, adversary data mismatches are induced via the same noisy and statistics-based synthesis strategies used for Berka. For both attacks, adversary data is modified by randomly selecting half of the numerical columns and half of the categorical columns and either applying the noisy or statistics-based synthesis strategies. Changes to the correlation structure for these manipulations are reported in Table \ref{tab:diabetes_distribution_mismatch}. In addition, for the TF attack, we also consider the setting of noising $25$\% of each column type or statistics-based synthesis for $100$\% of the columns.

\begin{table}[ht!]
\caption{Comparison of the statistical distributions of the training data for target model and the adversary data used for training the shadow models under different distribution mismatch scenarios for the Diabetes dataset. ``Corr.\ $\Delta$'' is the Frobenius-norm difference between correlation matrices.  ``MI $\Delta$'' is the Frobenius-norm difference between mutual-information matrices (categorical features only). ``\% Diff.'' is the fraction of feature columns whose empirical marginals differ significantly ($\alpha=0.05$). For numerical columns, distributions are compared using a Kolmogorov-Smirnov test, while categorical columns are compared using Total Variation Distance with significance established via a permutation test ($1000$ permutations).}
\centering
\begin{tabular}{lccc}
\toprule
Mismatch Type &
Corr.\ $\Delta$ &
MI $\Delta$ &
\% Diff. \\
\midrule
Statistics-Based Synthesis (50\%)
  & 1.87 & 3.31 & 0.0 \\
  Statistics-Based Synthesis (100\%)  & 1.90 & 3.33 & 0.0 \\
  Noisy Adversary Data (25\%)  & 1.16 & 3.18 & 22.9 \\
Noisy Adversary Data (50\%)
  & 1.78 & 2.21 & 47.9 \\
\bottomrule
\end{tabular}
\label{tab:diabetes_distribution_mismatch}
\end{table}

In experiments varying training size for the Diabetes dataset, there is an important caveat to the experiments when the training sizes were $15000$ and $20000$. Because the size of the overall Diabetes set is limited, compared to Berka, only four and three shadow models are trained for attack development, while two models are still present as target models. This does imply that, in these settings, the attack has access to fewer shadow models. However, the results varying the number of shadow models for TF attack training in Figure \ref{fig:attacker_variations_tf} demonstrate that this does not dramatically undermine attack success.

\section{Tartan Federer Attack: Default Settings and Other Details} \label{tf_appendix}

As described in Section \ref{methodology}, the TF attack constructs features for each input, $x_0$, by computing the loss $\Vert (m_{\theta}(x_t, t) - \epsilon \Vert_2^2$ for a fixed set of $\epsilon$ and $t$ values. Note that the attack loss is modified for TabDiff following the description in \citep{mahyar2026fermiexploitingrelationsmembership}. 

\subsection{ClavaDDPM Setup}

For the default number of diffusion steps, $2000$, following \citep{wu2025winning}, a static set of $300$ initial noise values, $\epsilon$, are sampled from a standard Gaussian and the timesteps are $t\in \{5, 10, 20, 30, 40, 50, 100\}$. This yields $2100$ features in total to be processed by the DNN for membership inference. When varying the number of timesteps in model architectures, models with timesteps in $\{500, 1000, 2000, 3000, 4000\}$ use the aforementioned set of timesteps. For models with $\{10, 20, 50, 80, 100\}$ timesteps, $t \in \{3, 4, 5, 6, 7, 8, 9\}$. These settings are used for both datasets. In all setups, the DNN used for membership classification has two hidden layers, each with dimension $200$. Hidden layers have $\tanh$ activations and the output layer is a sigmoid. Each shadow model contributes $3000$ member and $3000$ non-member samples. As such, in the Berka experiments with $20$ shadow models, this yields $120000$ training points for the network, whereas the Diabetes setting has seven shadow models, producing $42000$ training samples. The classifier is trained with standard binary cross-entropy loss, an Adam optimizer with default parameters, and a learning rate of 1e-04. Training proceeds for $5000$ steps with batch sizes of $6000$.

\subsection{TabDiff Setup}

The TF attack for TabDiff replicates the settings of \citet{mahyar2026fermiexploitingrelationsmembership}. The TF loss features are constructed at time values $t\in \{0.01, 0.05, 0.1, 0.2, 0.4, 0.6, 0.9\}$. Similar to ClavaDDPM, a fixed set of $500$ noise values are sampled. Numerical and categorical column loss features are constructed in distinct ways, consistent with \citet{mahyar2026fermiexploitingrelationsmembership}. As such, each row yields a feature vector of dimensionality
$7 \cdot 500 \cdot 2 = 7000$ for both datasets in the experiments. The DNN used for membership classification has four hidden layers, each with dimension $128$. Hidden layers have $\relu$ activations and the output layer is a sigmoid. Batch normalization and dropout, with a rate of $0.2$, follow each dense layer. Each shadow model contributes $3000$ member and $3000$ non-member samples. As such, in the Berka experiments with $10$ shadow models, this yields $60000$ training points for the network, whereas the Diabetes setting has seven shadow models, producing $42000$ training samples. The classifier is trained with standard binary cross-entropy loss, an Adam optimizer with default parameters, and a learning rate of 5e-03. Training proceeds for $5000$ steps with batch sizes of $1024$.

\section{Ensemble Attack: Default Settings and Other Details} \label{ensemble_appendix}

By default, the attack classifier uses $20000$ samples from the Berka dataset or $10000$ samples from the Diabetes dataset drawn from $\mathcal{D}_h$ to which the attacker has access. The entire sampled subset is used to train the target shadow model. Further, half of the sample set is then used to train each of the RMIA shadow models. More specifically, each sample is randomly assigned to be used as training data for half of the RMIA shadow models. To construct the DOMIAS features, the full population data, $\mathcal{D}_h$, available to the attacker is used as the reference dataset, in line with the attack setup for the MIDST competition \citep{MIDST2025}. For Berka, this consists of approximately $800000$ samples and $70000$ samples for the Diabetes dataset. The use of the entire reference dataset is relaxed in the ablations of Appendix \ref{ensemble_ablation_results}.

In the default settings, 16 different RMIA shadow models are trained, in agreement with the algorithm from \citep{MIDST2025}. These are split into models with pre-train and fine-tune phases and those strictly with a single training phase. The first category of models are pre-trained on a subset of $\mathcal{D}_h$ not used to train the primary target shadow model constructed for the Ensemble attack. For Berka, this set is of size $60000$, and for Diabetes it has size $50000$. Thereafter, they are fine-tuned on the constructed subset of member points, as described above, from the training data of the target shadow model. For all experiments, except those in the results of Appendix \ref{ensemble_ablation_results}, two models are pre-trained and four models are fine-tuned from each of these base models, for a total of eight. In the second category, eight RMIA shadow models are also simply trained directly on a subset of data points drawn from the training data of the primary target shadow model with no pre-training phase.

The Ensemble attack meta-classifier is an XGBoost model trained on the features constructed through distance-based, DOMIAS-based, and RMIA-based extraction. The hyperparameters of the XGBoost model are optimized with 5-fold validation through Optuna \citep{akiba2019optunanextgenerationhyperparameteroptimization}. The parameters optimized are
\begin{itemize}
\item \texttt{eta=trial.suggest\_float("eta", 0.0001, 0.1, log=True)},
\item \texttt{max\_depth=trial.suggest\_int("max\_depth", 3, 10)},
\item \texttt{subsample=trial.suggest\_float("subsample", 0.1, 1)},
\item \texttt{colsample\_bytree=trial.suggest\_float("colsample\_bylevel", 0.5, 1)},
\item \texttt{reg\_alpha=trial.suggest\_categorical("reg\_alpha", [0, 0.1, 0.5, 1, 5, 10])},
\item \texttt{reg\_lambda=trial.suggest\_categorical("reg\_lambda", [0, 0.1, 0.5, 1, 5, 10, 100])}.
\end{itemize}

For the Berka dataset, the meta-classifier training set is a random selection of $10000$ points from the target shadow model training as member points and an equal number of samples drawn from $\mathcal{D}_h$ constituting non-members. The same is true for the Diabetes dataset, but with $5000$ points for each.

\section{TF and Ensemble Attack Ablation Results} \label{ensemble_ablation_results}

\begin{figure}[ht!]
\begin{subfigure}{0.49\textwidth}
\includegraphics[width=0.9\linewidth]{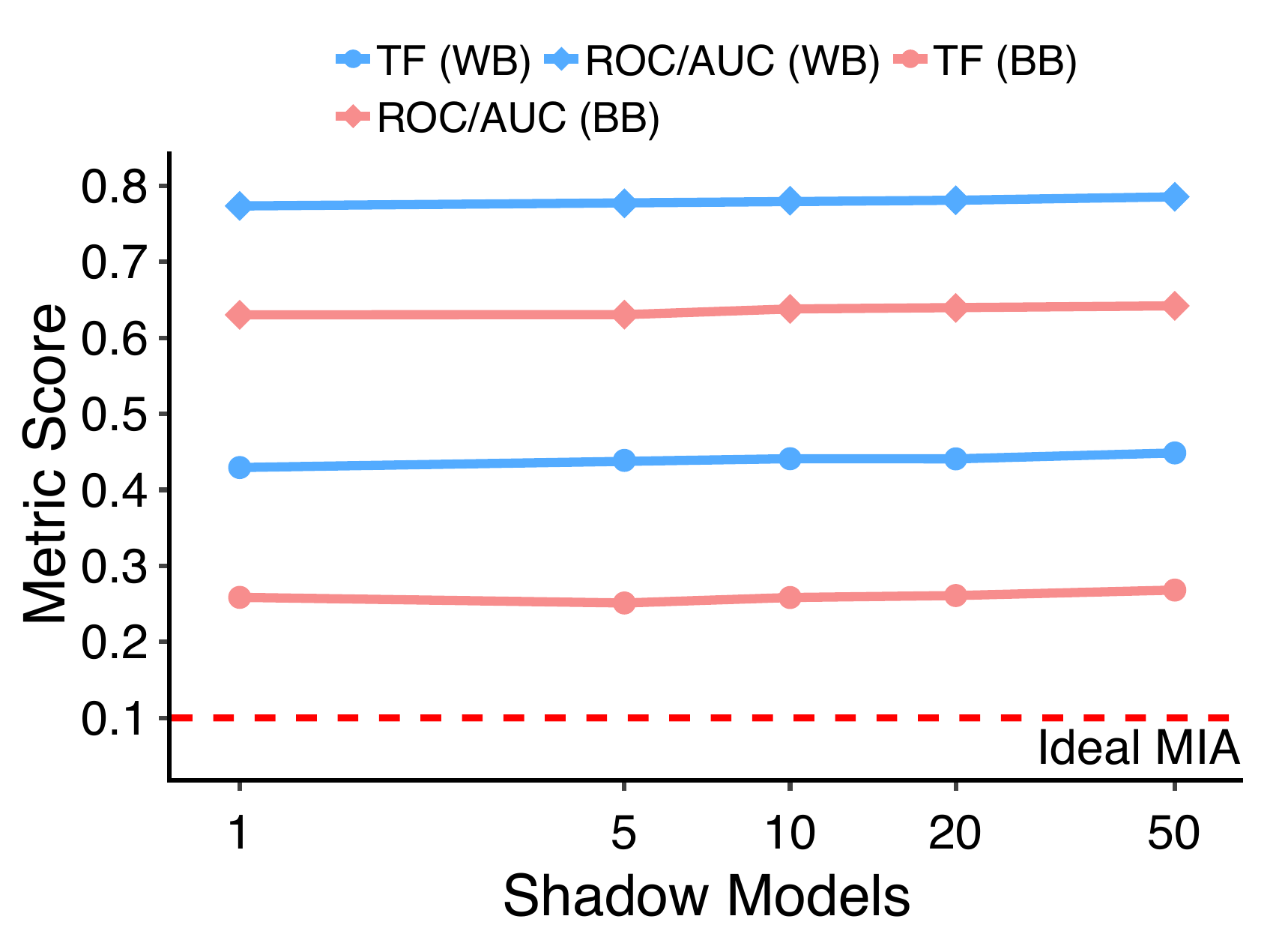}
\end{subfigure}
\begin{subfigure}{0.49\textwidth}
\includegraphics[width=0.9\linewidth]{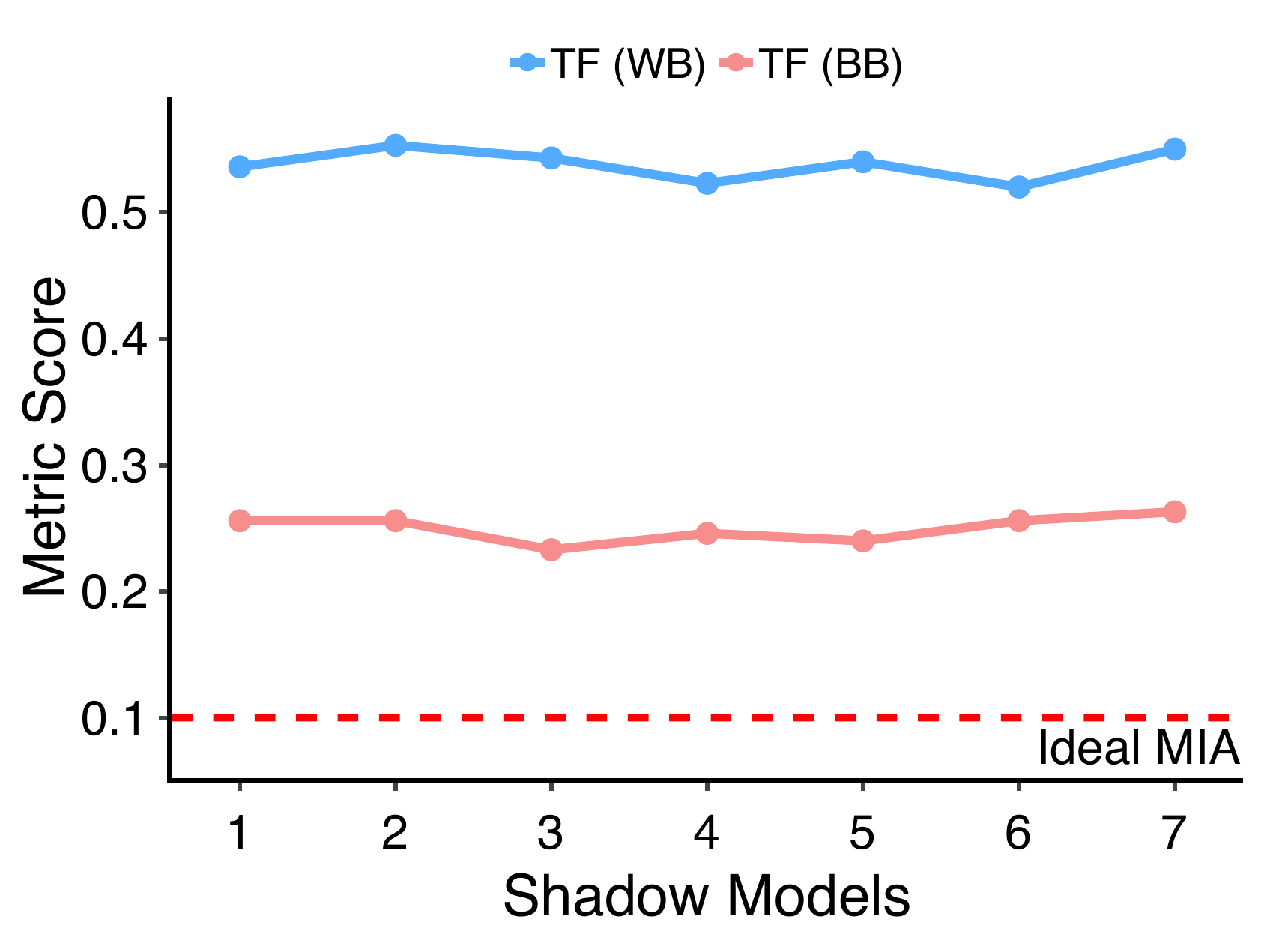}
\end{subfigure}
\caption{Variations in the number of shadow models used to construct the WB and BB TF attacks against ClavaDDPM target models for the Berka (left) and Diabetes (right) datasets. } \label{fig:tf_shadow_ablation}
\end{figure}

In this section, additional experimental results associated with ablations of the TF and Ensemble MIAs are shared. The results in Figure \ref{fig:tf_shadow_ablation} extend the experiments of Figure \ref{fig:attacker_variations_tf} (left) in terms of the number of shadow models trained for the WB and BB TF attacks across both datasets. While the success of the attack for both settings and across both datasets does improve with more models, the improvements are quite modest, implying that attacks need not have the compute or data resources required to train many shadow models in order to construct successful attacks.

\begin{figure}[ht!]
\begin{subfigure}{0.49\textwidth}
\includegraphics[width=0.9\linewidth]{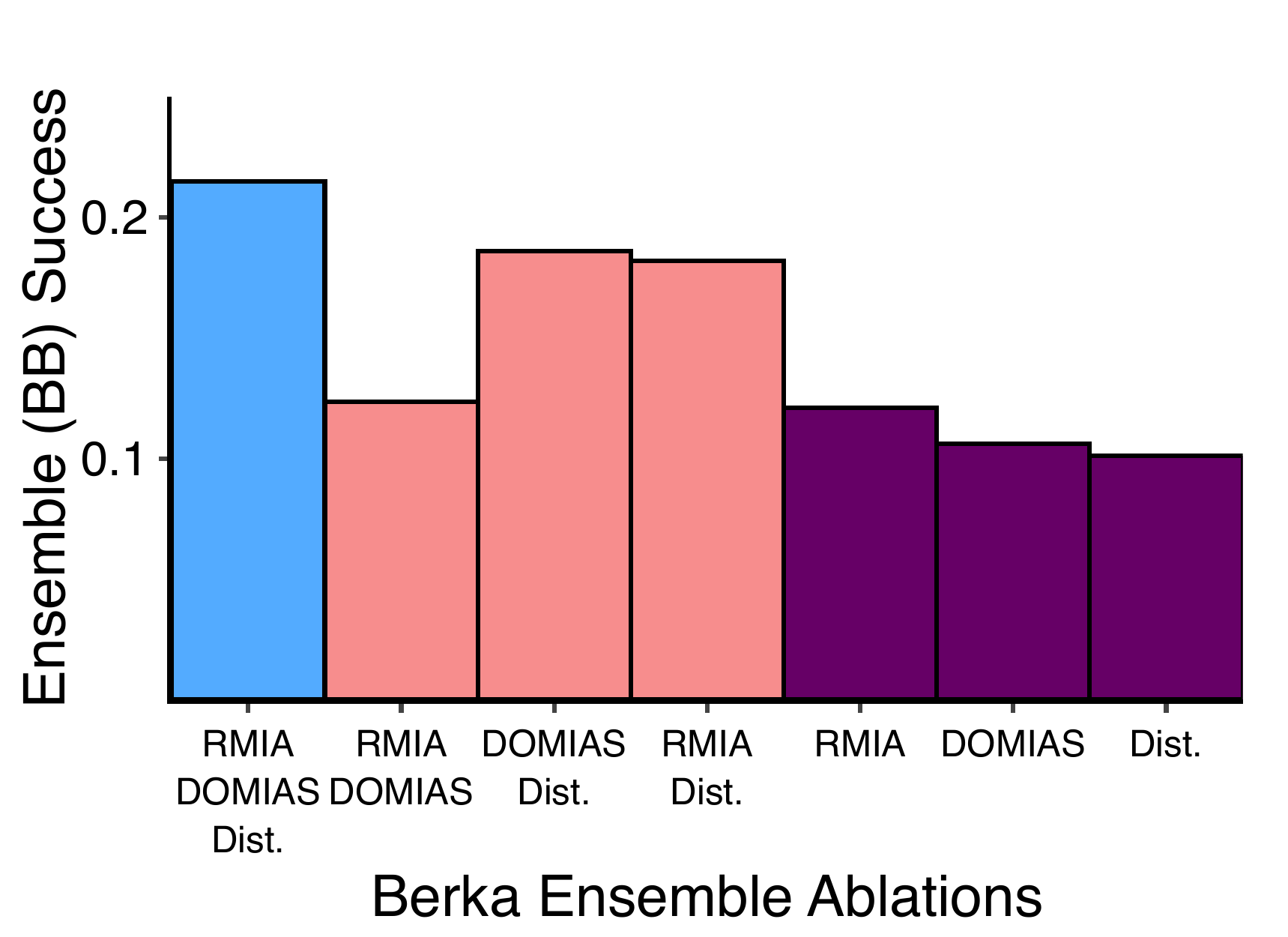}
\end{subfigure}
\begin{subfigure}{0.49\textwidth}
\includegraphics[width=0.9\linewidth]{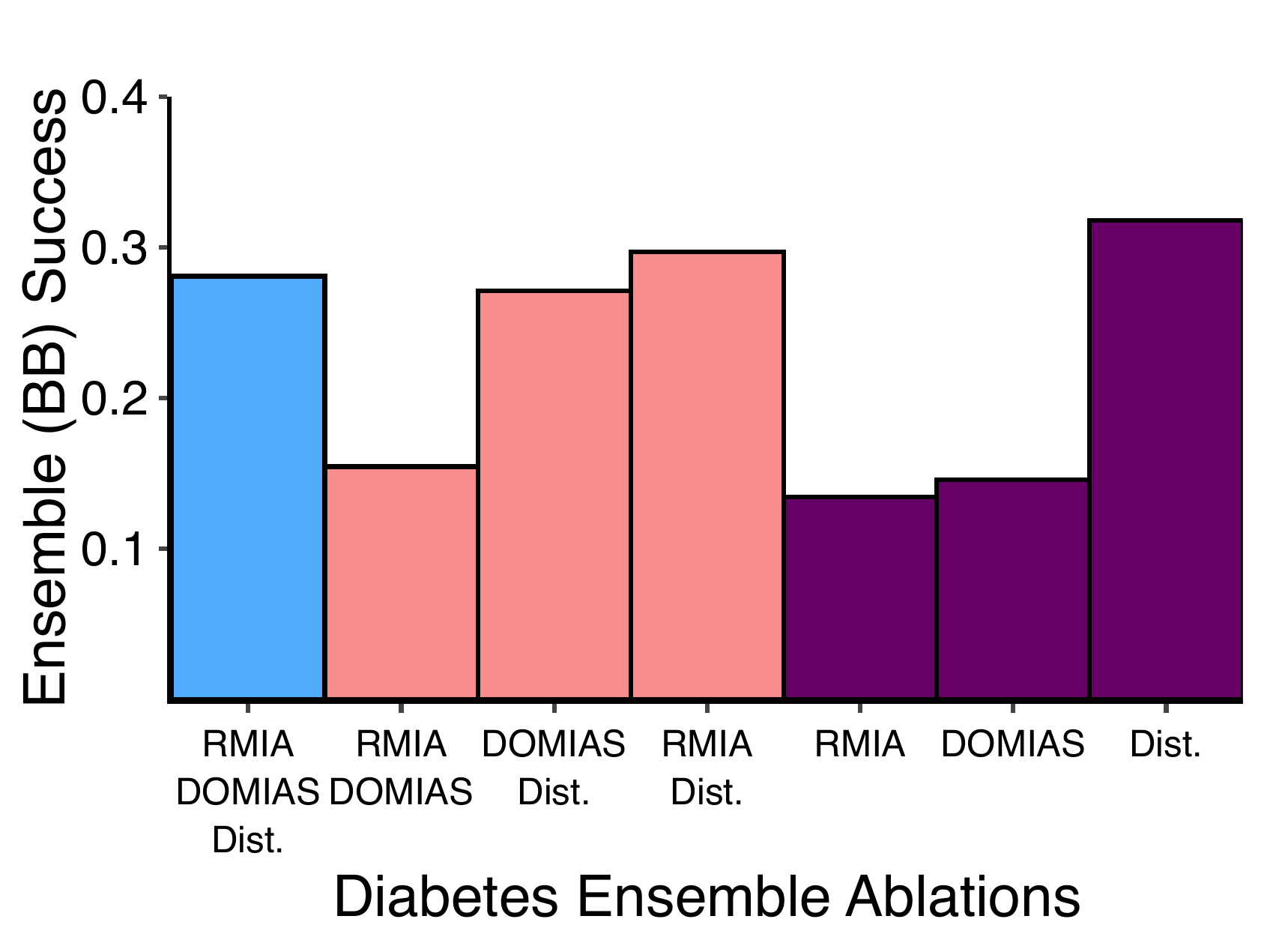}
\end{subfigure}
\begin{subfigure}{0.49\textwidth}
\includegraphics[width=0.9\linewidth]{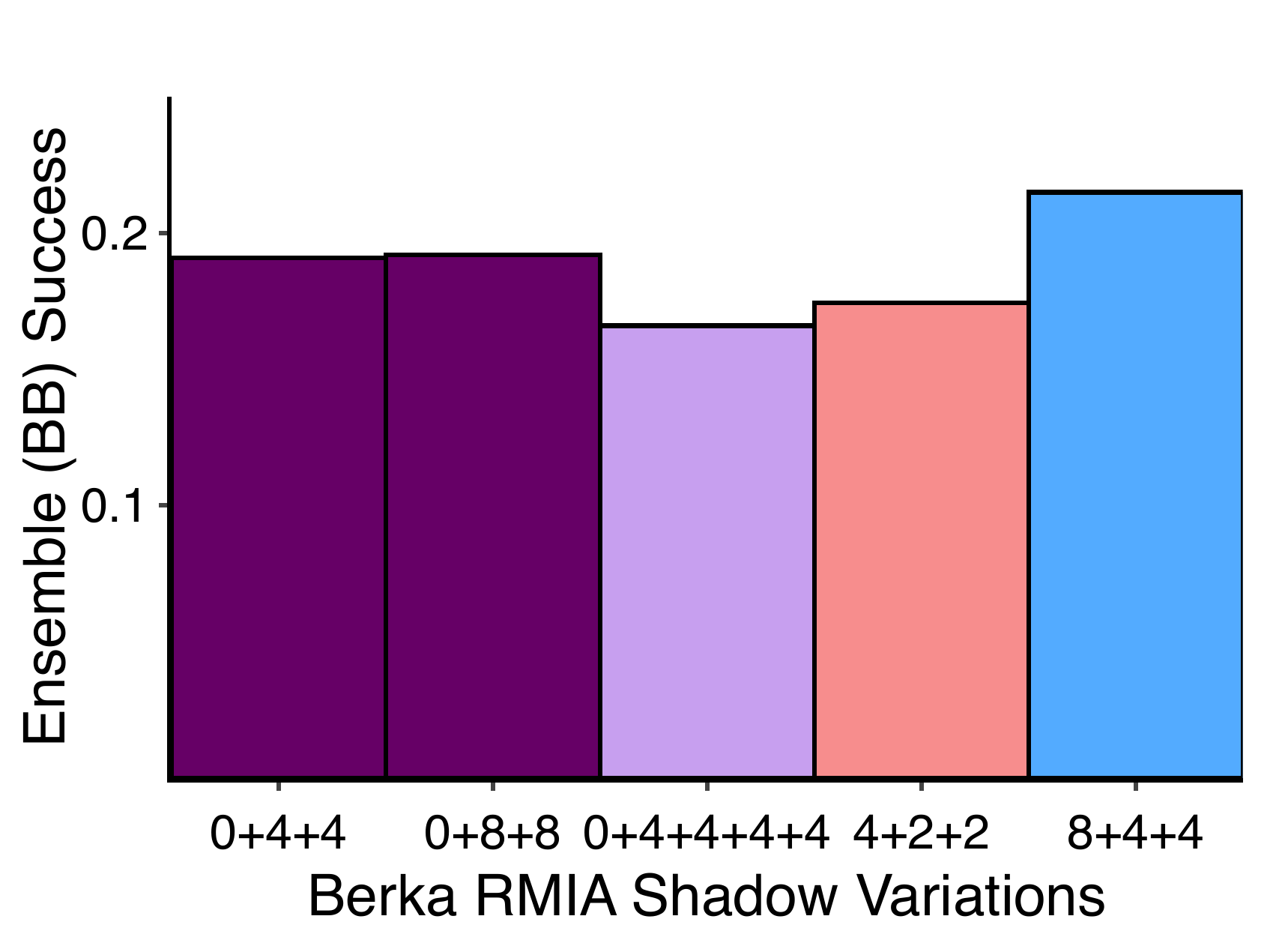}
\end{subfigure}
\begin{subfigure}{0.49\textwidth}
\includegraphics[width=0.9\linewidth]{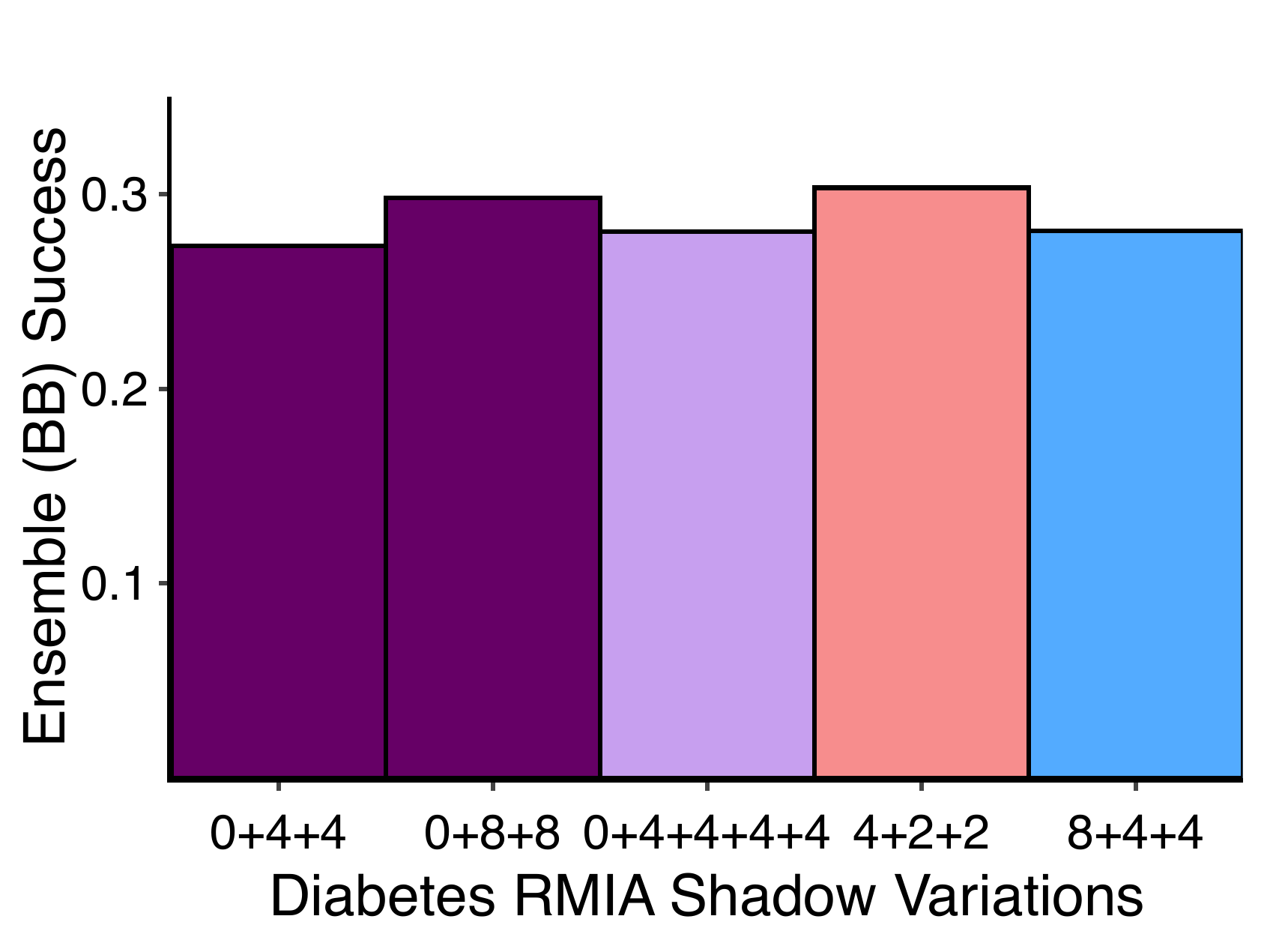}
\end{subfigure}
\caption{Ensemble attack ablations (top) and RMIA shadow model variations (bottom) for the Berka (left) and Diabetes (right) datasets. In the RMIA shadow experiments, the first number is single-phase training models, and the remaining are counts of fine-tuned models for each pre-trained base model. For example, 4+2+2 denotes four single-phase models and two sets of two models, each set trained on a distinct pre-trained base.} \label{fig:ensemble_ablation_summary}
\end{figure}

As previously discussed, the Ensemble attack combines three components in RMIA, DOMIAS, and Distance-based features through a trained meta-classifier. To understand the contributions of each component and their interactions, attacks are constructed excluding features associated with different combinations of the ensemble. The results are shown in the top row of Figure \ref{fig:ensemble_ablation_summary}. For the Berka dataset, there is a clear synergy associated with the components comprising the ensemble, as each one in isolation is insufficient to produce an MIA with significantly better than random performance. Interestingly, for the Diabetes dataset, the distance-based features are all that is necessary to produce a successful attack, whereas the RMIA and DOMIAS features alone are not effective. These results highlight that dataset composition plays a role in attack success and the strength of leveraging an ensemble-based approach to capture diverse inputs.

As detailed in Appendix \ref{ensemble_appendix}, the default construction of the RMIA approach is training 16 shadow models, eight single-stage and eight two-stage models. We experiment with alternative configurations to demonstrate robustness to this choice. The results appear in the bottom row of Figure \ref{fig:ensemble_ablation_summary}. While the most consistent arrangement for both datasets is the 8+4+4 split, other RMIA shadow-model compositions perform well above random chance even with half the number of models and a single stage of training.

\begin{figure}[ht!]
\centering
\includegraphics[width=0.49\linewidth]{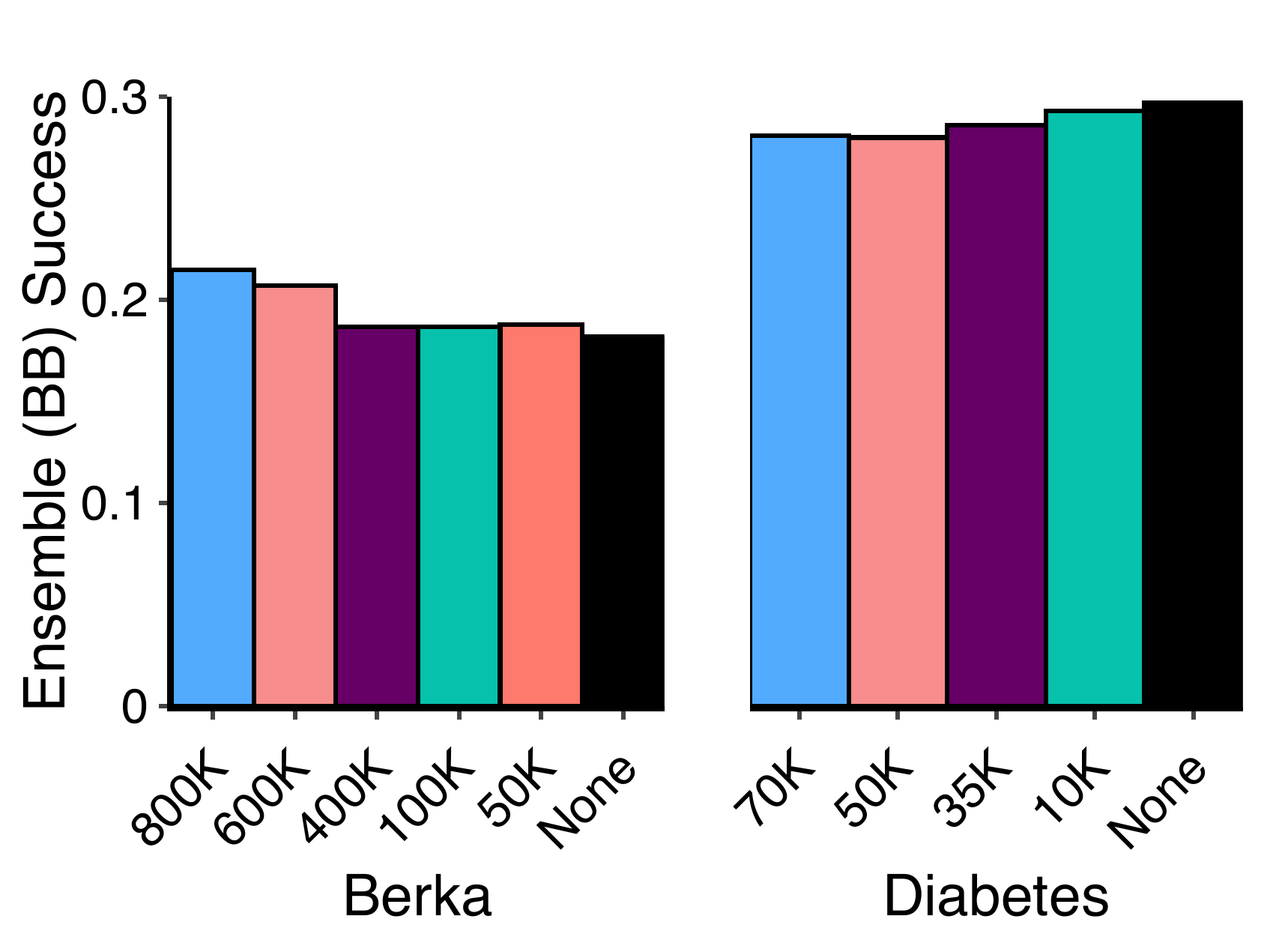}
\caption{Ensemble attack success with variations in the amount of reference data available for the construction of DOMIAS features for both datasets. The size of the reference data used for the attack is along the x-axis. None indicates that the DOMIAS features are not used in MIA classification.} \label{fig:ensemble_ablation_domias}
\end{figure}

In the final set of ablations, the size of the reference dataset used to construct the DOMIAS features for both datasets is considered. By default, the attack uses all available reference points to construct such features. The results presented in Figure \ref{fig:ensemble_ablation_domias} examine using decreasing amounts of data for these computations. For both datasets, using smaller pools of reference data for the DOMIAS calculations does not lead to attack collapse, implying that massive pools of reference data are not required for the Ensemble attack to be effective.

\section{Privacy Heuristics: DCR, NNDR, HR, and EIR} \label{other_privacy_heuristics}

In this appendix, definitions of the auxiliary pseudo-privacy risk metrics and computation details are provided. When computing DCR metrics for ClavaDDPM, $\vert \mathcal{D}_h^i \vert = 10000$ for both the Berka and Diabetes datasets. This implies an idea DCR of $0.\overline{66}$ and $0.5$ for these datasets, respectively, except for settings when the training dataset size is modified. On the other hand, for TabDiff, $\vert \mathcal{D}_h^i \vert$ is augmented for both datasets such that the ideal DCR always remains $0.5$, even when training dataset size is varied. The average DCR across all target models is reported. The distance measure used for DCR is the Euclidean norm with categorical variables one-hot encoded. Beyond DCR, the metrics described below are only computed for ClavaDDPM.

\textbf{Nearest Neighbor Distance Ratio (NNDR)} is the ratio of the $\ell_2$ distance between each synthetic data point and its two nearest neighbors in the training data (closest vs.\ second-closest). This measure is intended to capture whether a synthetic sample lies in a sparsely populated region of the training distribution, with values closer to~0 indicating this. Using a holdout dataset, comprising real data points not used to train the generative model, we recompute the NNDR and compare the two ratios. The difference between the train-based and holdout-based ratios yields a ``privacy loss'' score. In our experiments we report 1-NNDR values so that, similar to the other metrics, the larger the value, the more the synthetic data may reveal about the original training set. 

\textbf{Hitting Rate (HR)} measures how often real data points are ``replicated'' in synthetic data, based on whether numerical values fall within a specified percentage (3\% in these experiments) of a real training data point's value range and categorical values match exactly; the hitting (exact match) rate is the percentage of real points that meet these criteria, and lower rates indicate better privacy protection. 

\begin{figure}[ht!]
\begin{subfigure}{0.33\textwidth}
\includegraphics[width=0.99\linewidth]{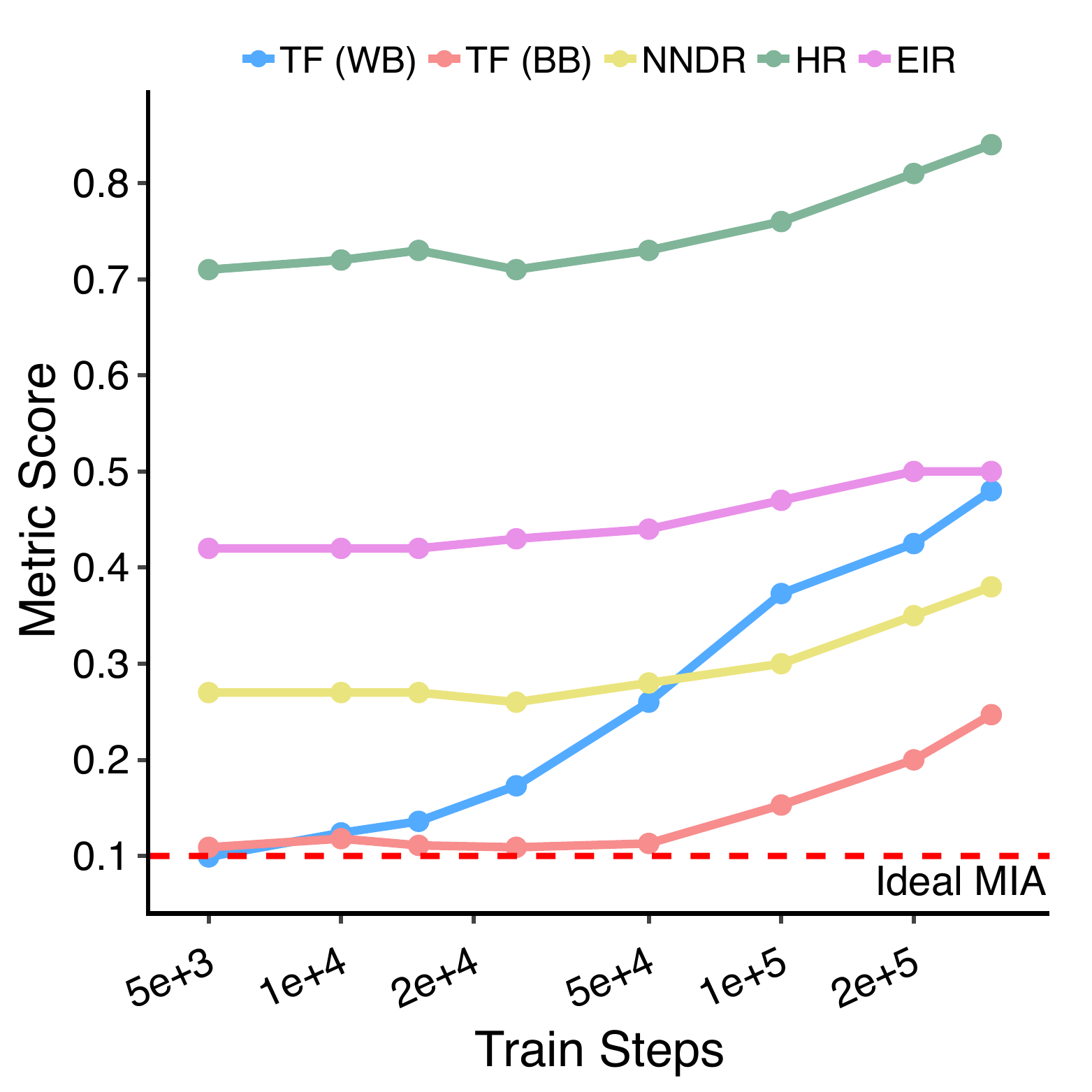}
\end{subfigure}
\begin{subfigure}{0.33\textwidth}
\includegraphics[width=0.99\linewidth]{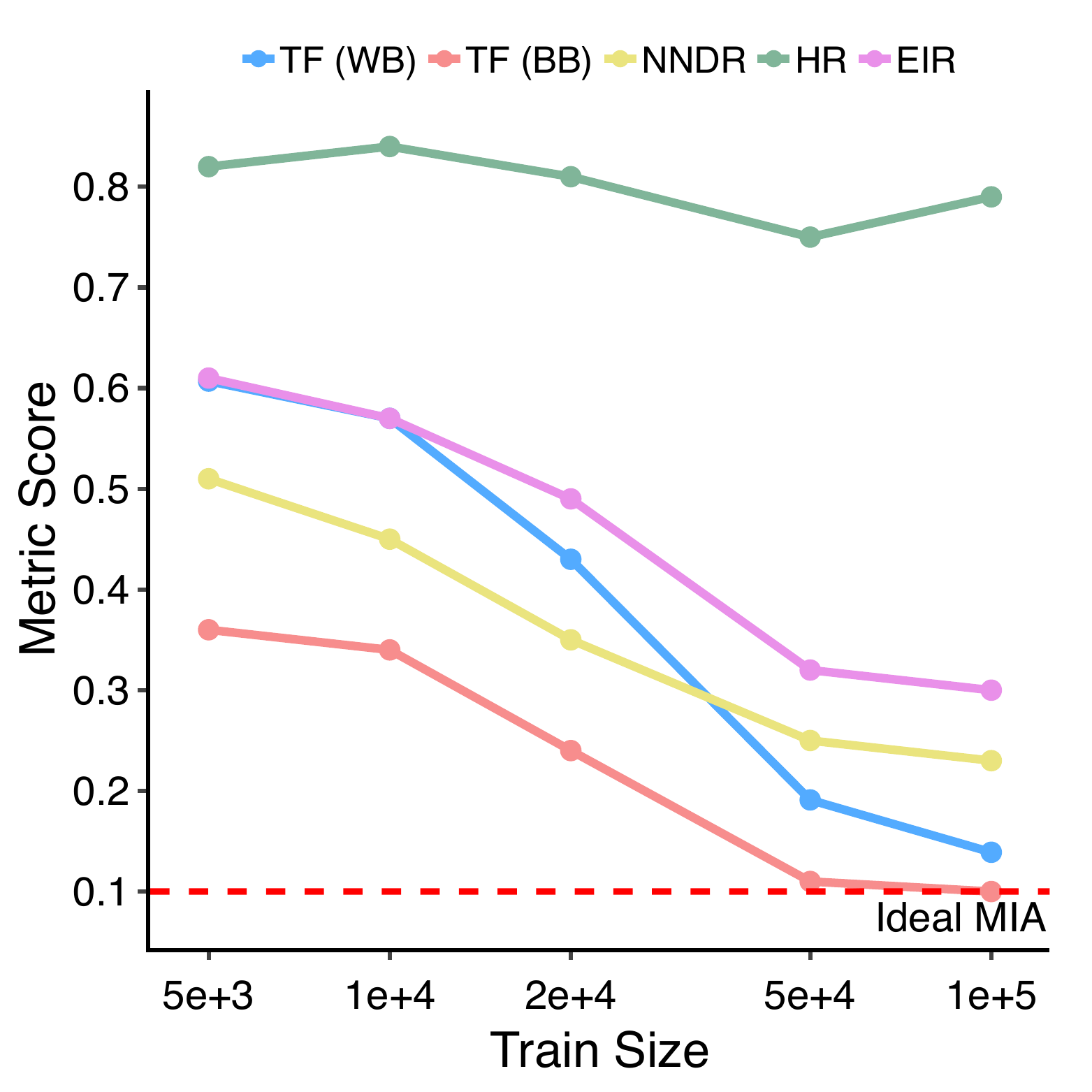}
\end{subfigure}
\begin{subfigure}{0.33\textwidth}
\includegraphics[width=0.99\linewidth]{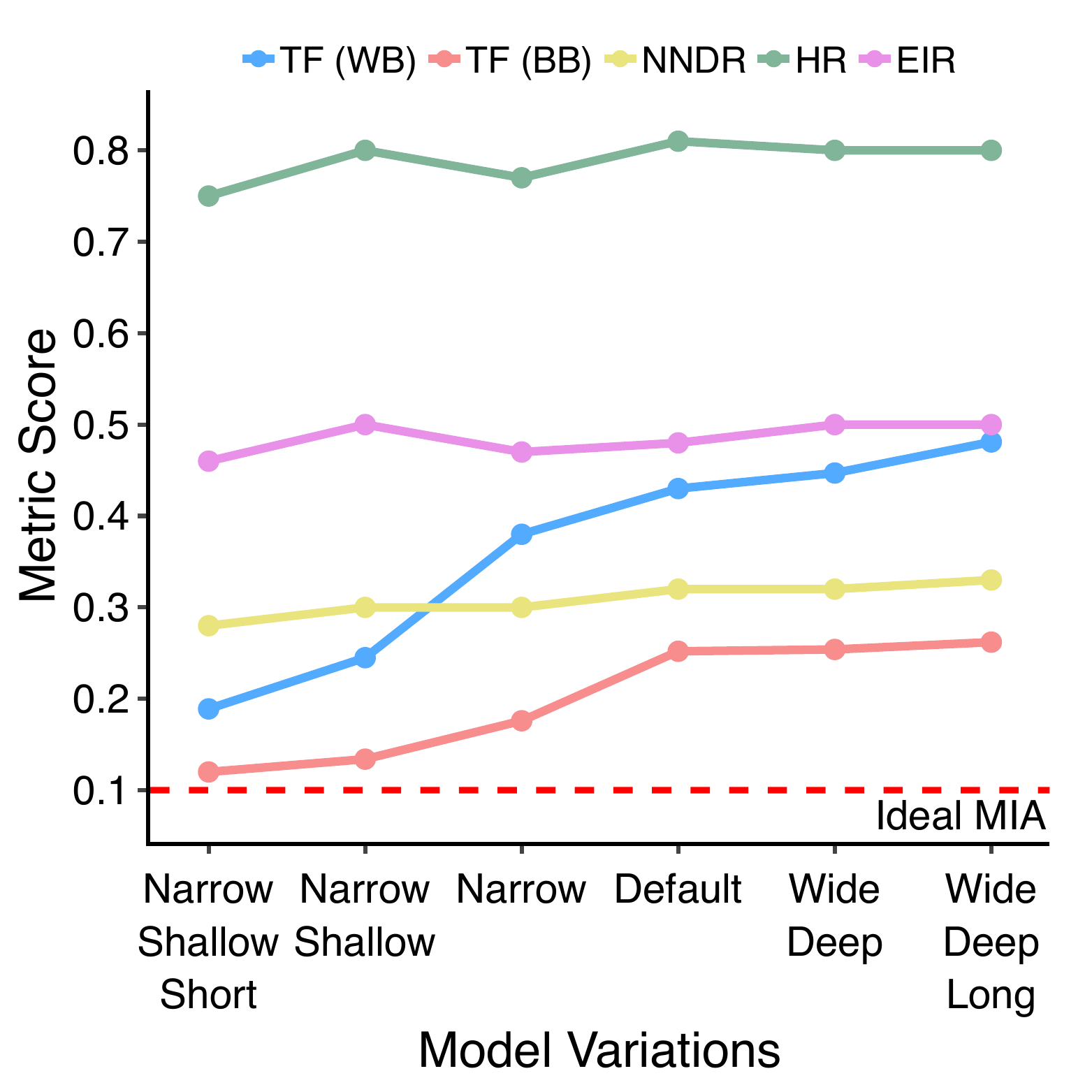}
\end{subfigure}
\begin{subfigure}{0.33\textwidth}
\includegraphics[width=0.99\linewidth]{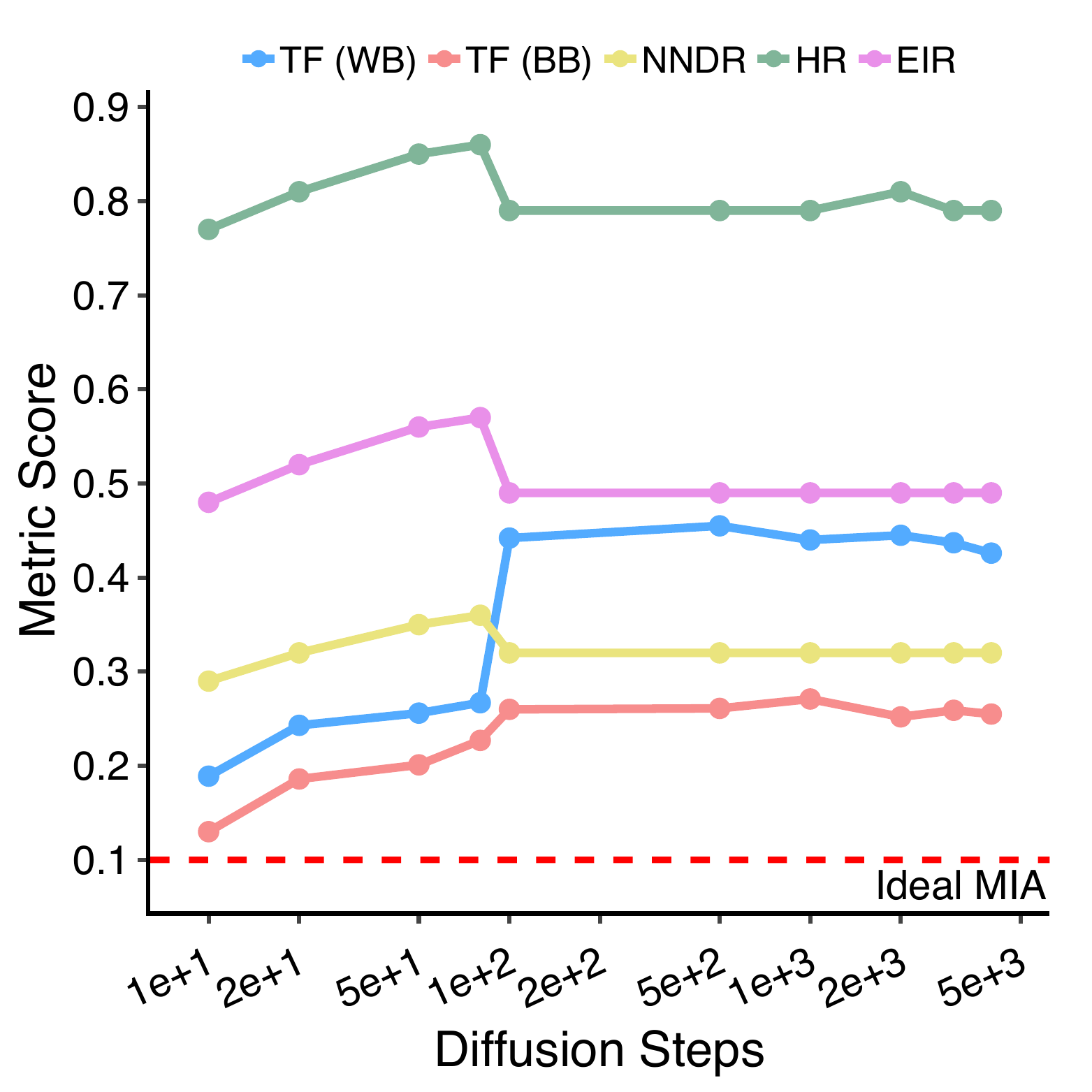}
\end{subfigure}
\begin{subfigure}{0.33\textwidth}
\includegraphics[width=0.99\linewidth]{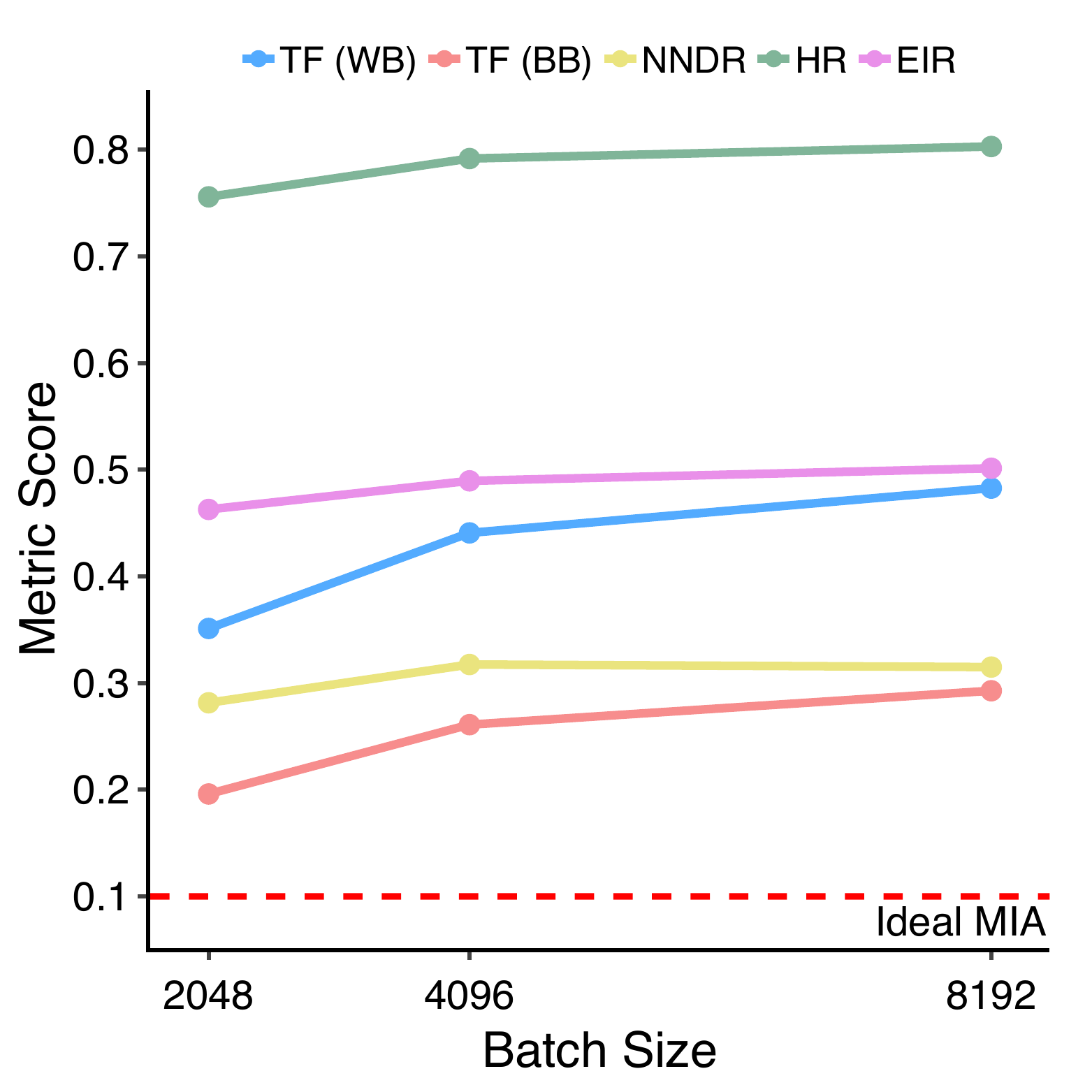}
\end{subfigure}
\begin{subfigure}{0.33\textwidth}
\includegraphics[width=0.99\linewidth]{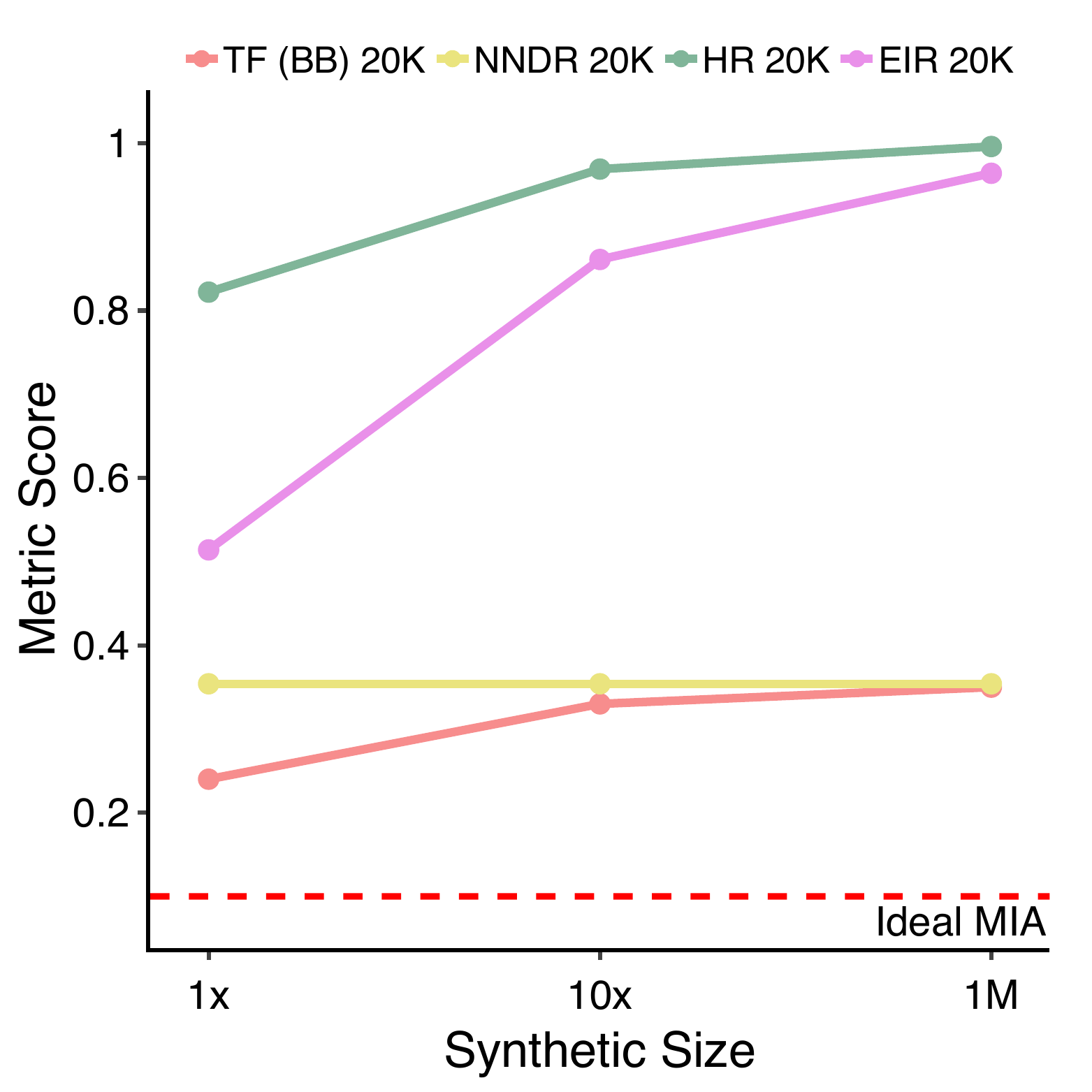}
\end{subfigure}
\caption{Training and synthesis levers that influence NNDR, HR, and EIR in comparison with TF MIA success for the Berka dataset.} \label{fig:berka_OtherPrivacy_training_variations}
\end{figure}

\begin{figure}[ht!]
\begin{subfigure}{0.33\textwidth}
\includegraphics[width=0.99\linewidth]{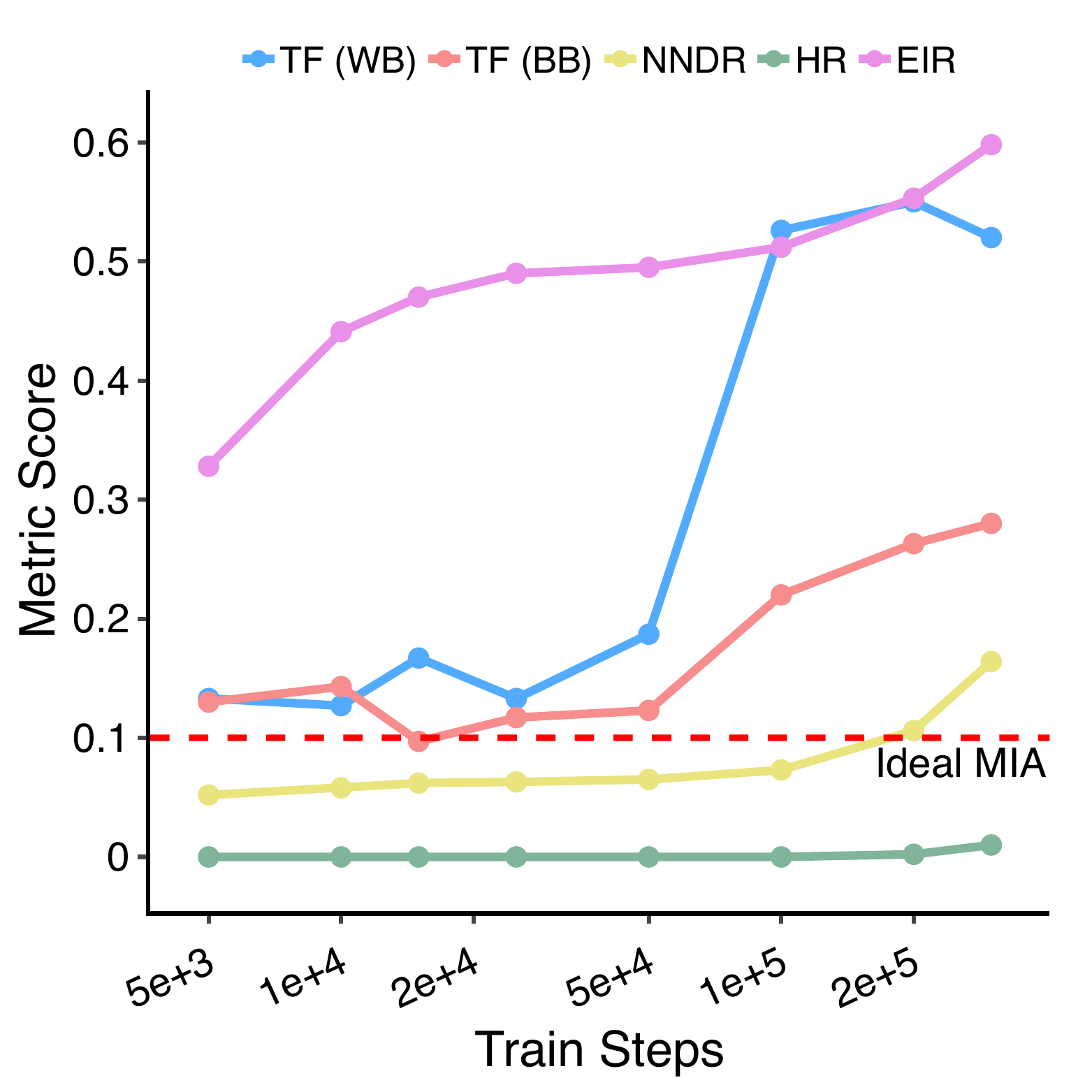}
\end{subfigure}
\begin{subfigure}{0.33\textwidth}
\includegraphics[width=0.99\linewidth]{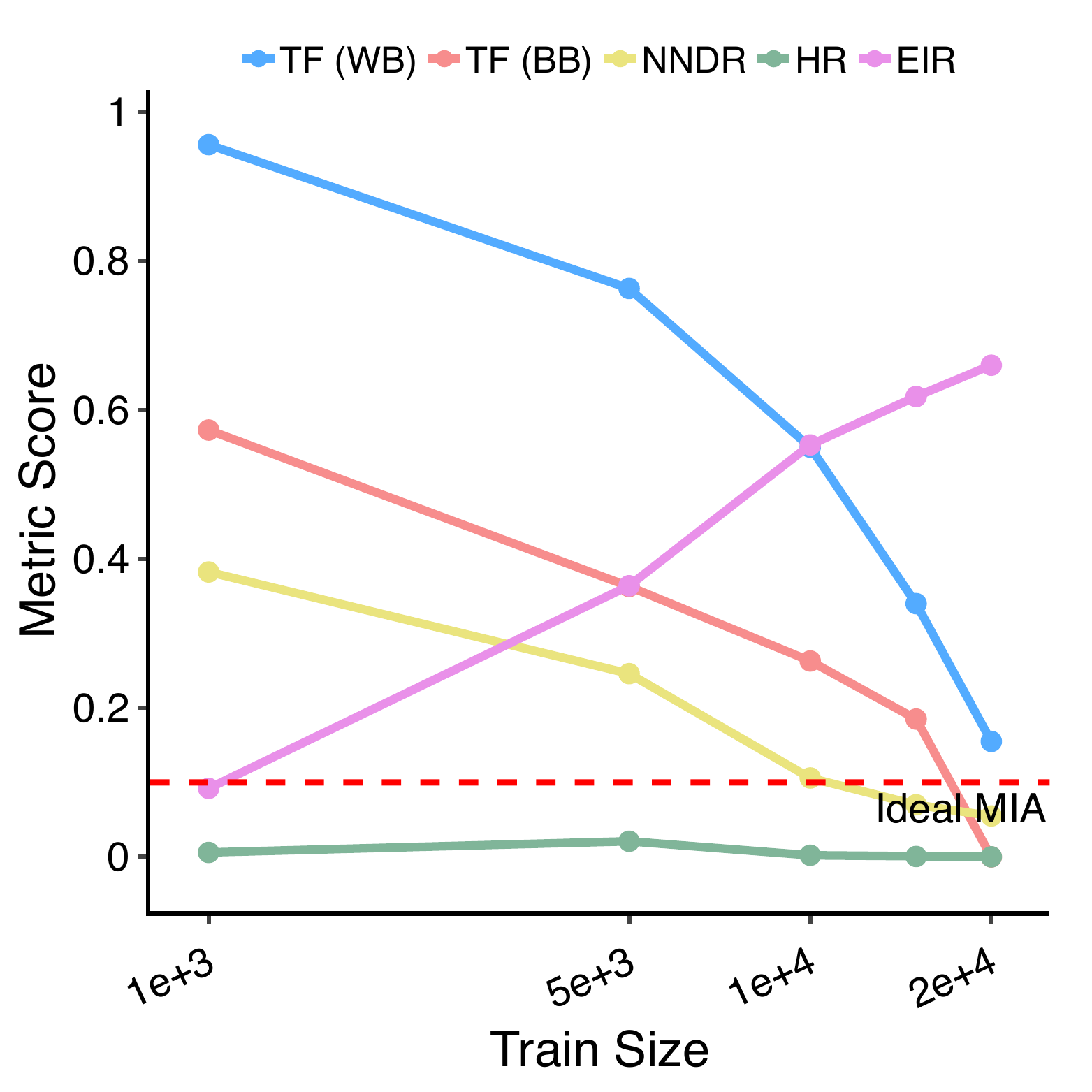}
\end{subfigure}
\begin{subfigure}{0.33\textwidth}
\includegraphics[width=0.99\linewidth]{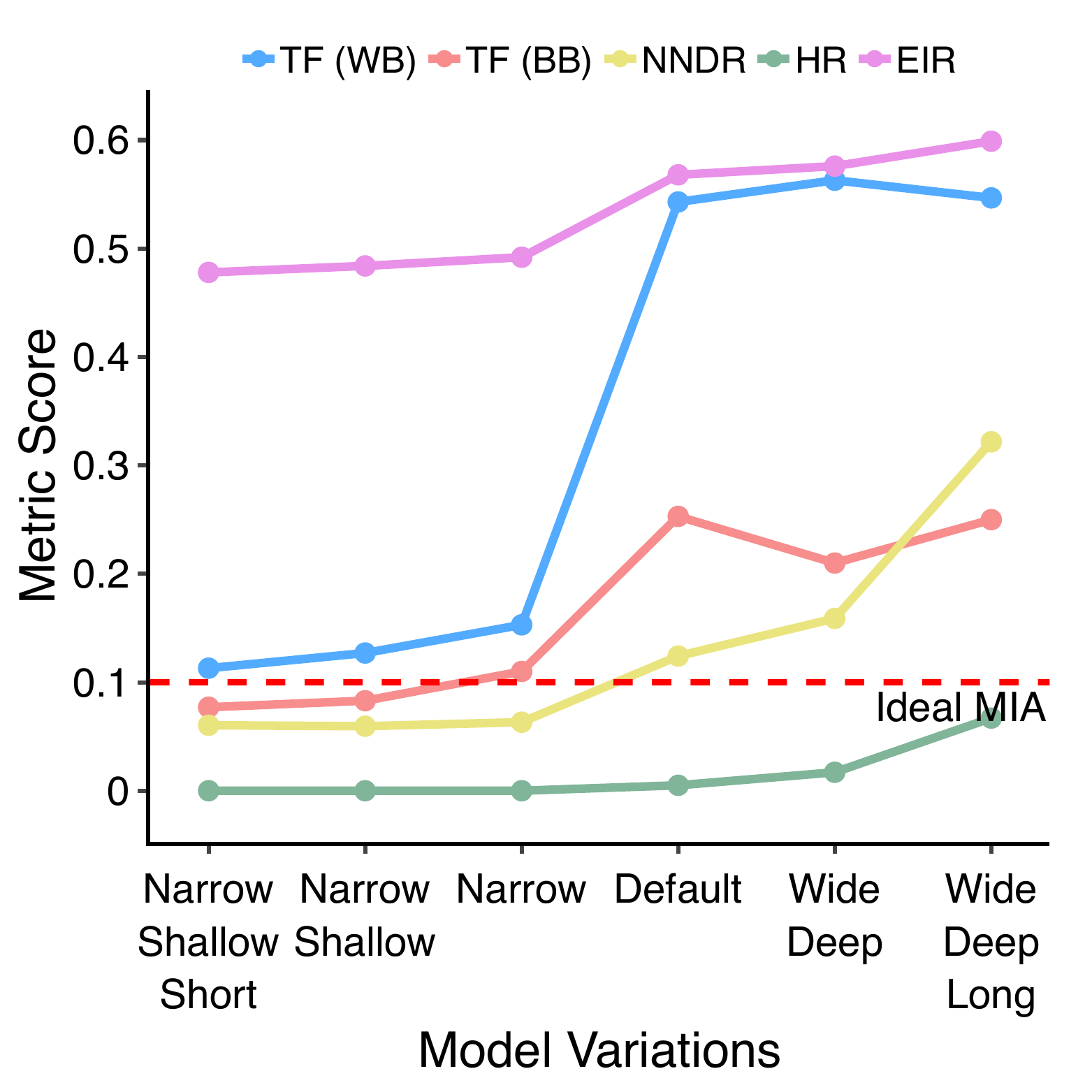}
\end{subfigure}
\begin{subfigure}{0.33\textwidth}
\includegraphics[width=0.99\linewidth]{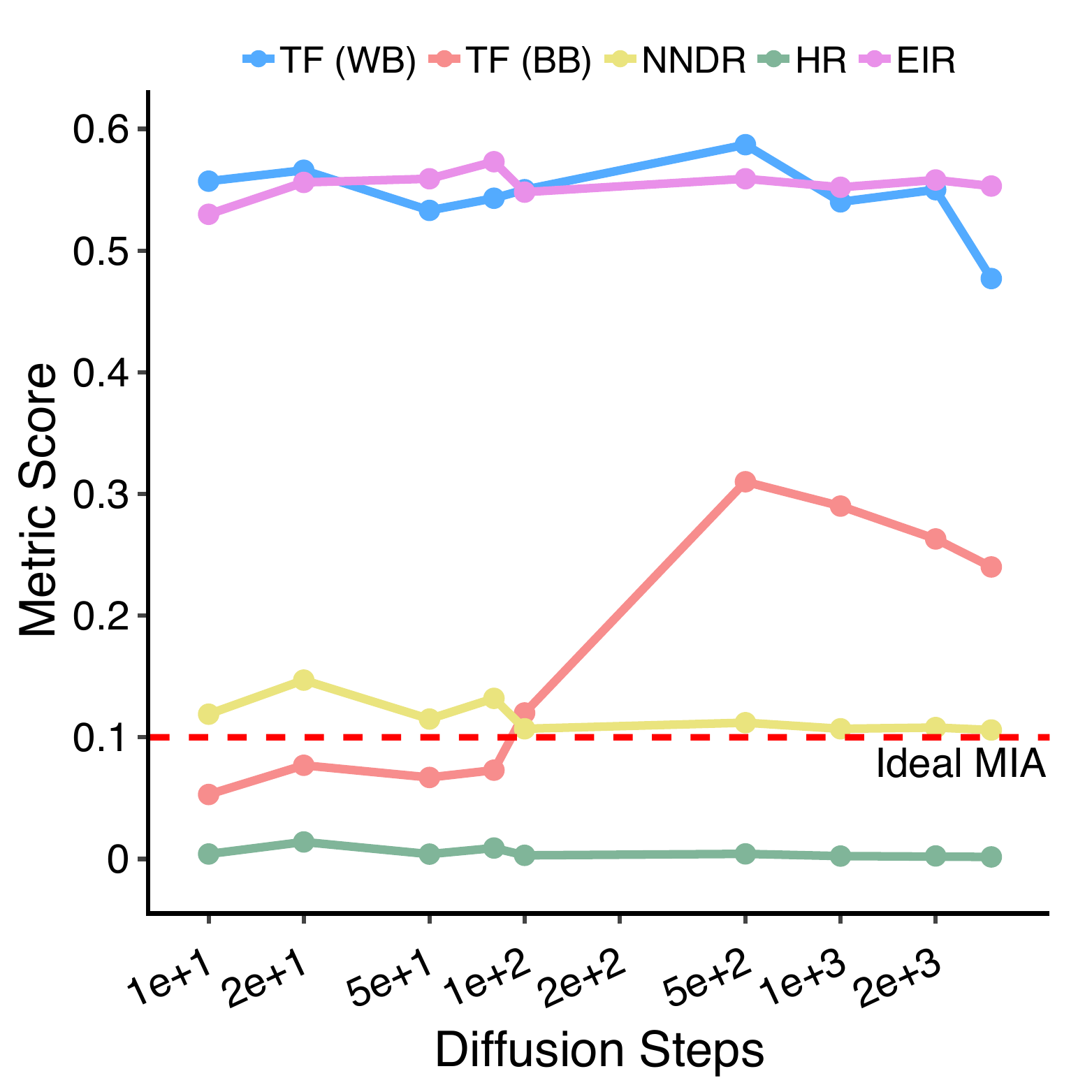}
\end{subfigure}
\begin{subfigure}{0.33\textwidth}
\includegraphics[width=0.99\linewidth]{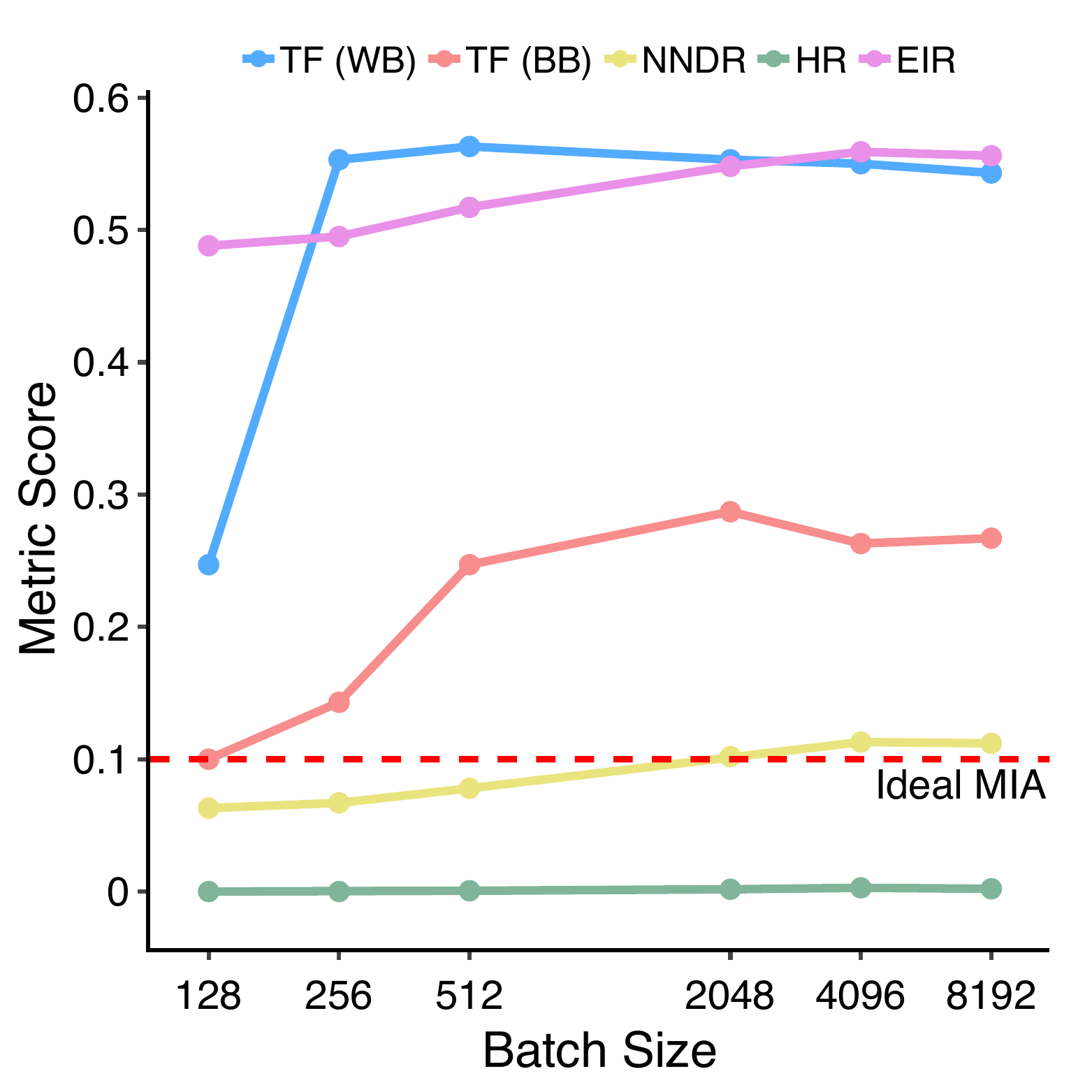}
\end{subfigure}
\begin{subfigure}{0.33\textwidth}
\includegraphics[width=0.99\linewidth]{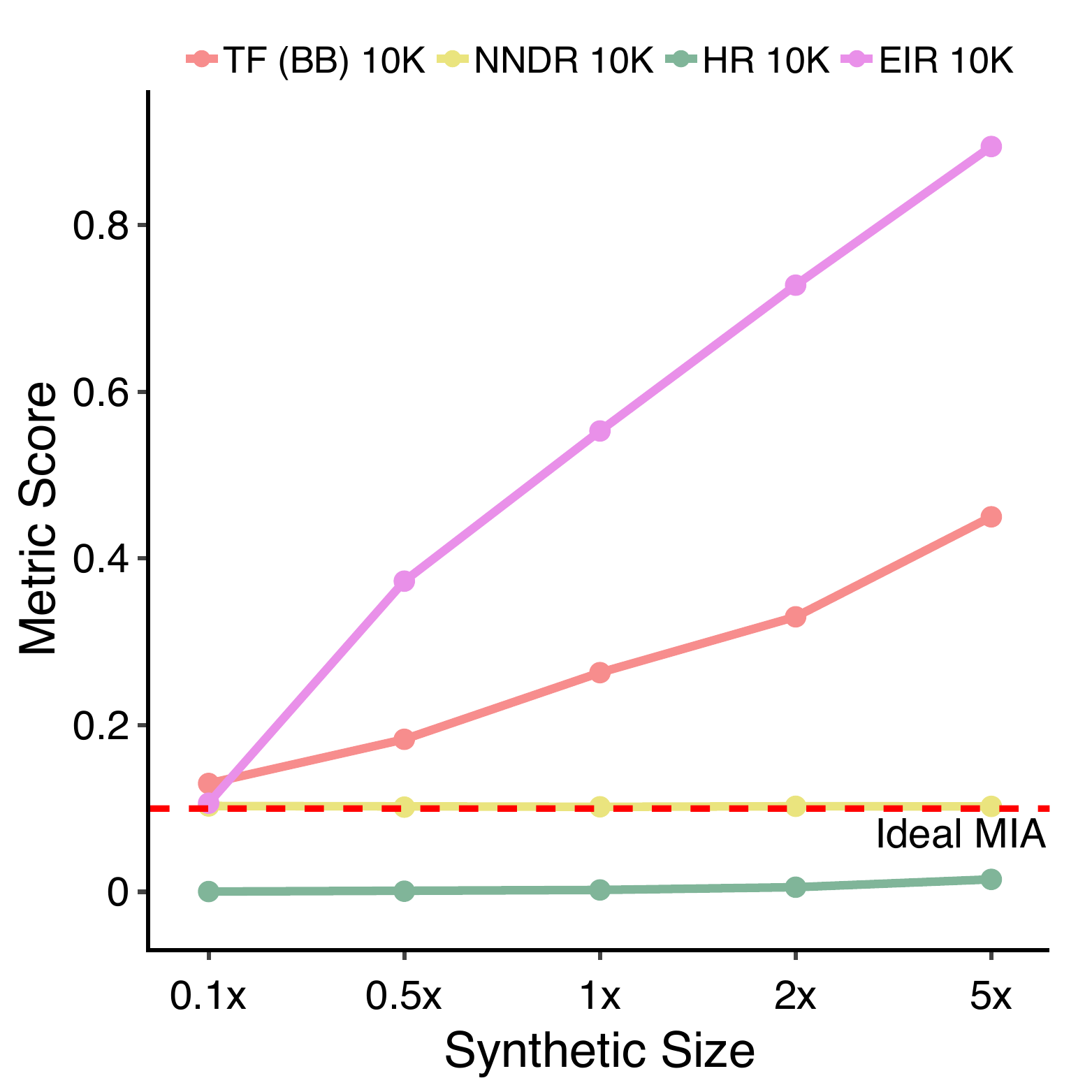}
\end{subfigure}
\caption{Training and synthesis levers that influence NNDR, HR, and EIR in comparison with TF MIA success for the Diabetes dataset.} \label{fig:diabetes_OtherPrivacy_training_variations}
\end{figure}

\textbf{Epsilon Identifiability Risk (EIR)} measures the percentage of training points for which a synthetic data point is closer than any other real point, indicating potential overfitting of the model that produced the synthetic data. Values closer to zero are meant to reflect lower privacy risk. When a holdout set is available, the same ratio is computed for holdout points, and the difference between the training and holdout ratios is reported, ideally near zero or negative showing that synthetic data is not memorizing training examples. The metric also weights features by their inverse entropy, giving greater influence to rare attributes.

Figures \ref{fig:berka_OtherPrivacy_training_variations} and \ref{fig:diabetes_OtherPrivacy_training_variations} show the effects of training and synthesis levers on NNDR, HR, and EIR privacy heuristics compared to ClavaDDPM TF MIA for the Berka and Diabetes datasets, respectively. Note that for, all metrics, lower values are meant to indicate lower privacy risk. The plots indicate that, similar to DCR, none of these privacy heuristics are reliable as a standalone metric for privacy leakage. Several observations support this argument. Some heuristics report high privacy risk even when the WB MIA indicates no privacy risk, as HR does in most of the experiments of Figure \ref{fig:berka_OtherPrivacy_training_variations}. Some follow a completely different pattern than the WB and BB MIA, as EIR does in the \textit{Train Size} experiment on the Diabetes dataset in Figure \ref{fig:berka_OtherPrivacy_training_variations}. Furthermore, some do not show any changes in privacy leakage with parameters variations unlike WB and BB MIA, as NNDR does in \textit{Batch Size} and \textit{Synthetic Size} experiments over the Diabetes dataset as well as HR in almost all experiments of Figure \ref{fig:diabetes_OtherPrivacy_training_variations}. 

\section{Quality Metrics and Results} \label{quality_metrics}

In this appendix, more information is provided on the quality metrics used in this work. Further, the quality metrics associated with the target models considered are reported. The metrics offer insight into the experimental results of Section \ref{main_results} in several ways. First, they showcase the relationship of synthetic-data quality to the various configuration changes and privacy risk. For example, increasing training iterations improves quality with diminishing returns while privacy risk continues to rise. Second, they demonstrate that the observed changes in MIA success are generally not tied to precipitous falloffs in synthetic quality. In most cases, generated data quality shows, at most, moderate and smooth variation with changes in the experimental configurations. 

When computing quality metrics, for model $i$, evaluation compares the synthetic data, $\mathcal{D}_s^i$, to the training data, $\mathcal{D}_t^i$, to measure how closely the synthetic data resembles the training data it is meant to mimic. For each setting below, five target models are chosen at random for quality measurement and the resulting metric values averaged. Note that, depending on the experimental configurations, $\vert \mathcal{D}_s^i \vert$ is not always equal to $\vert \mathcal{D}_t^i \vert$. As fidelity and diversity measures of synthetic tabular data, $\alpha$-precision and $\beta$-coverage (-recall) metrics from \citep{alaa2022faithful} are computed. The former measures how frequently generated samples fall within high-density regions of the real-data distribution. The latter measures the extent to which the synthetic distribution spans the diversity of the real data. To assess the extent to which generated data captures the statistical structure of the real data, differences in the one- and two-way marginals are computed. For numerical columns, a Kolmogorov-Smirnov (KS) test statistic is computed, comparing the synthetic and real dataset distributions for each column separately. Test statistics are then averaged over all columns. To compare real and synthetic categorical column distributions, total variation distance (TVD) is computed, also yielding a test statistics, which is averaged over all columns. In both cases, larger statistics imply wider differences in column-wise distributions. As a measure of column dependency preservation, pair-wise correlation and mutual information (MI) matrices are computed for synthetic and real tables. The Frobenius norm of the difference between these matrices is calculated. Correlation matrices are computed only for numerical column pairs, while MI matrices include both column types and are computed using SynthEval \citep{SynthEval2024}. Metrics are computed against holdout sets of size $20000$ and $10000$ for the Berka and Diabetes datasets, respectively, where required.

The final two metrics computed are so called machine learning efficacy (MLE) measures and only calculated for ClavaDDPM. These measures estimate how well synthetic data preserves dataset utility by training models on both synthetic and real data and comparing their task performance. Following \citep{Pang2025ClavaDDPM}, separate models are trained to predict each column value from the others. For numerical columns, a random forest (RF) regressor is trained and an RF classifier is constructed for categorical columns. For model $m_{\theta}^i$, MLE models are trained on either synthetic data generated by $m_{\theta}^i$ or the training data, $\mathcal{D}_t^i$. Regressor performance is measured via $R^2$ scores and classification with macro $F_1$. The $R^2$ or $F_1$ metrics are averaged across all columns in the respective categories and the difference in the averages is reported. For example, $\Delta F_1 = \text{avg}(F_1)_{\text{synthetic}} - \text{avg}(F_1)_{\text{real}}$. Larger values imply closer, or even improved performance, compared to real data. Reported values are for synthetic and training data from a single target model and RF performance is evaluated on training data from a separate target model as holdout data.

\begin{table}[ht!]
\caption{Quality metrics across training steps for ClavaDDPM models trained on Berka.}
\centering
\resizebox{0.99\linewidth}{!}{
\begin{tabular}{lcccccccccc}
\toprule
Train Steps
& 5e03 & 1e04 & 1.5e04  & 2.5e04  & 5e04  & 1e05 & 2e05 & 3e05 & 4e05 & 5e05 \\
\midrule
$\beta$ Cov. $\uparrow$ & 5.91e-01   & 5.78-e-01    & 5.91e-01   & 4.29e-01    & 4.51e-01    & 4.58e-01   & 4.79e-01   & 4.89e-01    & 5.04e-01   & 5.05e-01 \\
$\alpha$ Prec. $\uparrow$ & 9.31e-01    & 9.37e-01    & 9.33e-01    & 9.97e-01   & 9.94e-01    & 9.96e-01   & 9.93e-01  & 9.95e-01    & 9.96e-01    & 9.94e-01 \\
\midrule
Avg. KS $\downarrow$ & 7.02e-03  & 9.53e-03   & 8.48e-03   & 3.99e-02 & 4.03e-02   & 3.96e-02  & 3.90e-02  & 3.91e-02   & 3.91e-02   & 3.82e-02 \\
Avg. TVD $\downarrow$ & 1.28e-02   & 1.25e-02  & 1.35e-02  & 7.30e-03 & 9.30e-03    & 8.32e-03   & 5.82e-03  & 6.84e-03  & 4.85e-03   & 7.17e-03 \\
Corr. Diff. $\downarrow$ & 5.25e-02   & 3.81e-02   & 3.81e-02  & 3.09e-02    & 3.24e-02   & 3.35e-02   & 3.77e-02   & 2.36e-02   & 2.79e-02   & 4.32e-02 \\
MI Diff. $\downarrow$ & 4.18e-02   & 2.43e-02   & 1.19e-02   & 8.02e-03  & 4.11e-03 & 5.19e-03 & 3.58e-03 & 4.19e-03  & 4.44e-03 & 6.32e-03 \\
\midrule
$\Delta R^2$ $\uparrow$ & 6.20e-03       &  6.90e-03       &  5.80e-03     & 1.28e-02 & -6.29e-03         & -1.00e-03        & 2.40e-03    & -2.50e-03     & 8.00e-05 &  5.50e-05  \\
$\Delta F_1$ $\uparrow$ & -1.22e-02      &  -1.64e-02      & -1.05e-02   &  -1.53e-02      & -1.84e-02     &  -2.07e-02     & -1.92e-02 & -1.72e-02     & -1.72e-02 & -1.46e-02 \\
\bottomrule
\end{tabular}}
\label{berka_quality_vs_training_iteration}
\end{table}

\begin{table}[ht!]
\caption{Quality metrics across training dataset sizes for ClavaDDPM models trained on Berka.}
\centering
\begin{tabular}{lccccc}
\toprule
Training Size
& 5e03 & 1e04 & 2e04 & 5e04 & 1e05 \\
\midrule
$\beta$ Cov. $\uparrow$ & 6.06e-01    & 6.16e-01    & 5.44e-01    & 5.32e-01     & 3.37e-01 \\
$\alpha$ Prec. $\uparrow$ & 9.29e-01     & 9.29e-01     & 9.93e-01   & 9.31e-01     & 9.95e-01 \\
\midrule
Avg. KS $\downarrow$ & 4.24e-02   & 4.17e-02   & 3.90e-02   & 3.91e-02    & 3.81e-02 \\
Avg. TVD $\downarrow$ & 8.90e-03    & 8.55e-03    & 5.82e-03  & 5.75e-03    & 2.75e-03 \\
Corr. Diff. $\downarrow$  & 6.92e-02   & 5.01e-02   & 3.77e-02    & 1.32e-02    & 1.75e-02 \\
MI Diff. $\downarrow$ & 7.01e-03 & 7.06e-03   & 3.58e-03    & 4.98e-03    & 8.86e-03 \\
\midrule
$\Delta R^2$ $\uparrow$ & -3.90e-04      &  -3.27e-02        & -2.10e-04   &  -1.11e-01     &  -1.21e-01  \\
$\Delta F_1$ $\uparrow$ & -2.20e-03 & -8.60e-02    & -1.01e-02  &   -1.25e-02   &  -1.68e-02 \\
\bottomrule
\end{tabular}
\label{berka_quality_versus_training_size}
\end{table}

\begin{table}[ht!]
\caption{Quality metrics across different ClavaDDPM model configurations trained on the Berka dataset. Scenario (1) is ``narrow, shallow, short,'' (2) is ``narrow, shallow,'' (3) is ``narrow,'' (4) is default, (5) is ``wide, deep,'' (6) is ``wide, deep, long.''}
\centering
\begin{tabular}{lcccccc}
\toprule
Scenario & (1) & (2) & (3) & (4) & (5) & (6) \\
\midrule
$\beta$ Cov. $\uparrow$  &  5.95e-01 &     5.75e-01  & 6.06e-01  & 4.70e-01  & 6.13e-01  & 5.87e-01 \\
$\alpha$ Prec. $\uparrow$ & 9.30e-01 & 9.47e-01 & 9.30e-01 & 9.90e-01  & 9.30e-01  & 9.33e-01 \\
\midrule
Avg. KS $\downarrow$ & 7.47e-03 & 7.05e-03  & 8.99e-03 & 3.90e-03 & 6.21e-03 & 8.80e-03 \\
Avg. TVD $\downarrow$ & 8.60e-03 & 7.47e-03 & 7.15e-03 & 4.30e-03  & 5.94e-03 & 8.47e-03 \\
Corr. Diff. $\downarrow$ & 2.92e-02  & 2.49e-02 & 4.56e-02 & 5.64e-02 & 3.21e-02 & 3.27e-02 \\
MI Diff. $\downarrow$ & 1.32e-02 & 4.80e-03 & 5.92e-03 & 4.23e-03 & 1.12e-02 & 4.53e-03 \\
\midrule
$\Delta R^2$ $\uparrow$ &  -8.90e-03 &  -4.32e-03 & 1.70e-03 & 2.40e-03 & -3.40e-03 &  4.00e-03 \\
$\Delta F_1$ $\uparrow$ &  -1.16e-02 &  -5.40e-03 & 2.10e-03 & -3.70e-03 & 2.70e-03 &  4.10e-03 \\
\bottomrule
\end{tabular}
\label{berka_quality_vs_model_setup}
\end{table}

\subsection{ClavaDDPM: Berka Quality Results}

Synthetic data quality measures for target models trained on the Berka dataset are reported in Tables 
\ref{berka_quality_vs_training_iteration}--\ref{berka_quality_vs_batch_size}. The primary take away from these tables is that synthetic data quality does vary with training configuration changes, but it does not sharply degrade in any of the scenarios. As such, the main privacy risk results presented in Section \ref{main_results} are based on models with reasonable utility and reflect legitimate modeling-privacy trade-offs.

\begin{table}[ht!]
\caption{Quality metrics across diffusion timesteps for ClavaDDPM models trained on Berka.}
\centering
\resizebox{0.99\linewidth}{!}{
\begin{tabular}{lcccccccccc}
\toprule
Diffusion Steps & 10 & 20  & 50  & 80 & 100  & 500  & 1000  & 2000  & 3000 & 4000 \\
\midrule
$\beta$ Cov. $\uparrow$ & 5.20e-01 & 5.90e-01 & 6.13e-01 & 5.17e-01 & 4.94e-01 & 4.94e-01 & 4.97e-01 & 4.79e-01 & 4.98e-01 & 4.95e-01 \\
$\alpha$ Prec. $\uparrow$ & 9.30e-01 & 9.22e-01 & 9.28e-01 & 9.87e-01 & 9.90e-01 & 9.90e-01 & 9.90e-01 & 9.97e-01 & 9.90e-01 & 9.90e-01 \\
\midrule
Avg. KS $\downarrow$ & 2.63e-02 & 1.23e-02 & 8.55e-03 & 7.73e-03 & 7.38e-03 & 8.05e-01 & 7.80e-03 & 3.96e-03 & 7.67e-03 & 7.82e-03 \\
Avg. TVD $\downarrow$ & 5.90e-02 & 2.55e-02 & 1.07e-02 & 9.52e-03 & 6.92e-03 & 7.47e-03 & 4.77e-03 & 4.35e-03 & 5.45e-03 & 7.12e-03 \\
Corr. Diff. $\downarrow$& 1.46e-01 & 7.12e-02 & 3.23e-02 & 2.98e-02 & 4.64e-02 & 3.87e-02 & 3.32e-02 & 5.64e-02 & 3.44e-02 & 4.02e-02 \\
MI Diff. $\downarrow$ & 1.24e-01 & 4.32e-02 & 1.33e-02 & 7.29e-03 & 7.66e-03 & 9.13e-03 & 1.64e-02 & 4.23e-03 & 6.24e-03 & 8.20e-03 \\
\midrule
$\Delta R^2$ $\uparrow$ & -1.48e-02 & -1.23e-02 & -6.40e-03 & -1.51e-03 & -1.30e-03 & -6.30e-03 & 7.10e-03 & 2.40e-03 & 8.90e-03 & 7.90e-03 \\
$\Delta F_1$ $\uparrow$ & -2.33e-02 & -1.57e-02 & -1.49e-02 & -1.82e-02 & -5.30e-03 & -6.60e-03 & -8.00e-03 & -4.80e-03 & 2.20e-03 & -4.40e-03 \\
\bottomrule
\end{tabular}}
\label{berka_quality_vs_timesteps}
\end{table}

\begin{table}[ht!]
\caption{Quality metrics across batch sizes for ClavaDDPM models trained on Berka.}
\centering
\begin{tabular}{lccc}
\toprule
Batch Size & 2048 & 4096  & 8192 \\
\midrule
$\beta$ Cov. $\uparrow$ &  4.90e-01 & 4.90e-01 & 4.80e-01 \\
$\alpha$ Prec. $\uparrow$ &  9.90e-01 & 9.90e-01 & 9.90e-01 \\
\midrule
Avg. KS $\downarrow$ &  7.40e-03 & 7.60e-03 & 8.10e-03 \\
Avg. TVD $\downarrow$ &  4.90e-03 & 4.70e-03 & 6.80e-03 \\
Corr. Diff. $\downarrow$ &  4.44e-02 & 2.90e-02 & 4.02e-02 \\
MI Diff. $\downarrow$ &  4.30e-03 & 3.90e-03 & 6.10e-03 \\
\midrule
$\Delta R^2$ $\uparrow$ & 2.30e-03 & 2.40e-03 & -5.40e-03 \\
$\Delta F_1$ $\uparrow$ & -1.22e-02 & -1.17e-02 & -1.43e-02 \\
\bottomrule
\end{tabular}
\label{berka_quality_vs_batch_size}
\end{table}

For the Berka dataset, synthetic quality is impacted by training choices in different ways. In the ranges tested, some parameter changes have fairly limited impact on quality, such as batch size or model architecture changes. Others, like training iterations and diffusion steps, generally improve the quality metrics with larger values. Finally, training dataset size has a more nuanced impact on quality, where some metrics improve with more data and some improve to a point and then degrade.

\begin{table}[ht!]
\caption{Quality metrics across training steps for ClavaDDPM models trained on Diabetes.}
\centering
\resizebox{0.99\linewidth}{!}{
\begin{tabular}{lcccccccccc}
\toprule
Train Steps
& 5e03 & 1e04 & 1.5e04  & 2.5e04  & 5e04  & 1e05 & 2e05 & 3e05 & 4e05 & 5e05 \\
\midrule
$\beta$ Cov. $\uparrow$ & 3.19e-01 & 4.28e-01 & 4.61e-01 & 4.89e-01 & 5.30e-01 & 6.01e-01 & 6.60e-01 & 6.93e-01 & 6.61e-01 & 6.73e-01 \\
$\alpha$ Prec. $\uparrow$ & 8.00e-01 & 8.09e-01 & 7.77e-01 & 8.19e-01 & 8.33e-01 & 7.81e-01 & 7.53e-01 & 7.77e-01 & 7.58e-01 & 7.63e-01 \\
\midrule
Avg. KS $\downarrow$ & 1.93e-01 & 8.00e-02 & 4.30e-02 & 2.10e-02 & 2.00e-02 & 3.60e-02 & 3.31e-02 & 3.10e-02 & 3.50e-02 & 3.20e-02 \\
Avg. TVD $\downarrow$ & 1.75e-01 & 8.10e-02 & 5.10e-02 & 3.20e-02 & 2.60e-02 & 3.30e-02 & 3.64e-02 & 3.20e-02 & 3.60e-02 & 3.50e-02 \\
Corr. Diff. $\downarrow$ & 2.11e-00 & 1.66e+00 & 1.25e+00 & 5.33e-01 & 3.44e-01 & 3.25e-01 & 1.42e-01 & 9.40e-02 & 1.41e-01 & 1.23e-01 \\
MI Diff. $\downarrow$ & 7.27e-00 & 5.78e+00 & 4.40e+00 & 2.81e+00 & 5.85e+00 & 4.31e+00 & 5.84e+00 & 5.84e+00 & 7.35e+00 & 4.07e+00 \\
\midrule
$\Delta R^2$ $\uparrow$ & -5.77e-01 & -5.40e-02 & -2.70e-02 & -3.10e-02 & 1.90e-02 & -1.60e-02 & -1.78e-01 & -1.30e-02 & -2.70e-02 & 1.60e-02 \\
$\Delta F_1$ $\uparrow$ & -4.00e-02 & -2.80e-02 & -2.40e-02 & -1.80e-02 & -1.80e-02 & -1.60e-02 & -4.10e-02 & -1.00e-02 & -3.00e-03 & -9.00e-03 \\
\bottomrule
\end{tabular}}
\label{diabetes_quality_vs_training_iteration}
\end{table}

\subsection{ClavaDDPM: Diabetes Quality Results}

Tables \ref{diabetes_quality_vs_training_iteration}--\ref{diabetes_quality_vs_batch_size} report the quality metrics for variations in the target model training setups on the Diabetes dataset. As discussed in Section \ref{datasets_and_defaults}, the Diabetes dataset is smaller, in total size, than Berka, but has a larger number of columns. As such, there is more variation in the target model quality results with changes to the experimental configurations. It should be noted that, with the increased number of columns, the correlation and MI matrices are also much larger, thereby inducing larger Froebenius norms for similar quality statistics.

\begin{table}[ht!]
\caption{Quality metrics across training data sizes for ClavaDDPM models trained on Diabetes.}
\centering
\begin{tabular}{lccccc}
\toprule
Training Size
& 1e03 & 5e03 & 1e04 & 1.5e04 & 2e04 \\
\midrule
$\beta$ Cov. $\uparrow$ & 1.53e-01 & 4.95e-01 & 6.60e-01 & 6.93e-01 & 7.12e-01 \\
$\alpha$ Prec. $\uparrow$ & 7.22e-01 & 7.16e-01 & 7.53e-01 & 7.81e-01 & 8.29e-01 \\
\midrule
Avg. KS $\downarrow$ & 4.80e-02 & 4.29e-02 & 3.31e-02 & 3.20e-02 & 2.90e-02 \\
Avg. TVD $\downarrow$ & 5.10e-02 & 4.40e-02 & 3.64e-02 & 3.20e-02 & 2.90e-02 \\
Corr. Diff. $\downarrow$  & 4.27e-01 & 2.71e-01 & 1.42e-01 & 1.54e-01 & 1.43e-01 \\
MI Diff. $\downarrow$ & 8.84e+00 & 8.72e+00 & 5.84e+00 & 5.84e+00 & 5.31e+00 \\
\midrule
$\Delta R^2$ $\uparrow$ & -8.80e-02 & -3.60e-02 & -1.78e-01 & 1.00e-02 & 4.10e-02 \\
$\Delta F_1$ $\uparrow$ & -2.60e-02 & -1.25e-02 & -4.10e-02 & -5.00e-03 & 1.00e-03 \\
\bottomrule
\end{tabular}
\label{diabetes_quality_versus_training_size}
\end{table}

\begin{table}[ht!]
\caption{Quality metrics across different ClavaDDPM model configurations trained on the Diabetes dataset. Scenario (1) is ``narrow, shallow, short,'' (2) is ``narrow, shallow,'' (3) is ``narrow,'' (4) is default, (5) is ``wide, deep,'' (6) is ``wide, deep, long.''}
\centering
\begin{tabular}{lcccccc}
\toprule
Scenario & (1) & (2) & (3) & (4) & (5) & (6) \\
\midrule
$\beta$ Cov. $\uparrow$  & 4.73e-01 & 4.78e-01 & 4.96e-01 & 6.85e-01 & 6.81e-01 & 7.42e-01 \\
$\alpha$ Prec. $\uparrow$ & 8.18e-01 & 8.51e-01 & 8.55e-01 & 7.66e-01 & 7.63e-01 & 7.74e-01 \\
\midrule
Avg. KS $\downarrow$ & 3.80e-02 & 4.00e-02 & 2.60e-02 & 3.50e-02 & 3.60e-02 & 3.20e-02 \\
Avg. TVD $\downarrow$ & 4.10e-02 & 4.30e-02 & 2.70e-02 & 3.50e-02 & 3.60e-02 & 3.30e-02 \\
Corr. Diff. $\downarrow$ & 1.14e+00 & 1.54e+00 & 8.87e-01 & 3.27e-01 & 3.11e-01 & 1.43e-01 \\
MI Diff. $\downarrow$ & 4.95e+00 & 5.16e+00 & 3.20e+00 & 4.01e+00 & 4.01e+00 & 5.84e+00 \\
\midrule
$\Delta R^2$ $\uparrow$ & -3.10e-02 & -6.40e-02 & -1.90e-02 & -2.30e-02 & -1.90e-02 & -1.08e-01 \\
$\Delta F_1$ $\uparrow$ & -3.10e-02 & -2.10e-02 & -1.50e-02 & -8.00e-03 & -1.30e-02 & -1.60e-02 \\
\bottomrule
\end{tabular}
\label{diabetes_quality_vs_model_setup}
\end{table}

\begin{table}[ht!]
\caption{Quality metrics across diffusion timesteps for ClavaDDPM models trained on Diabetes.}
\centering
\resizebox{0.99\linewidth}{!}{
\begin{tabular}{lccccccccc}
\toprule
Diffusion Steps & 10 & 20  & 50  & 80 & 100  & 500  & 1000  & 2000  & 3000 \\
\midrule
$\beta$ Cov. $\uparrow$ & 6.55e-01 & 6.79e-01 & 6.67e-01 & 6.63e-01 & 6.62e-01 & 6.55e-01 & 6.59e-01 & 6.68e-01 & 6.66e-01 \\
$\alpha$ Prec. $\uparrow$ & 6.52e-01 & 7.04e-01 & 7.44e-01 & 7.45e-01 & 7.25e-01 & 7.58e-01 & 7.59e-01 & 7.52e-01 & 7.72e-01 \\
\midrule
Avg. KS $\downarrow$ & 4.90e-02 & 4.20e-02 & 3.80e-02 & 3.60e-02 & 3.70e-02 & 3.60e-02 & 3.70e-02 & 3.50e-02 & 3.40e-02 \\
Avg. TVD $\downarrow$ & 4.90e-02 & 4.20e-02 & 3.80e-02 & 3.60e-02 & 3.80e-02 & 3.70e-02 & 3.70e-02 & 3.60e-02 & 3.40e-02 \\
Corr. Diff. $\downarrow$ & 1.52e-01 & 1.18e-01 & 1.35e-01 & 1.18e-01 & 1.86e-01 & 1.27e-01 & 1.38e-01 & 1.49e-01 & 1.24e-01 \\
MI Diff. $\downarrow$ & 7.36e+00 & 7.35e+00 & 5.84e+00 & 4.01e+00 & 7.00e-03 & 7.35e+00 & 5.84e+00 & 4.01e+00 & 5.84e+00 \\
\midrule
$\Delta R^2$ $\uparrow$ & -1.50e-02 & -3.40e-02 & -4.60e-02 & -1.70e-02 & -1.80e-02 & -1.60e-02 & -2.90e-02 & -2.60e-02 & -2.50e-02 \\
$\Delta F_1$ $\uparrow$ & -9.00e-03 & -9.00e-03 & -1.10e-02 & -8.00e-03 & -1.70e-02 & -1.10e-02 & -1.60e-02 & -9.00e-03 & -1.00e-02 \\
\bottomrule
\end{tabular}}
\label{diabetes_quality_vs_timesteps}
\end{table}

Similar to the Berka results, generated data quality does not collapse across a wide variety of training setups. However, there are a few instances, at the extreme ends of the hyperparameter spectra, where metrics indicate larger deterioration. For instance, when training for only a few thousand steps in Table \ref{diabetes_quality_vs_training_iteration}, many of the metrics are significantly worse. Similarly very small training sizes also produce low $\beta$-coverage and large correlation differences.

\begin{table}[ht!]
\caption{Quality metrics across batch sizes for ClavaDDPM models trained on Diabetes.}
\centering
\begin{tabular}{lcccccc}
\toprule
Batch Size & 128 & 256 & 512 & 2048 & 4096  & 8192 \\
\midrule
$\beta$ Cov. $\uparrow$ & 4.87e-01 & 5.65e-01 & 6.02e-01 & 6.62e-01 & 6.71e-01 & 6.64e-01 \\
$\alpha$ Prec. $\uparrow$ & 8.67e-01 & 8.06e-01 & 7.68e-01 & 7.49e-01 & 7.70e-01 & 7.64e-01 \\
\midrule
Avg. KS $\downarrow$ & 3.60e-02 & 2.90e-02 & 3.62e-02 & 3.60e-02 & 3.40e-02 & 3.50e-02 \\
Avg. TVD $\downarrow$ & 3.90e-02 & 2.90e-02 & 3.62e-02 & 3.60e-02 & 3.40e-02 & 3.50e-02 \\
Corr. Diff. $\downarrow$ & 6.99e-01 & 2.32e-01 & 2.15e-01 & 1.60e-01 & 1.40e-01 & 1.06e-01 \\
MI Diff. $\downarrow$ & 3.45e+00 & 5.84e+00 & 2.66e+00 & 5.84e+00 & 5.84e+00 & 5.84e+00 \\
\midrule
$\Delta R^2$ $\uparrow$ & -1.54e-02 & -6.00e-03 & -1.56e-02 & -2.50e-02 & -2.00e-02 & -6.00e-03 \\
$\Delta F_1$ $\uparrow$ & -3.90e-02 & -1.30e-02 & -1.11e-02 & -1.30e-02 & -8.00e-03 & -1.30e-02 \\
\bottomrule
\end{tabular}
\label{diabetes_quality_vs_batch_size}
\end{table}

The dependencies of the quality metrics for target models trained on the Diabetes dataset have some similarities and notable differences compared to the Berka-trained models. Quality metrics are inconsistently affected by changes in batch size and increasing training steps generally improves such measurements. On the other hand, for the Diabetes dataset, larger models and training dataset sizes mostly improve synthetic quality, while the impact of changing the number of diffusion steps is varied.

\begin{table}[ht!]
\centering
\caption{Synthetic data quality metrics across experimental settings for target TabDiff models trained on Berka.}
\resizebox{0.99\linewidth}{!}{
\begin{tabular}{llcccccc}
\toprule
Category & Setting
& $\beta$ Cov.$\uparrow$
& $\alpha$ Prec.$\uparrow$
& Avg. KS$\downarrow$
& Avg. TVD$\downarrow$
& Corr. Diff.$\downarrow$
& MI Diff.$\downarrow$ \\
\midrule
Train Steps
& 5e03  & 0.592 & 0.914 & 0.020 & 0.014 & 0.147 & 0.107 \\
& 1e04  & 0.590 & 0.925 & 0.012 & 0.010 & 0.091 & 0.101 \\
& 2e04  & 0.519 & 0.953 & 0.009 & 0.007 & 0.053 & 0.090 \\
& 4e04  & 0.534 & 0.957 & 0.010 & 0.008 & 0.057 & 0.099 \\
& 6e04 & 0.583 & 0.926 & 0.010 & 0.007 & 0.057 & 0.097 \\
& 8e04 & 0.580 & 0.934 & 0.009 & 0.007 & 0.064 & 0.094 \\
& 1e05 & 0.587 & 0.944 & 0.009 & 0.006 & 0.059 & 0.096 \\
\midrule
Architectures
& 512x1  & 0.544 & 0.950 & 0.009 & 0.007 & 0.071 & 0.101 \\
& 1024x1 & 0.562 & 0.941 & 0.009 & 0.006 & 0.061 & 0.103 \\
& 2048x1 & 0.558 & 0.936 & 0.009 & 0.013 & 0.061 & 0.098 \\
& 512x2  & 0.584 & 0.933 & 0.008 & 0.006 & 0.063 & 0.095 \\
& 1024x2 & 0.587 & 0.944 & 0.009 & 0.006 & 0.059 & 0.096 \\
& 2048x2 & 0.525 & 0.938 & 0.006 & 0.008 & 0.048 & 0.089 \\
& 512x3  & 0.567 & 0.935 & 0.009 & 0.006 & 0.062 & 0.095 \\
& 1024x3 & 0.589 & 0.927 & 0.009 & 0.008 & 0.064 & 0.099 \\
& 2048x3 & 0.525 & 0.951 & 0.008 & 0.005 & 0.067 & 0.097 \\
\midrule
Diffusion Steps
& 5   & 0.360 & 0.728 & 0.205 & 0.054 & 1.155 & 1.003 \\
& 50  & 0.587 & 0.944 & 0.009 & 0.006 & 0.059 & 0.096 \\
& 300 & 0.515 & 0.965 & 0.009 & 0.005 & 0.045 & 0.024 \\
\midrule
Train Size
& 2e04  & 0.587 & 0.944 & 0.009 & 0.006 & 0.059 & 0.096 \\
(Equal Synth.)& 1e05  & 0.579 & 0.933 & 0.038 & 0.005 & 0.064 & 0.091 \\
& 2e05 & 0.526 & 0.944 & 0.037 & 0.005 & 0.069 & 0.090 \\
\midrule
Batch Size
& 2048 & 0.566 & 0.925 & 0.009 & 0.008 & 0.060 & 0.097 \\
& 4096 & 0.587 & 0.944 & 0.009 & 0.006 & 0.059 & 0.096 \\
& 8192 & 0.588 & 0.928 & 0.009 & 0.007 & 0.080 & 0.098 \\
\midrule
Synth. Size
& 2e04  & 0.547 & 0.944 & 0.009 & 0.006 & 0.059 & 0.096 \\
(2e04 Train) & 1e05 & 0.585 & 0.927 & 0.007 & 0.005 & 0.056 & 0.100 \\
& 2e05 & 0.564 & 0.938 & 0.006 & 0.004 & 0.052 & 0.097 \\
\bottomrule
\end{tabular}}
\label{tabdiff_berka_quality}
\end{table}

\subsection{TabDiff: Berka Quality Results}

Quality metrics for TabDiff, trained on the Berka dataset, for the various levers investigated in Sections \ref{steps_size_model} and \ref{diffusion_batch_ratio} are displayed in Table \ref{tabdiff_berka_quality}. Overall, the quality metrics are generally stable across the range of hyperparameters and alternative settings investigated. At the extremes, there are some signs of model quality degradation. For example, when training for only $5000$ steps, $\alpha$ precision, Avg. KS, and Corr. Diff begin to show signs of stress. Similarly, using a very small number of diffusion steps reduces synthetic data quality along a number of metrics. Even still, the attack shows non-trivial success for this dataset. Several takeaways are shared with the results for ClavaDDPM. Training models for longer improves quality to a point. However, beyond that, privacy leakage significantly scales without in-kind quality benefits. More training data does not uniformly improve all quality metrics, but it does markedly reduce privacy risk. Finally, larger models can produce some quality improvements, but also increased leakage. Measuring quality, along with privacy, allows for informed tradeoff decisions by the data modeler.

An important observation from the quality metrics is that synthetic data quality for Berka does not vary significantly with batch size. There are some variations with batch size changes for Diabetes, but they are not consistent in the sense that most of the ``best'' metric values occur at the default batch size of $4096$ rather than on either end of the batch size spectrum. This is interesting due to the observed differences in MIA success rate between ClavaDDPM and TabDiff. Data utility does not appear to suggest clear explanations for the attenuating MIA success with increasing batch size seen in the TabDiff experiments. Future work is required to more deeply understand the mechanisms at work in these results.

\subsection{TabDiff: Diabetes Quality Results}  

Table \ref{tabdiff_diabetes_quality} exhibits the quality metrics for TabDiff models trained using the Diabetes dataset. The table includes measures across the different configurations reported in Sections \ref{steps_size_model} and \ref{diffusion_batch_ratio}. As noted above, the Diabetes dataset is smaller than Berka, but has more columns and representation diversity. It also tends to exhibit privacy leakage at a higher rate for both ClavaDDPM and TabDiff models. While large synthetic data quality degradation is only present at certain setting extremes, model quality is more sensitive for Diabetes, compared to Berka. For example, When training for only $3000$ steps, $\alpha$ precision is much lower, Avg. KS and MI Diff. are much higher than such values at $24000$ steps. Similar quality drop off is seen when using the smallest model configurations and very few diffusion steps to generate the data. Nevertheless, the tradeoffs around utility and privacy seen for TabDiff models trained on Berka are also seen here.

\begin{table}[ht!]
\centering
\small
\caption{Synthetic data quality metrics across experimental settings for target TabDiff models trained on Diabetes.}
\resizebox{0.99\linewidth}{!}{
\begin{tabular}{llcccccc}
\toprule
Category & Setting
& $\beta$ Cov.$\uparrow$
& $\alpha$ Prec.$\uparrow$
& Avg. KS$\downarrow$
& Avg. TVD$\downarrow$
& Corr. Diff.$\downarrow$
& MI Diff.$\downarrow$ \\
\midrule
Train Steps
& 3e03 & 0.420 & 0.781 & 0.047 & 0.015 & 1.168 & 4.008 \\
& 6e03 & 0.450 & 0.865 & 0.032 & 0.011 & 1.256 & 3.543 \\
& 9e03 & 0.449 & 0.862 & 0.028 & 0.009 & 1.037 & 3.807 \\
& 1.2e04 & 0.444 & 0.822 & 0.025 & 0.008 & 0.967 & 3.375 \\
& 1.5e04 & 0.449 & 0.852 & 0.026 & 0.008 & 0.979 & 2.894 \\
& 1.8e04 & 0.456 & 0.894 & 0.026 & 0.009 & 0.981 & 3.089 \\
& 2.1e04 & 0.452 & 0.891 & 0.026 & 0.008 & 1.102 & 2.337 \\
& 2.4e04 & 0.462 & 0.936 & 0.026 & 0.008 & 0.992 & 2.010 \\
\midrule
Architecture
& 512x1 & 0.443 & 0.845 & 0.026 & 0.008 & 1.016 & 3.624 \\
& 1024x1 & 0.454 & 0.894 & 0.025 & 0.008 & 0.991 & 2.997 \\
& 512x2  & 0.452 & 0.908 & 0.026 & 0.008 & 0.994 & 4.126 \\
& 1024x2 & 0.462 & 0.936 & 0.026 & 0.008 & 0.992 & 2.010 \\
& 512x3  & 0.448 & 0.881 & 0.025 & 0.008 & 1.055 & 2.956 \\
& 1024x3 & 0.456 & 0.892 & 0.024 & 0.008 & 1.005 & 2.162 \\
\midrule
Diffusion Steps
& 5   & 0.102 & 0.242 & 0.308 & 0.087 & 2.437 & 5.748 \\
& 50  & 0.462 & 0.936 & 0.026 & 0.008 & 0.992 & 2.010 \\
& 300 & 0.454 & 0.891 & 0.026 & 0.008 & 1.062 & 2.888 \\
\midrule
Train size
& 5e03  & 0.391 & 0.732 & 0.048 & 0.010 & 1.243 & 3.951 \\
(Equal Synth.) & 1e04 & 0.462 & 0.936 & 0.026 & 0.008 & 0.992 & 2.010 \\
& 2e04 & 0.476 & 0.921 & 0.016 & 0.008 & 0.687 & 2.940 \\
\midrule
Batch size
& 2048 & 0.467 & 0.895 & 0.026 & 0.011 & 1.460 & 4.547 \\
& 4096 & 0.462 & 0.936 & 0.026 & 0.008 & 0.992 & 2.010 \\
& 8192 & 0.432 & 0.834 & 0.028 & 0.008 & 1.255 & 3.946 \\
\midrule
Synth. size
& 5e03  & 0.419 & 0.764 & 0.026 & 0.011 & 1.021 & 4.827 \\
(1e04 Train) & 1e04 & 0.462 & 0.936 & 0.026 & 0.008 & 0.992 & 2.010 \\
& 2e04 & 0.452 & 0.885 & 0.025 & 0.006 & 0.962 & 2.009 \\
\bottomrule
\end{tabular}}
\label{tabdiff_diabetes_quality}
\end{table}

\section{Dataset Preprocessing} \label{dataset_prepocessing}

Prior to training the ClavaDDPM target models in the experiments, both datasets undergo light preprocessing steps. Any columns corresponding to unique identifiers are dropped. Missing values for categorical columns are treated as a distinct category. For Diabetes, some categorical columns also have entries of ``NaN'' and ``?''. These values are similarly treated as distinct categories for the columns in which they appear. For numerical columns in Berka, missing values are imputed as $0$, while no missing numerical values are present for the Diabetes data. Finally, categorical values are ordinally encoded for uniform representation with encoding performed over the entire dataset to avoid issues with less frequent values. For the TabDiff experiments, the Diabetes dataset is preprocessed in an identical way to the ClavaDDPM experiments. For Berka, we use the preprocessed dataset from \citep{mahyar2026fermiexploitingrelationsmembership}.

When computing the metrics discussed in Appendix \ref{quality_metrics} and DCR, preprocessing differed depending on the metric computed. For $\alpha$-precision and $\beta$-coverage, categorical values are one-hot encoded and numerical columns are left unchanged. When computing KS and TVD statistics or Correlation and MI matrix differences, the SynthEval \citep{SynthEval2024} preprocessing approach is applied before downstream computations, which leaves categorical columns ordinally encoded and min-max transforms numerical values. For the MLE quality measures, min-max transforms are also applied to numerical values, but categorical columns are one-hot encoded. Finally, for DCR, categorical variables are one-hot encoded and numerical columns are normalized to the interval $[-1, 1]$.

\section{Compute Resources}

Generally, the experiments conducted in this work are quite computationally intensive. In addition to training the target models, both the TF and Ensemble attacks require training numerous shadow models, each of which is a large tabular diffusion model itself. In the default setting for Berka, the TF MIA trains $20$ such models, while the Ensemble approach constructs $17$, one primary shadow model and $16$ RMIA models. Both attacks also need to train membership classifiers and require non-trivial feature extraction computations. For the TF attack, this involves extracting loss values via forward and backward diffusion processes for each data point for $2100$ ($\epsilon, t)$ pairs for ClavaDDPM and $5000$ such pairs for TabDiff. In the Ensemble MIA, computationally intensive distance-, DOMIAS-, and RMIA-based features need to be derived.

The experiments in this work leverage both A40 and A100 GPUs with 48GB and 80GB of GPU memory, respectively. Depending on the experimental setup, end-to-end attack training and inference times vary. For example, attacks applied to larger models require more time compared to the default settings. However, a single default ClavaDDPM configuration for the TF approach requires approximately $2.5$ hours for Berka and $1$ hour for the Diabetes dataset on A40 GPUs. For an individual setup, the Ensemble attack takes around $16$ hours to complete for Berka and $11$ hours to complete for Diabetes using A100s. This increases to $22$ and $16$ hours, respectively, on A40 GPUs.

\end{document}